\documentclass{article}

  \usepackage[preprint]{neurips_2026}

\usepackage[utf8]{inputenc}
\usepackage[T1]{fontenc}

\usepackage{amsmath}
\usepackage{amssymb}
\usepackage{amsfonts}

\usepackage{graphicx}
\usepackage{booktabs}
\usepackage{multirow}
\usepackage{colortbl}
\usepackage{tabularx}
\usepackage{enumitem}
\usepackage{subcaption}

\usepackage{nicefrac}
\usepackage{siunitx}
\usepackage{xspace}
\usepackage{xcolor}
\usepackage{url}
\usepackage{makecell} 
\usepackage[ruled]{algorithm2e}

\usepackage{kotex}

\usepackage{microtype}

\usepackage{hyperref}

\usepackage[capitalize]{cleveref}

\newcommand{\reflectgs}{RGS}

\newcommand{\first}[1]{\cellcolor{red!25}#1}
\newcommand{\secondbest}[1]{\cellcolor{orange!25}#1}
\newcommand{\thirdbest}[1]{\cellcolor{yellow!25}#1}

\makeatletter
\DeclareRobustCommand\onedot{%
  \futurelet\@let@token\@onedot
}
\def\@onedot{%
  \ifx\@let@token.\else.\null\fi\xspace
}
\makeatother

\newcommand{\eg}{\emph{e.g}\onedot}

\newcommand{\vs}{\emph{vs}\onedot}

\newcommand{\paragrapht}[1]{%
\vspace{0.5em}\noindent\textbf{#1.}\ }

\usepackage{amsmath,amsfonts,bm}

\def\eqref#1{equation~\ref{#1}}

\def\1{\bm{1}}

\def\vb{{\bm{b}}}
\def\vc{{\bm{c}}}

\def\vl{{\bm{l}}}

\def\vp{{\bm{p}}}

\def\vs{{\bm{s}}}
\def\vt{{\bm{t}}}
\def\vu{{\bm{u}}}

\def\vx{{\bm{x}}}

\def\mN{{\bm{N}}}

\def\mZ{{\bm{Z}}}

\DeclareMathAlphabet{\mathsfit}{\encodingdefault}{\sfdefault}{m}{sl}
\SetMathAlphabet{\mathsfit}{bold}{\encodingdefault}{\sfdefault}{bx}{n}

\title{Pygmalion Effect in Vision: Image-to-Clay \\ Translation for Reflective Geometry Reconstruction}

\author{
Gayoung Lee$^{1,2}$ \quad
Junho Kim$^{1}$ \quad
Jin-Hwa Kim$^{1,3}$\setcounter{footnote}{1}\thanks{Corresponding authors.} \quad
Junmo Kim$^{2}$\footnotemark[2] \\
\texttt{{\footnotesize
\{gayoung.lee, jhkim.ai, j1nhwa.kim\}@navercorp.com, junmo.kim@kaist.ac.kr
}} \\[-4pt]
\\
$^{1}$NAVER AI Lab \quad
$^{2}$KAIST \quad
$^{3}$SNU AIIS
}

\begin{document}

\maketitle

\begin{abstract}
Understanding reflection remains a long-standing challenge in 3D reconstruction due to the entanglement of appearance and geometry under view-dependent reflections. In this work, we present the Pygmalion Effect in Vision, a novel framework that metaphorically “sculpts” reflective objects into clay-like forms through image-to-clay translation. 
We then introduce a dual-branch design in which a BRDF-based reflective branch and a clay-guided branch share the same Gaussian geometry but operate through independent rendering paths. The clay-guided branch, supervised by the synthesized clay images, provides reflection-free geometric guidance that complements the photometric supervision of the BRDF branch. Experiments on both synthetic and real datasets show consistent improvements in both geometry and photometric metrics. Beyond technical gains, our framework reveals that seeing by unshining, translating radiance into neutrality, can serve as a powerful inductive bias for reflective object geometry learning.
\end{abstract} 
\section{Introduction}
\label{sec:intro}
Reconstructing three-dimensional structures from images has long been a fundamental problem in computer vision.
The emergence of Neural Radiance Fields (NeRF)~\cite{mildenhall2020nerf,barron2021mip,muller2022instant} marked a major breakthrough in continuous scene representation and photorealistic novel-view synthesis. Subsequent studies~\cite{yariv2021volume,wang2021neus,azinovic2022neural} bridge neural rendering and mesh-based modeling by extracting explicit geometry.
More recently, 3D Gaussian Splatting (3DGS)~\cite{kerbl20233d} introduced a point-based differentiable representation that enables real-time, high-fidelity rendering without neural queries at inference.

Despite these advances, reconstructing reflective objects remains a significant challenge. While NeRF- and 3DGS-based models can capture certain non-Lambertian cues, they still implicitly rely on view-invariant radiance consistency, which reflective surfaces violate since their appearance changes with view-dependent environmental reflections. To address this, several inverse-rendering approaches attempt to recover surface reflectance and environment illumination by explicitly modeling specular reflectance parameters within a BRDF formulation~\cite{ngan2005experimental,low2012brdf,yao2025reflective}. However, jointly optimizing geometry and material properties is challenging, as appearance and shape are entangled through reflection, making it difficult to attribute observed color variations to either geometry or reflectance. A parallel line of work~\cite{tong2025gs,zhangmaterialrefgs} introduces auxiliary normal supervision from off-the-shelf estimators to constrain geometry directly. Yet such signals are tightly coupled to the reflective rendering path: even small normal errors are amplified when computing reflection directions for environment-map lookup. Moreover, accurate per-pixel normal estimation on reflective surfaces remains an open problem.

\begin{figure}[t]
  \centering
  \setlength{\tabcolsep}{2pt} 

  \begin{tabular}{c c c c}
    \includegraphics[width=0.23\linewidth]{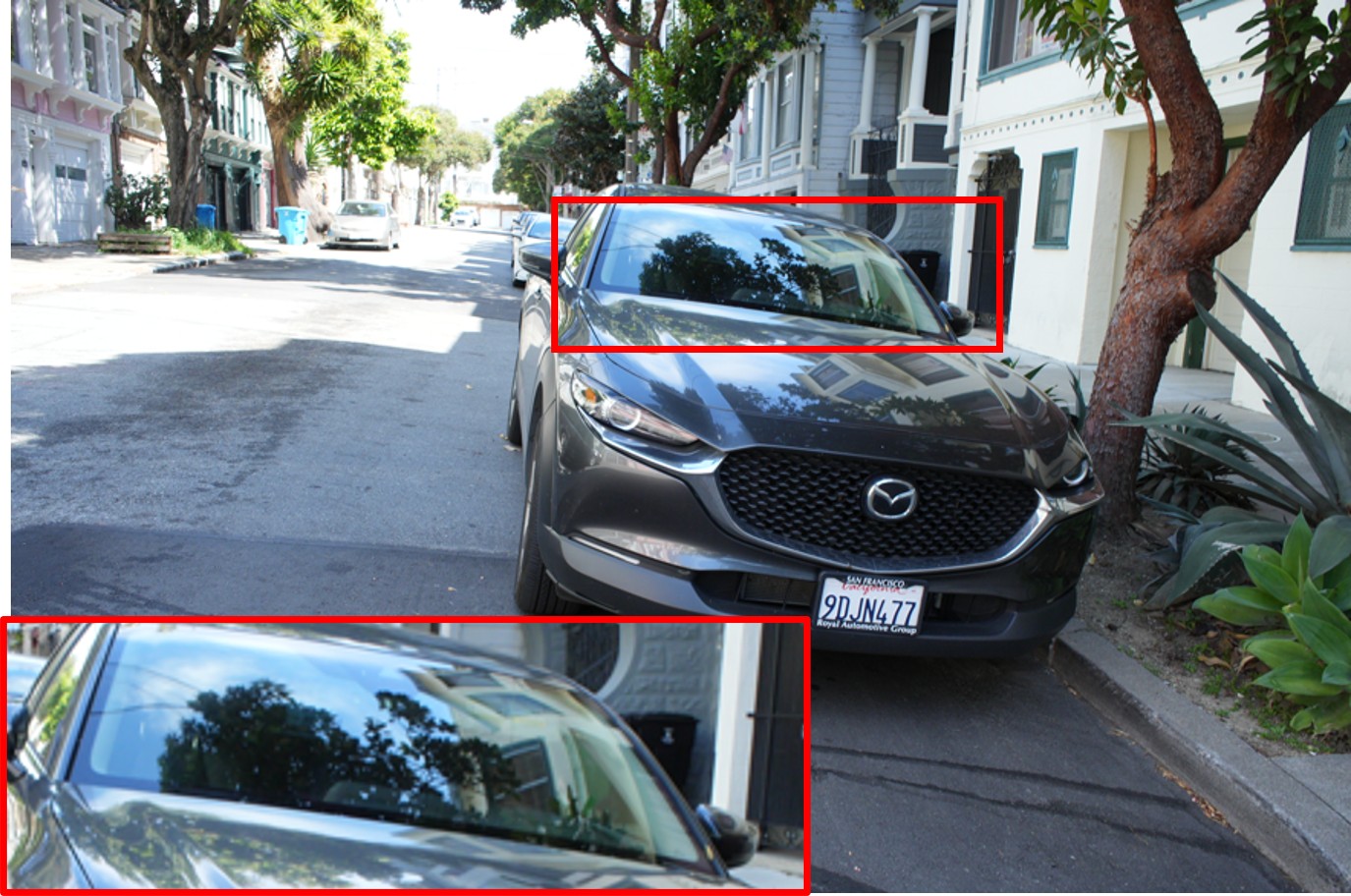} &
    \includegraphics[width=0.23\linewidth]{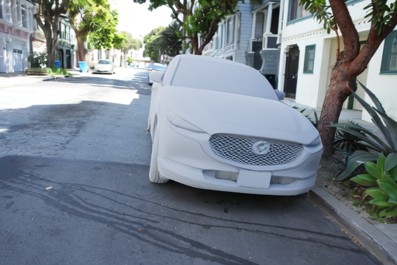} &
    \includegraphics[width=0.23\linewidth]{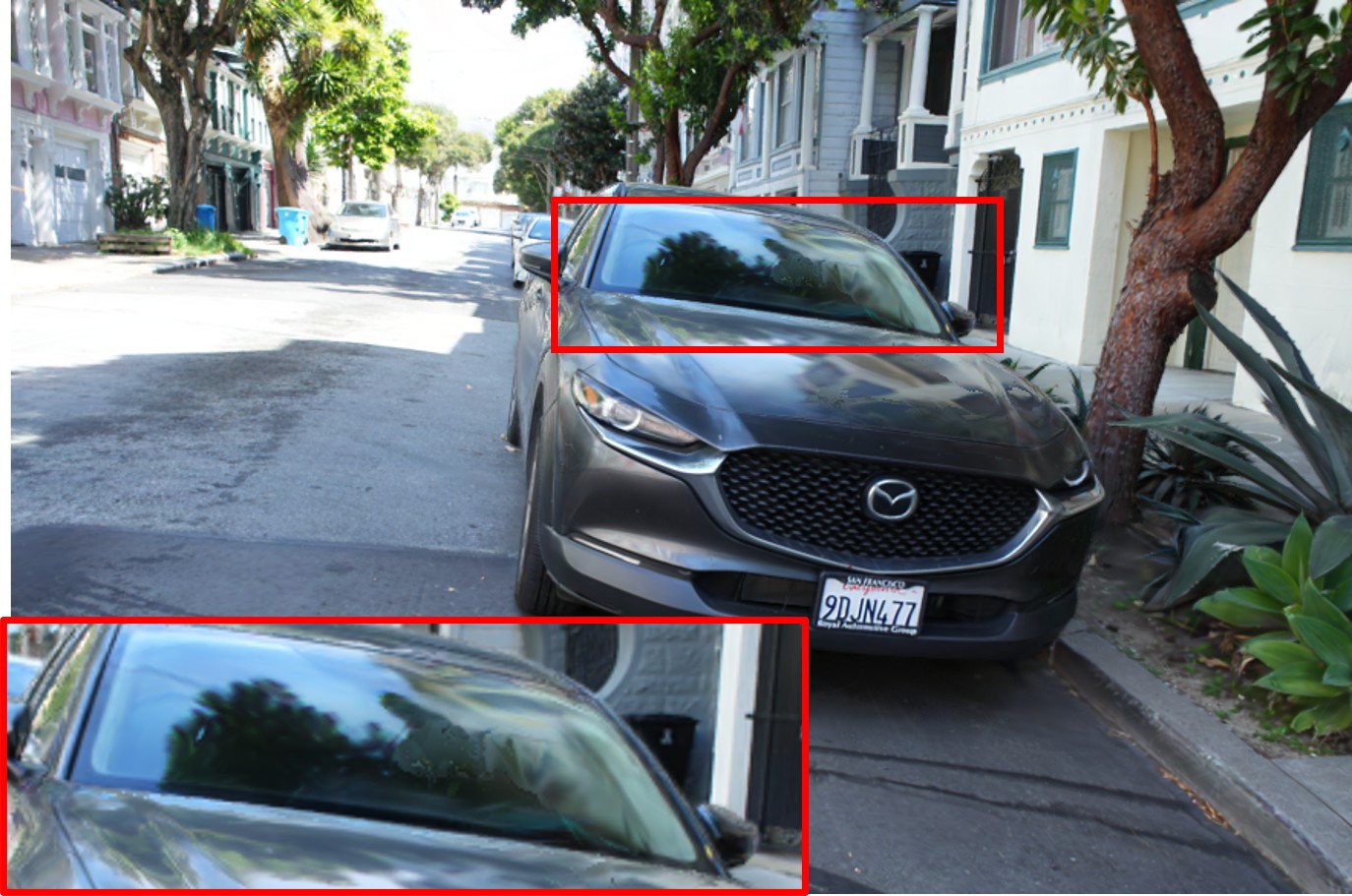} &
    \includegraphics[width=0.23\linewidth]{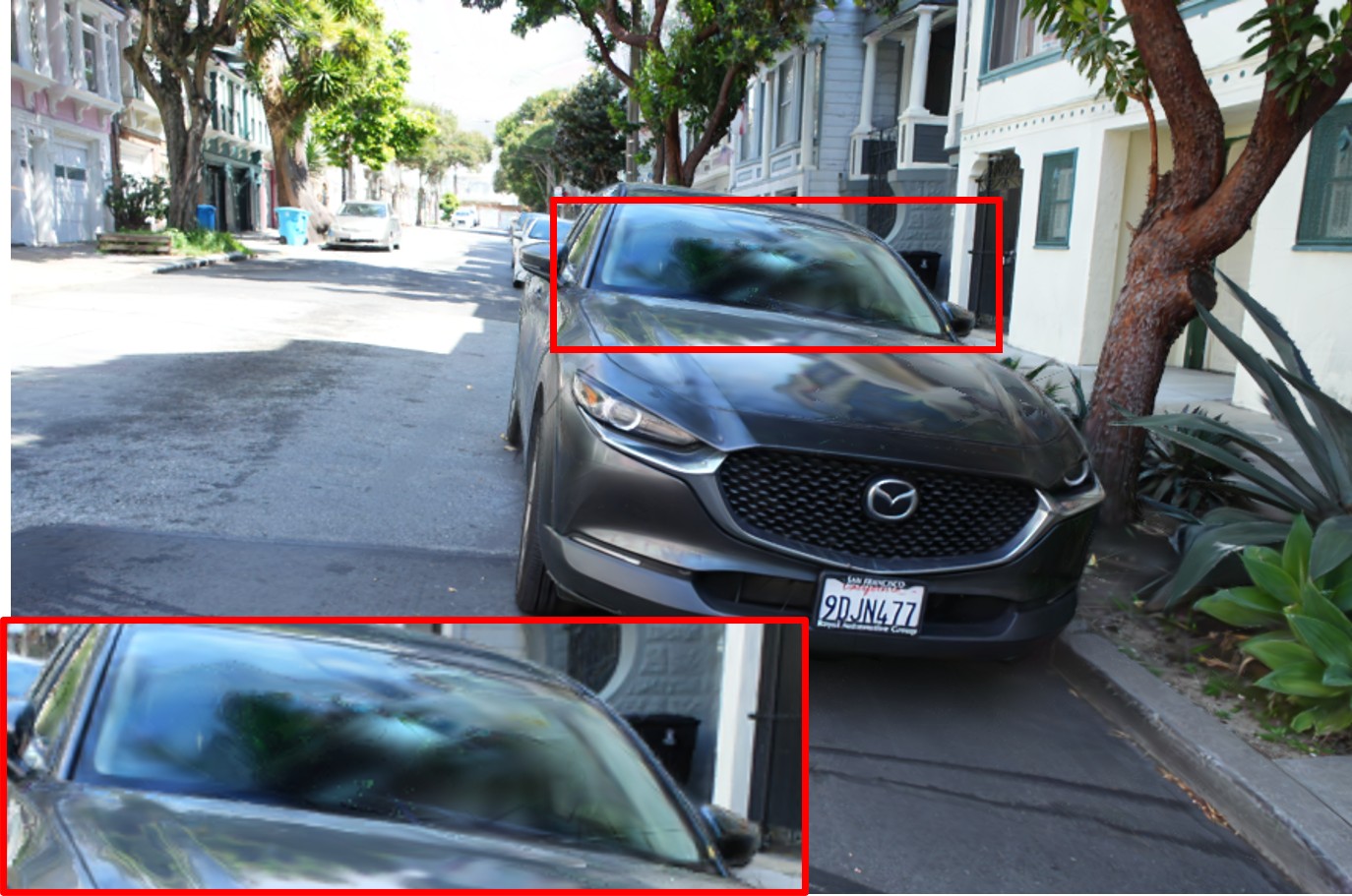} \\

    \includegraphics[width=0.23\linewidth]{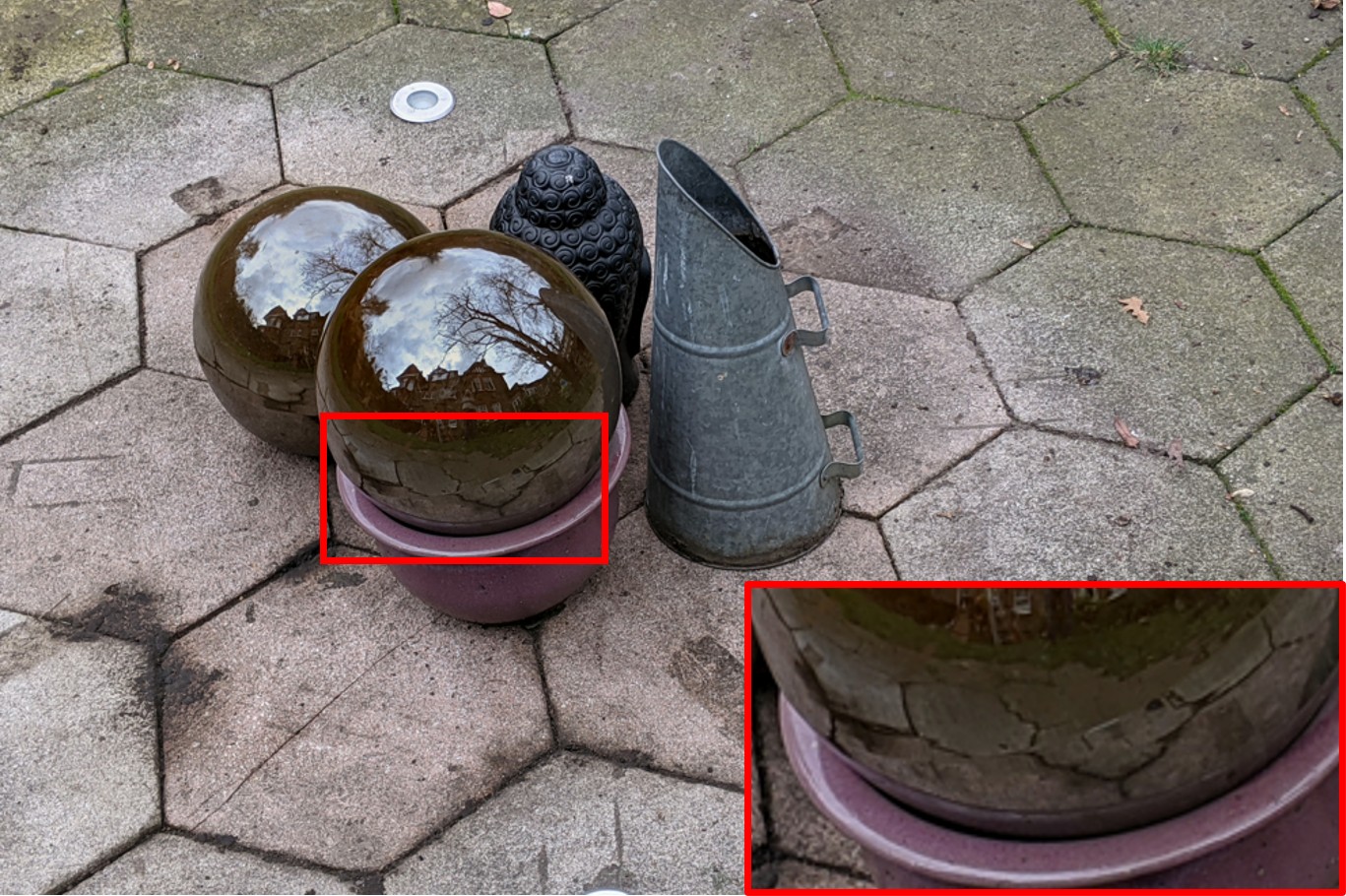} &
    \includegraphics[width=0.23\linewidth]{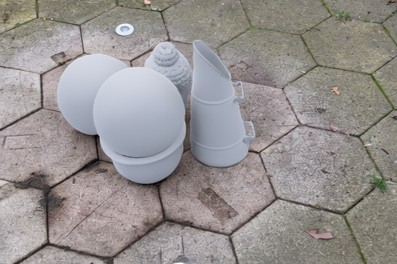} &
    \includegraphics[width=0.23\linewidth]{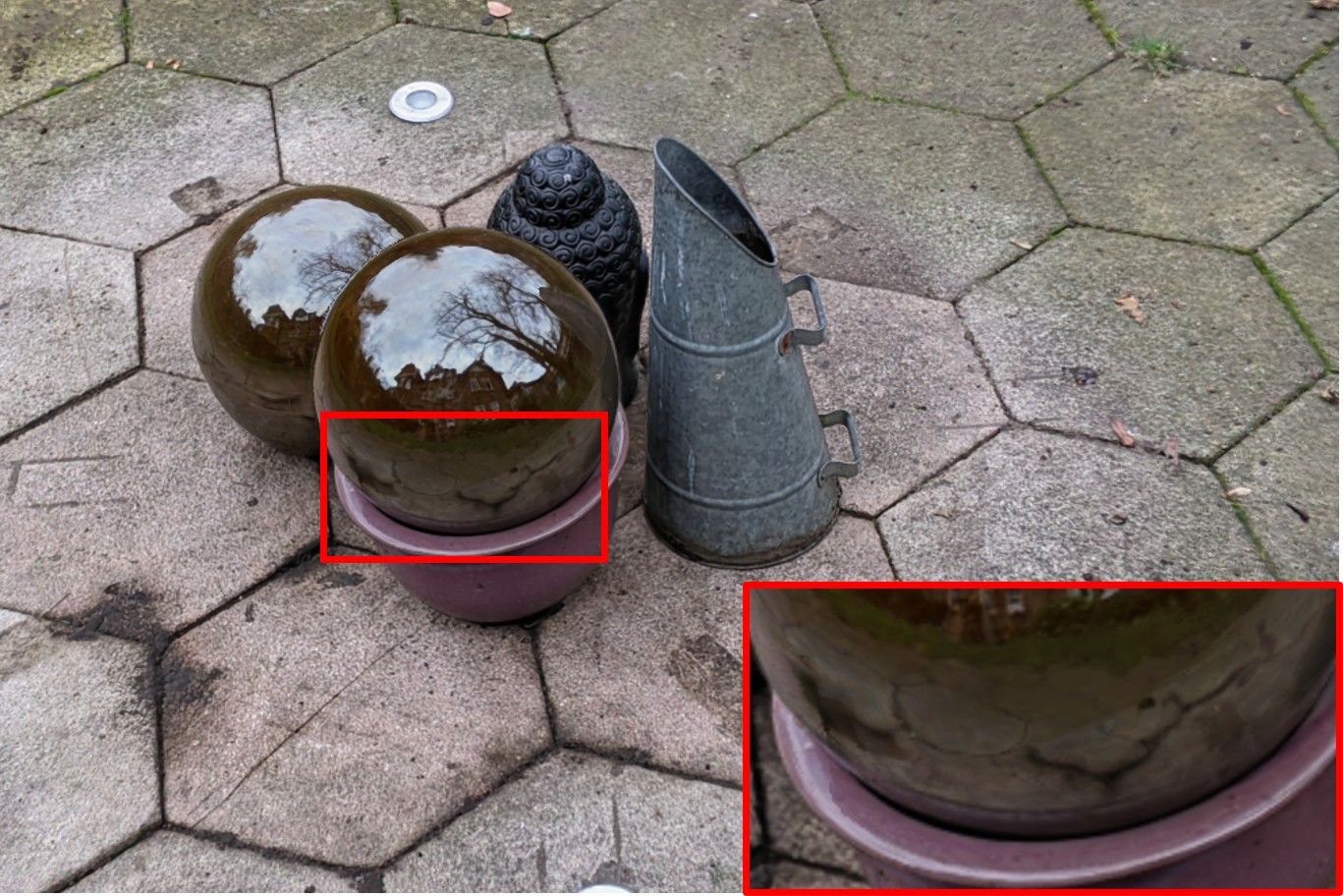} &
    \includegraphics[width=0.23\linewidth]{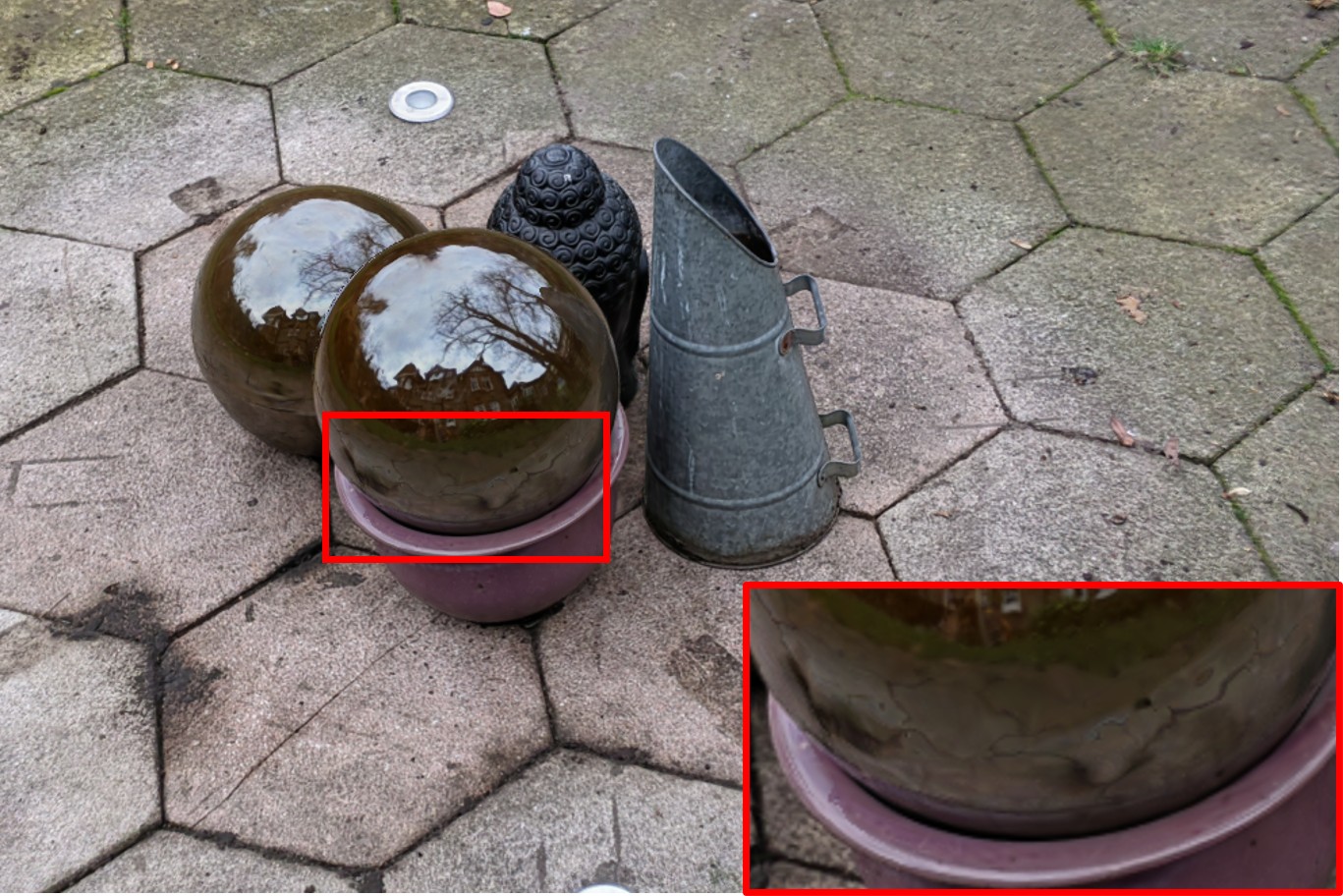} \\
    (a) Ground Truths & (b) Clay Images & (c) Ours & (d) MaterialRefGS 
  \end{tabular}

  \caption{By translating reflective images (a) into clay-like renderings (b) for geometric guidance, our method produces sharper, more accurate reflective-surface geometry (c) than a strong baseline (d).}
  \label{fig:teasure_neurips}
\end{figure}

In contrast to prior works that build increasingly complex models for reflective objects, we reformulate the problem as reconstruction by converting reflective appearances into non-reflective ones using image-to-image translation models. For this conversion, we adopt white clay as the target material. Clay renders, widely used in pipelines such as Blender~\cite{blender}, effectively convey geometric structure without interference from material appearance. This process can be viewed as an instance of the Pygmalion effect in vision (for our analogy, see \cref{appx:analogy}), where the model learns to reshape the visual world according to its own learned expectations, turning reflective complexity into a self-consistent, interpretable form.
We find that clay-like appearances can be reliably synthesized by modern generative models, due to the abundance of similar matte-white objects in real-world images. To ensure cross-view consistency of the translated clay images, we adapt the pretrained video diffusion model LTX-2~\cite{hacohen2026ltx} with lightweight LoRA~\cite{hu2022lora} fine-tuning, reinterpreting its temporal coherence as a view-consistency prior.

To leverage these clay images for geometry optimization, we introduce a dual-branch reflective Gaussian splatting design that can be readily integrated into existing reflective splatting frameworks. Our clay-guided branch shares the Gaussian geometry with the BRDF-based branch but uses an auxiliary clay color parameter, decoupling reflection-free geometric guidance from specular evaluation. We adopt MaterialRefGS~\cite{zhangmaterialrefgs} as our primary backbone and show that adding clay supervision substantially improves geometry. Notably, our clay supervision is complementary to the monocular normal prior estimated by an off-the-shelf model, operating through an independent rendering path. Moreover, we will show that it can stand alone as a sufficient source of geometric guidance, even without any normal prior. Experimental results on multiple datasets demonstrate superior mesh quality compared to existing approaches, as illustrated in \cref{fig:teasure_neurips}.

In summary, our contributions are summarized as follows:
\begin{itemize}[leftmargin=2em, itemsep=0pt, topsep=0pt]
    \item A novel framework that translates reflective images into clay-like renderings as a reflection-free supervision signal, providing geometric guidance.
    \item A multi-view-consistent clay translation model built on LTX-2~\cite{hacohen2026ltx}, a pretrained video diffusion backbone. We reinterpret its temporal axis as a view axis, enabling cross-view coherence with only lightweight LoRA adaptation.
\item A dual-branch reflective Gaussian splatting design that augments existing BRDF-based pipelines with a clay-guided branch sharing the same geometry, providing reflection-free geometric supervision through an independent rendering path.
    \item Experimental validation demonstrates clear and consistent gains in mesh accuracy and reconstruction stability across multiple reflective object datasets.  
\end{itemize}

\section{Related work}
\label{sec:related_work}

\paragrapht{Novel-view synthesis with implicit geometry}
NeRF~\cite{mildenhall2020nerf} introduced volumetric radiance fields, inspiring extensions for anti-aliasing~\cite{barron2021mip}, training efficiency~\cite{muller2022instant}, and high-fidelity surface reconstruction~\cite{li2023neuralangelo,wang2023neus2}. In parallel, surface-based approaches~\cite{yariv2021volume,oechsle2021unisurf,yu2022monosdf} explicitly model signed-distance fields for improved geometry. Recently, point-based representations such as 3D Gaussian Splatting~\cite{kerbl20233d} and 2DGS~\cite{huang20242d} have enabled real-time rendering with improved geometric consistency.

\paragrapht{Reflective reconstruction}
Reconstructing reflective objects is challenging because appearance and geometry are entangled under view-dependent color variations.  
Early attempts such as TensoSDF~\cite{li2024tensosdf} and NeRO~\cite{liu2023nero} combine volumetric radiance fields with explicit reflectance modeling to jointly learn geometry and BRDF parameters.  
Gaussian-based approaches address the same issue through differentiable point representations. 
GShader~\cite{jiang2024gaussianshader} integrates shading functions into Gaussian splatting for reflective rendering, while GS-IR~\cite{liang2024gsir} formulates inverse rendering within the Gaussian domain to estimate reflectance and lighting.  
R3DG~\cite{gao2024r3dg} introduces a ray-tracing formulation to better handle high-frequency specular highlights. Further, Ref-GS~\cite{zhang2025refgs} and \reflectgs{}~\cite{yao2025reflective} focus on disentangling view-dependent radiance for more accurate reflective reconstruction. GS-2DGS~\cite{tong2025gs} and MaterialRefGS~\cite{zhangmaterialrefgs} enhance geometric consistency by supervising reflective Gaussians with external models. Recent works~\cite{zhangmaterialrefgs, verbin2024nerfcasting, moenne20243dgs_raytracing} also utilize ray-tracing to more accurately model indirect illumination. 

S2Dnet~\cite{wu2018specular} previously explored the idea of translating specular images into diffuse counterparts for geometry reconstruction, using a multi-view-aware GAN. However, S2Dnet relies on local correspondence-based consistency and uses diffuse translation as preprocessing for classical MVS, without BRDF learning and with limited mesh reconstruction accuracy. In contrast, our approach leverages a video diffusion prior with an anchor mechanism for cross-view consistency and integrates clay supervision into reflective Gaussian splatting optimization to support both geometry reconstruction and BRDF learning.  An empirical comparison is provided
in \cref{appx:s2dnet}.

\paragrapht{Image and video editing}
Image-to-image translation learns mappings between visual domains for controllable appearance transformation using paired adversarial learning~\cite{isola2017image}, unpaired cycle-consistency~\cite{zhu2017unpaired}, disentangled multimodal~\cite{huang2018multimodal}, contrastive~\cite{park2020contrastive} formulations, and unified spatial-semantic control~\cite{tan2025ominicontrol}. 
Extending these ideas to the temporal domain, instruction-based video editors~\cite{mou2025instructx,yu2026vifeedit} and stylization methods~\cite{li2026dreamstyle,zhang2026telestyle,zhang2026omnitransfer} enable coherent edits across frames. Unlike prior works that primarily focus on temporal coherence, our goal is multi-view consistent material translation: we reinterpret the temporal coherence prior of a pretrained video diffusion model as a view-consistency prior, generating clay-like views that provide reflection-free supervision for accurate and efficient geometry optimization.
\begin{figure*}[t]
  \centering
  \includegraphics[width=1.0\linewidth]{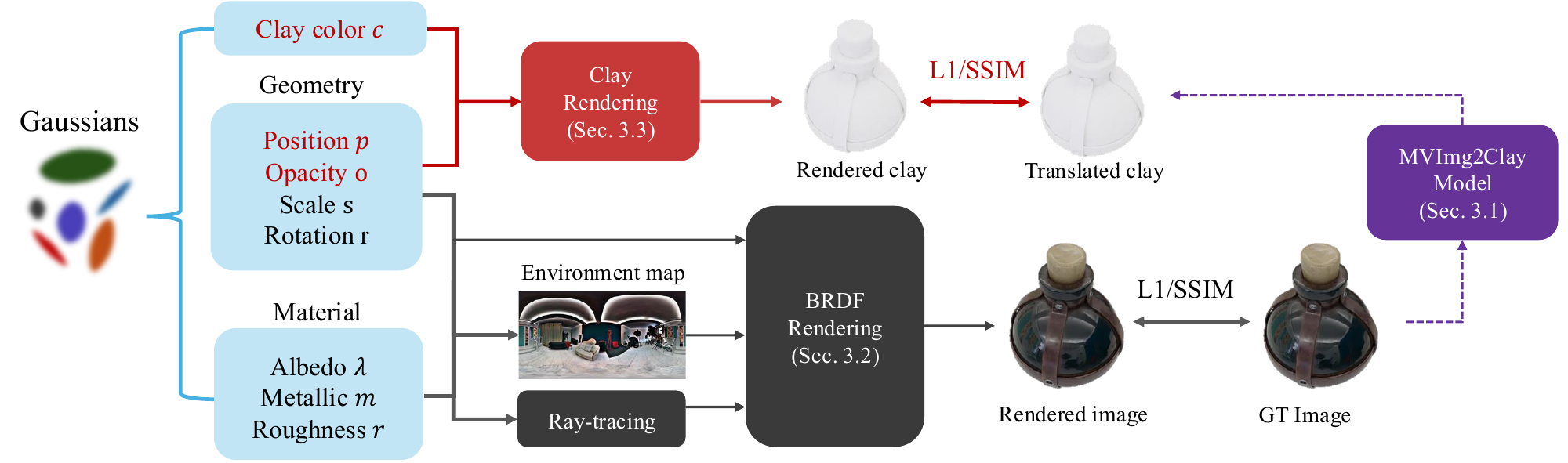}
  \caption{Overview of our dual-branch pipeline.
A BRDF-based reflective branch and a clay-guided branch share the same Gaussian geometry.
Clay-like images provide reflection-free supervision during early training, stabilizing geometry and improving surface normals.}
  \label{fig:model_pipe}
\end{figure*}

\section{Method}

Our overall framework is illustrated in \cref{fig:model_pipe}. We first develop 
a multi-view image-to-clay translation network in \cref{sec:im2clay_model}, 
trained on the dataset described in \cref{appx:data-prep-i2c}. We then introduce our 
clay-guided reflective Gaussian splatting method in \cref{sec:refl_gs} and 
\cref{sec:clayrs}.

\begin{figure}[t]   
  \centering
  \includegraphics[width=0.99\linewidth]{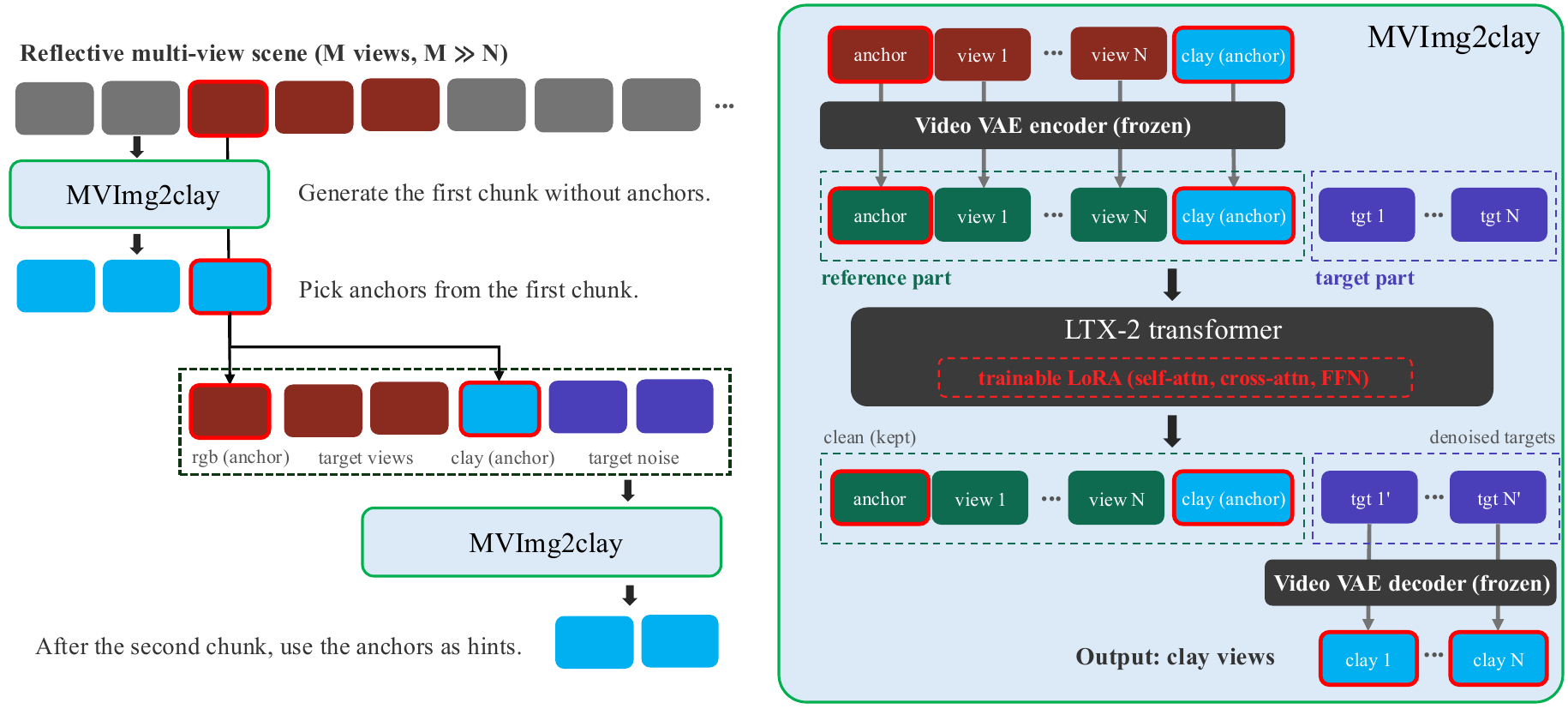}
  \caption{Pipeline of our LTX-2-based multi-view image-to-clay translation model. We arrange multi-view images along the temporal axis of the video diffusion model, repurposing temporal coherence as cross-view consistency. For scenes split into multiple chunks, we use anchor views and their clay translations from the first chunk as conditioning to maintain consistency across chunks.}
  \label{fig:multiview_pipeline}
\end{figure}

\subsection{Multi-view image-to-clay translation}
\label{sec:im2clay_model}

To translate reflective images into diffuse clay-like renderings that can serve as multi-view geometric supervision, we require a translation model that not only suppresses specular cues but also produces outputs consistent across viewpoints. Per-image translation models~\cite{wu2025qwen,tan2025ominicontrol} can produce high-quality clay translations on individual images, but offer no mechanism to enforce coherence across views. 

We address this by repurposing a pretrained video diffusion model, LTX-2~\cite{hacohen2026ltx}, which natively enforces coherence across video frames through its temporal attention. Our key insight is that the temporal axis of a video model can be reinterpreted as a view axis. While multi-view images are not a continuous video sequence in the strict sense, we find that the inter-frame coherence prior learned by the video model transfers naturally to multi-view inputs with only lightweight fine-tuning, producing translations that remain consistent across views. We provide a brief overview of LTX-2 and In-Context LoRA in \cref{appx:ltx_preliminary}. An overview of our multi-view image-to-clay translation pipeline is shown in \cref{fig:multiview_pipeline}.

\paragraph{Multi-view sequence construction.}
Given $N$ multi-view images of a reflective object, we encode each view independently using LTX-2's video VAE encoder, producing $N$ single-frame latent tokens. We then concatenate these latents along the temporal axis and use them as conditioning.

The full token sequence consists of two parts:
\begin{equation}
    \underbrace{[\mZ_1^{\text{cond}}, \dots, \mZ_N^{\text{cond}}]}_{\text{conditioning view (clean)}} \;\big|\; \underbrace{[\mZ_1^{\text{tgt}}, \dots, \mZ_N^{\text{tgt}}]}_{\text{target (noisy)}},
\end{equation}
where the conditioning part contains the input reflective views as clean condition tokens, and the target part contains the corresponding clay views as the diffusion target. 
Following IC-LoRA~\cite{huang2024iclora}, we assign per-token timesteps to distinguish conditioning tokens within a unified sequence:
\begin{equation}
    t_i = \begin{cases} 0 & \text{if } m_i = 1 \\ \sigma & \text{if } m_i = 0 \end{cases},
\end{equation}
where $m_i$ denotes conditioning mask and $\sigma$ is the noise level sampled from a shifted logit-normal distribution. This per-token timestep design lets the transformer treat reference tokens as fully observed while denoising target tokens, all within a single forward pass.

Within the pseudo-video sequence, we order the views randomly rather than by camera angle proximity. We empirically found that random ordering generalizes better across datasets with diverse camera trajectories. 

\paragraph{Training data and objective.}
We train the translation model on a dataset of paired multi-view renderings. Each pair shares the same camera pose, HDRI lighting, and geometry, but differs in material. The full rendering details are provided in \cref{appx:data-prep-i2c}.

  We train the lightweight LoRA~\cite{hu2022lora} using a standard flow-matching velocity prediction objective. Given a clean target latent $\mZ^{\text{tgt}}$ and a noise sample $\boldsymbol{\epsilon}$, the noisy latent is $\mZ_\sigma = (1 - \sigma) \mZ^{\text{tgt}} + \sigma \boldsymbol{\epsilon}$, and the model predicts the velocity $\mathbf{v} = \boldsymbol{\epsilon} - \mZ^{\text{tgt}}$. We apply the loss only on noisy target tokens, masking out reference tokens and any clean anchor tokens:
\begin{equation}
    \mathcal{L}_{\text{flow}} = \mathbb{E}_{\sigma, \boldsymbol{\epsilon}} \left[ \sum_{i: m_i = 0} \| \mathbf{v}_\theta(\mZ_\sigma; t)_i - (\boldsymbol{\epsilon}_i - \mZ^{\text{tgt}}_i) \|_2^2 \right],
\end{equation}
where $\mathbf{v}_\theta$ is the LTX-2 transformer with our LoRA adaptation.

\begin{figure*}[t]
\centering
\begin{minipage}[c]{0.55\textwidth}
    \centering
    \includegraphics[width=\linewidth]{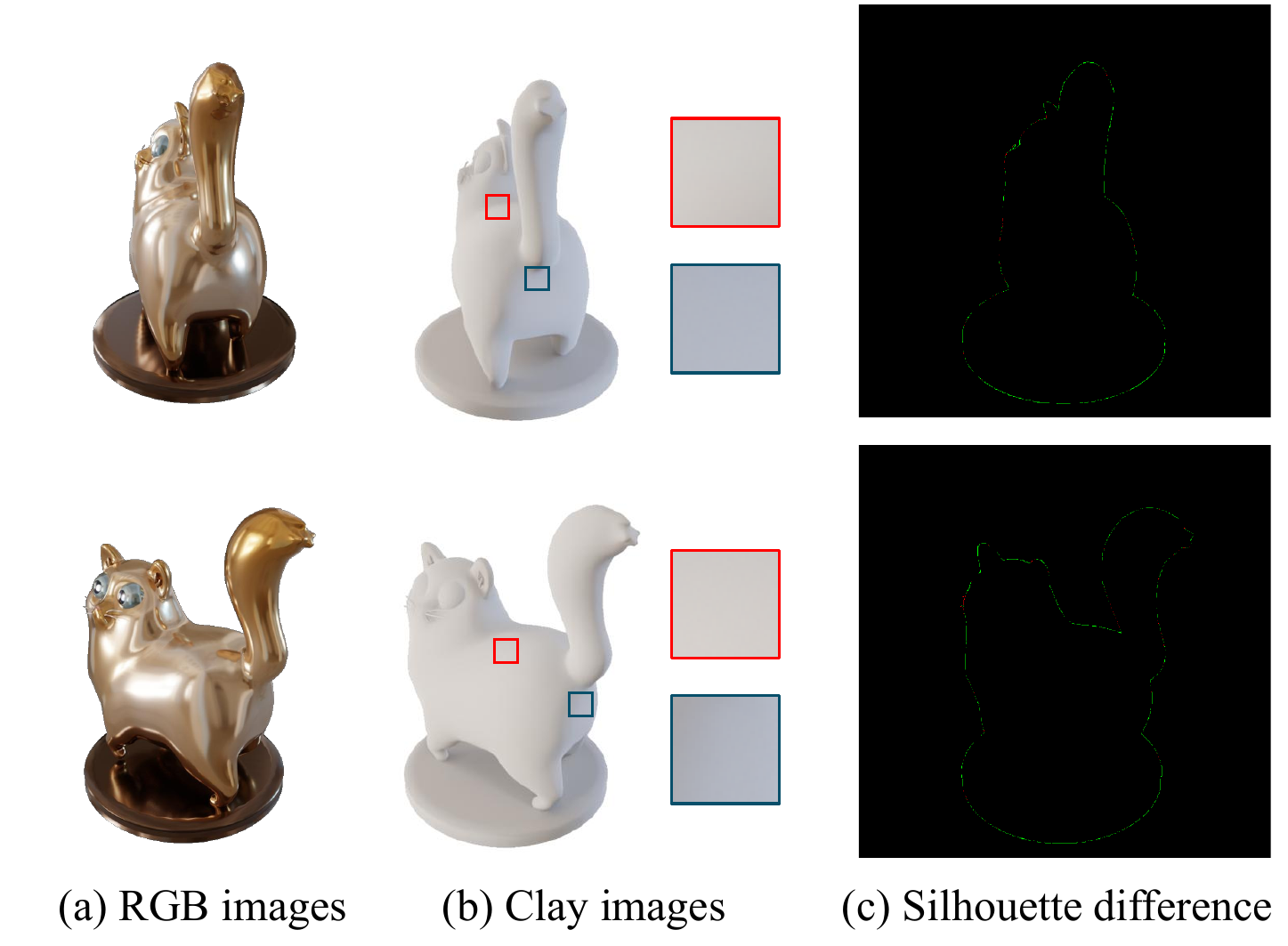}
\end{minipage}%
\hfill
\begin{minipage}[c]{0.42\textwidth}
    \centering
    \captionof{table}{\textbf{Quantitative analysis of clay image consistency.} 
    Top: foreground/background MSE between RGB and clay images, and IoU between the clay and the GT mask. 
    Bottom: cross-view color variance at the same 3D point compared with the overall variance.}
    \label{tab:clay_analysis}
    \vspace{1pt}
    \footnotesize
    \renewcommand{\arraystretch}{1.1}
\begin{tabular*}{\linewidth}{@{\hspace{15pt}\extracolsep{\fill}}ccc@{\hspace{15pt}}}
        \toprule
        MSE$_{\text{fg}}$ & MSE$_{\text{bg}}$ & Mask IoU \\
        \midrule
        0.1407 & 0.0001 & 0.9915 \\
        \bottomrule
    \end{tabular*}
    \\[3pt]
    \begin{tabular*}{\linewidth}{@{\hspace{6pt}\extracolsep{\fill}}cccc@{\hspace{6pt}}}
        \toprule
        Type & Cross var. & Global var. & Ratio $R$$\downarrow$ \\
        \midrule
        RGB  & 0.0575     &  0.1318     & 0.4109     \\
        Clay & 0.0163 & 0.0436 & 0.3661 \\
        \bottomrule
    \end{tabular*}
\end{minipage}
\caption{\textbf{Clay images preserve object silhouettes and view-consistent appearance.} 
Visualization of clay images alongside GT RGB on Cat (from Glossy Synthetic). The clay translation removes view-dependent specular cues while preserving object silhouettes.}
\label{fig:clay_analysis}
\end{figure*}

\paragraph{Anchor mechanism and inference for arbitrary view counts.}
\label{anchor}
Training uses fixed-length sequences ($N = 24$), but real scenes often
contain hundreds of views. To bridge this gap, we introduce an anchor
mechanism trained jointly with the denoising objective. During training, with
probability $0.5$ we replace $n_{\text{anchor}} = 4$ randomly selected target
slots with clean ground-truth clay latents, while the remaining targets are
denoised as usual; otherwise, all targets remain noisy. This yields two modes
within one model: anchor-free generation, and inpainting-like generation
conditioned on a few clean anchor translations. At inference, for a scene
with $M \gg N$ views, we partition the views into chunks of size $N$. The
first chunk is generated in the anchor-free mode; $n_{\text{anchor}}$ of its
input views (\emph{anchors}) and their translated outputs
(\emph{anchor-clay} images) are then injected into every subsequent chunk,
maintaining coherence across the full capture. A fixed text prompt
describing the clay material is provided in both modes.

\paragraph{Silhouette preservation and view consistency.} 
\cref{fig:clay_analysis} and \cref{tab:clay_analysis} show that our generated clay images effectively suppress view-dependent color variations while preserving object silhouettes.
\cref{fig:clay_analysis} (b) shows that the same surface point across different views is mapped to consistent colors.
\cref{fig:clay_analysis} (c) shows the silhouette mismatch between GT and clay-derived masks (red: GT-only, green: clay-only). Discrepancies appear only along thin boundaries, consistent with the near-zero background MSE and near-unity mask IoU in \cref{tab:clay_analysis} (top). The bottom of \cref{tab:clay_analysis} reports the color variance at corresponding 3D points obtained via GT depth. Variances and ratios are computed per scene and then averaged. Although clay images appear visually more uniform than the original images, the cross-view variance at a given surface point is substantially smaller than the global variance, yielding a lower variance ratio than the original images. This indicates that the clay translation produces view-independent appearances, providing an ideal supervision signal for geometry learning. We further explain the metric and analyze this in \cref{sec:ablation}. Additional examples of the generated clay images are provided in \cref{appx:additional_clayvis}.

\subsection{Reflective Gaussian splatting baseline}
\label{sec:refl_gs}

We build upon MaterialRefGS~\cite{zhangmaterialrefgs}, a deferred shading-based reflective Gaussian Splatting framework that extends 2D Gaussian Splatting (2DGS)~\cite{huang20242d} with physically-based material modeling. Following recent BRDF-based reflective splatting approaches~\cite{yao2025reflective, zhang2025refgs, tong2025gs}, each 2D Gaussian primitive is augmented with material attributes such as albedo $\lambda$, metallic $m$, roughness $r$, and surface normal $\mathbf{n}$, which are alpha-blended into screen-space material maps:
\begin{equation}
    \mathbf{B}(\vx) = \sum_{i=1}^{N} \vb_i\, \alpha_i\, \hat{\mathcal{G}}_i(\mathbf{u}(\vx))\prod_{j=1}^{i-1} \left( 1 - \alpha_j\, \hat{\mathcal{G}}_j(\mathbf{u}(\vx)) \right),
    \label{eq:blending_function}
\end{equation}
where $\vb_i = [\lambda_i, m_i, r_i, \mathbf{n}_i]$ are the per-Gaussian material parameters, and $\hat{\mathcal{G}}_i$, $\alpha_i$ denote the 2D Gaussian and opacity. The rasterized material maps are then evaluated under a physically-based BRDF, where the specular term is approximated via the split-sum formulation~\cite{munkberg2022splitsum}, separating the integral into a view-dependent BRDF lookup and a prefiltered environment-map evaluation. A complete formulation of 2DGS rasterization, BRDF, and split-sum integration is provided in \cref{appx:2dgs}.

Beyond the standard BRDF framework, MaterialRefGS introduces a multi-view material consistency loss $\mathcal{L}_{mv}$ that aligns rasterized material maps across nearby views via homography warping. Also, a monocular normal prior $\mathcal{L}_n$ from a pretrained estimator~\cite{hu2024metric3d} provides auxiliary geometric supervision. Indirect illumination is additionally handled through differentiable Gaussian ray tracing~\cite{xie2024envgs}. The total baseline loss is
\begin{equation}
    \mathcal{L}_{\text{base}} = \mathcal{L}_{rgb} + \lambda_{n\text{-}d} \mathcal{L}_{n\text{-}d} 
    + \lambda_n \mathcal{L}_n + \lambda_\text{mv} \mathcal{L}_\text{mv} + \lambda_\text{ref} \mathcal{L}_\text{ref},
\end{equation}
combining photometric, depth-normal consistency, normal prior, multi-view material, and reflection-strength regularization terms.

\subsection{Clay-guided reflective Gaussian splatting}
\label{sec:clayrs}

In addition to the BRDF-based rendering branch (\emph{BRDF branch}) of MaterialRefGS (\cref{sec:refl_gs}), we introduce an auxiliary \emph{clay branch} that renders view-independent, clay-like images supervised by our multi-view image-to-clay translation outputs (\cref{sec:im2clay_model}). The two branches share the same underlying Gaussian geometry but operate on independent appearance parameters, providing geometric supervision via a rendering path that is fully decoupled from the specular color evaluation.

To be specific, we introduce an additional clay color parameter $\hat{\vc}_i$  for each Gaussian $i$, alongside the existing BRDF parameters $[\lambda_i, m_i, r_i, \mathbf{n}_i]$. Both BRDF and clay branches share the same geometric attributes: position $\vp_i$, tangential direction $\vt_i$, scaling $\vs_i$, opacity $\alpha_i$, and surface normal $\mathbf{n}_i$. So, the geometric supervision from the clay branch directly improves the shared geometry. The clay color is rendered through the same alpha blending used for material maps:
\begin{equation}
    \hat{\vc}(\vx) = \sum_{i=1}^{N} \hat{\vc}_i\, \alpha_i\, \hat{\mathcal{G}}_i(\vu(\vx)) \prod_{j=1}^{i-1} \left( 1 - \alpha_j\, \hat{\mathcal{G}}_j(\vu(\vx)) \right).
\end{equation}

Unlike the specular branch, $\hat{\vc}(\vx)$ is view-independent by design as the clay surfaces exhibit only diffuse shading variations. The clay branch is supervised by the multi-view-consistent clay images $\mathbf{I}_{\text{clay}}$ generated by our translation model, using a weighted combination of $\mathcal{L}_1$ and SSIM losses:
\begin{equation}
\begin{split}
    \mathcal{L}_{\text{clay}} =\;& (1 - \lambda_{\text{dssim}})\, \|\hat{\mathbf{I}}_{\text{clay}} - \mathbf{I}_{\text{clay}}\|_1
    + \lambda_{\text{dssim}}\, \big(1 - \text{SSIM}(\hat{\mathbf{I}}_{\text{clay}}, \mathbf{I}_{\text{clay}})\big),
\end{split}
\end{equation}
with $\lambda_{\text{dssim}} = 0.2$, following the standard photometric weighting in 2DGS-based methods.

\paragraph{Selective gradient routing.}
A naive joint optimization can produce conflicting gradients on shared parameters: the RGB loss ties geometry to view-dependent specular evaluation, while the clay loss prefers geometry that explains diffuse shading. We resolve this conflict by selectively routing gradients. Rotation $\vt_i$ and scale $\vs_i$ receive both gradients at full strength, since surface normals (derived from rotation) determine reflection directions and must remain refinable by the BRDF branch. Position $\vp_i$ and opacity $\alpha_i$, in contrast, are optimized primarily by $\mathcal{L}_{\text{clay}}$ with only a weak residual gradient from $\mathcal{L}_{\text{rgb}}$:
\begin{equation}
    \tilde{\nabla}_\theta = \nabla_\theta \mathcal{L}_{\text{clay}} + \beta \, \nabla_\theta \mathcal{L}_{\text{rgb}}, \quad \theta \in \{\vp_i, \alpha_i\}, \quad \beta = 0.1,
\end{equation}
implemented via a stop-gradient with a residual scale on the RGB path.

The total training loss combines the MaterialRefGS baseline loss $\mathcal{L}_{\text{base}}$ (\cref{sec:refl_gs}) with our clay supervision as follows:
\begin{equation}
    \mathcal{L}_{\text{total}} = \mathcal{L}_{\text{base}} + \lambda_{\text{clay}}\, \mathcal{L}_{\text{clay}}.
\end{equation}
We use $\lambda_{\text{clay}} = 1.0$ in all experiments. After the initial phase in which geometry is learned through clay loss with gradient routing, we disable the clay branch entirely and continue training with the RGB loss only for synthetic data. For real data, we instead keep both losses active, jointly optimizing with RGB and clay losses without further gradient routing. This is because reflections in real scenes are more complex and harder to disentangle, making continued reflection-free supervision beneficial.
\begin{table*}[t]
\scriptsize
\centering
\caption{Chamfer-$\mathcal{L}_1$ ($\downarrow$) distances of 3D reconstruction results on the Glossy Synthetic dataset. The intensity of the red color signifies a better result.}
\begin{tabular}{@{}lccccccccc}
\toprule[1pt]
\textbf{Method} & \textbf{Angel} & \textbf{Bell} & \textbf{Cat} & \textbf{Horse} & \textbf{Luyu} & \textbf{Potion} & \textbf{Tbell} & \textbf{Teapot} & \textbf{Average} \\ 
\midrule
GShader~\cite{jiang2024gaussianshader} 
    & 0.0060 & 0.0078 & 0.0175 & 0.0072 & 0.0101 & 0.0382 & 0.0308 & 0.0178 & 0.0169 \\
GS-IR~\cite{liang2024gsir}             
    & 0.0110 & 0.1097 & 0.0566 & 0.0149 & 0.0224 & 0.0593 & 0.0989 & 0.0693 & 0.0553 \\
R3DG~\cite{gao2024r3dg}                
    & 0.0090 & 0.0403 & 0.0326 & 0.0117 & 0.0151 & 0.0380 & 0.0472 & 0.0488 & 0.0303 \\
Ref-GS~\cite{zhang2025refgs}           
    & \thirdbest{0.0049} & 0.0084 & 0.0170 & 0.0060 & 0.0098 & 0.0090 & \thirdbest{0.0064} & 0.0084 & 0.0087 \\
GS-2DGS~\cite{tong2025gs}              
    & 0.0061 & \first{0.0037} & \secondbest{0.0074} & \thirdbest{0.0050} & \thirdbest{0.0076} & 0.0088 & 0.0078 & \thirdbest{0.0083} & \thirdbest{0.0068} \\
\reflectgs{}~\cite{yao2025reflective}  
    & 0.0050 & \thirdbest{0.0071} & 0.0161 & 0.0069 & 0.0093 & \thirdbest{0.0081} & 0.0065 & 0.0090 & 0.0085 \\
MaterialRefGS~\cite{zhangmaterialrefgs}  
    & \secondbest{0.0038} & 0.0087 & \thirdbest{0.0130} & \first{0.0047} & \secondbest{0.0061} & \secondbest{0.0067} & \secondbest{0.0049} & \secondbest{0.0059} & \secondbest{0.0067} \\
Ours                                    
    & \first{0.0035} & \secondbest{0.0069} & \first{0.0057} & \first{0.0047} & \first{0.0057} & \first{0.0064} & \first{0.0046} & \first{0.0057} & \first{0.0054} \\
\bottomrule[1pt]
\end{tabular}
\label{tab:exp_geometry_grouped}
\end{table*}

\section{Experiments}

In this section, we compare our method with existing approaches both quantitatively and qualitatively to demonstrate its effectiveness. Our method shows notably superior performance in mesh quality, confirming the effectiveness of utilizing clay images for geometry learning.

\subsection{Implementation details}
\label{sec:imple_details}
For our multi-view image-to-clay translation model, we adapt LTX-2~\cite{hacohen2026ltx} with LoRA~\cite{hu2022lora} modules of rank $32$ and train for $50{,}000$ iterations on $8 \times$ NVIDIA H100 GPUs with a batch size of $1$ per GPU. Each sample consists of $24$ multi-view images of a single object. To enable the anchor-based inference, with probability $0.5$ we expose four randomly selected target views as clean anchors during training. Our reconstruction model is built on the official codebase of MaterialRefGS~\cite{zhangmaterialrefgs}. Following the baseline, we estimate monocular normal maps using Metric3Dv2~\cite{hu2024metric3d} for the normal prior loss. Additional implementation details are provided in \cref{appx:imple_details}.

\subsection{Datasets and metrics}
We use multiple datasets widely adopted for reflective object reconstruction. 
For synthetic evaluation, we use Shiny Blender~\cite{verbin2024refnerf} and 
Glossy Synthetic~\cite{liu2023nero}, both of which render non-Lambertian 
objects under environment maps. For real-world evaluation, we use 
Ref-Real~\cite{verbin2024refnerf}, Glossy-Real~\cite{liu2023nero}, and 
NeRF-casting~\cite{verbin2024nerfcasting}, which capture diverse reflective 
objects in real scenes. Ground-truth meshes are available for the Glossy Synthetic dataset, and ground-truth surface normals are 
additionally provided for Shiny Blender. We assess mesh quality using the 
Chamfer distance computed with the official evaluation code, and surface 
normal accuracy using the mean angular error (MAE). The photometric quality 
of novel-view synthesis is also evaluated by comparing rendered results with 
ground-truth images using PSNR, SSIM, and LPIPS.

\subsection{Baselines}
\label{sec:baselines}

We compare our method with various state-of-the-art Gaussian-based approaches 
for reflective object reconstruction, including 2DGS~\cite{huang20242d}, 
GShader~\cite{jiang2024gaussianshader}, R3DG~\cite{gao2024r3dg}, 3DGS-DR~\cite{ye2024gsdr}, 
Ref-GS~\cite{zhang2025refgs}, RGS~\cite{yao2025reflective}, and 
MaterialRefGS~\cite{zhangmaterialrefgs}. We use the official implementation of the baselines. Note that the reflection prior in the MaterialRefGS code is not yet available, so this component is excluded from our experiments. This may cause a discrepancy between the results reported in their paper and our reproduced results. Among these, MaterialRefGS serves as 
the underlying reflective Gaussian splatting backbone of our method 
(\cref{sec:refl_gs}), upon which we add the proposed clay-guided supervision.

\begin{table*}[t]
\scriptsize
\centering
\caption{Generalization across backbones and the effect of normal prior. We report Chamfer-$\mathcal{L}_1$ ($\downarrow$) distances on the Glossy Synthetic dataset. \textit{``+ Clay''} integrates our clay-guided branch on top of each backbone.  \textit{``- Normal''} further removes the monocular normal prior to test whether clay supervision alone suffices. The intensity of the red color signifies a better result.}
\begin{tabular}{@{}lccccccccc}
\toprule[1pt]
\textbf{Method} & \textbf{Angel} & \textbf{Bell} & \textbf{Cat} & \textbf{Horse} & \textbf{Luyu} & \textbf{Potion} & \textbf{Tbell} & \textbf{Teapot} & \textbf{Average} \\ 
\midrule
\reflectgs{}~\cite{yao2025reflective}  
    & 0.0050 & 0.0071 & 0.0161 & 0.0069 & 0.0093 & 0.0081 & 0.0065 & 0.0090 & 0.0085 \\
\reflectgs{} \textit{+ Clay}
    & \first{0.0032} & \secondbest{0.0067} & 0.0142 & \first{0.0042} & 0.0066 & 0.0077 & 0.0061 & 0.0110 & 0.0074 \\
MaterialRefGS~\cite{zhangmaterialrefgs}  
    & \thirdbest{0.0038} & 0.0087 & \thirdbest{0.0130} & \thirdbest{0.0047} & \secondbest{0.0061} & \thirdbest{0.0067} & \thirdbest{0.0049} & \secondbest{0.0059} & \thirdbest{0.0067} \\
MaterialRefGS \textit{+ Clay}                                    
    & \secondbest{0.0035} & \thirdbest{0.0069} & \first{0.0057} & \thirdbest{0.0047} & \first{0.0057} & \first{0.0064} & \secondbest{0.0046} & \first{0.0057} & \first{0.0054} \\
MaterialRefGS \textit{+ Clay - Normal}
    & \secondbest{0.0035} & \first{0.0047} & \secondbest{0.0087} & \secondbest{0.0044} & \thirdbest{0.0063} & \first{0.0064} & \first{0.0034} & \thirdbest{0.0060} & \first{0.0054} \\
\bottomrule[1pt]
\end{tabular}
\label{tab:exp_ablation_normal_backbone}
\end{table*}

\begin{table*}[t]
\centering
\small
\caption{\textbf{Average image quality on five benchmark datasets.} For each dataset, we report the average PSNR / SSIM / LPIPS over all scenes. Per-scene scores are reported in \cref{tab:main_compare_synthetic} and \cref{tab:main_compare_real}. The intensity of the red color signifies a better result.}
\label{tab:main_compare_avg}
\resizebox{\textwidth}{!}{%
\begin{tabular}{l|ccc|ccc|ccc|ccc|ccc}
\toprule[1pt]
\textbf{Datasets} 
         & \multicolumn{3}{c|}{\textbf{Shiny Blender}~\citep{verbin2024refnerf}} 
         & \multicolumn{3}{c|}{\textbf{Glossy Synthetic}~\citep{liu2023nero}} 
         & \multicolumn{3}{c|}{\textbf{Ref-Real}~\citep{verbin2024refnerf}} 
         & \multicolumn{3}{c|}{\textbf{Glossy-Real}~\citep{liu2023nero}} 
         & \multicolumn{3}{c}{\textbf{NeRF-casting}~\citep{verbin2024nerfcasting}} \\
\textbf{Metric}   
         & PSNR$\uparrow$ & SSIM$\uparrow$ & LPIPS$\downarrow$
         & PSNR$\uparrow$ & SSIM$\uparrow$ & LPIPS$\downarrow$
         & PSNR$\uparrow$ & SSIM$\uparrow$ & LPIPS$\downarrow$
         & PSNR$\uparrow$ & SSIM$\uparrow$ & LPIPS$\downarrow$
         & PSNR$\uparrow$ & SSIM$\uparrow$ & LPIPS$\downarrow$ \\
\midrule
2DGS          & 29.58 & 0.946 & 0.084 & 26.07 & 0.918 & 0.088 & 24.15 & 0.661 & 0.292 & 21.20 & 0.804 & 0.243 & 24.66 & 0.822 & 0.225 \\
Ref-GS        & \thirdbest{34.80} & \thirdbest{0.973} & \thirdbest{0.056} & 30.16 & 0.956 & 0.055 & 24.44 & 0.682 & \thirdbest{0.225} & \thirdbest{22.33} & \thirdbest{0.814} & \thirdbest{0.228} & \thirdbest{28.09} & \first{0.855} & \thirdbest{0.185} \\
\reflectgs{}  & \secondbest{35.04} & \thirdbest{0.973} & \thirdbest{0.056} & \thirdbest{30.37} & \thirdbest{0.958} & \thirdbest{0.049} & \thirdbest{24.61} & \thirdbest{0.685} & 0.252 & 22.00 & 0.802 & 0.254 & \secondbest{28.35} & \thirdbest{0.849} & 0.211 \\
MaterialRefGS & 34.68 & \secondbest{0.975} & \secondbest{0.054} & \secondbest{30.60} & \secondbest{0.960} & \secondbest{0.048} & \secondbest{24.73} & \secondbest{0.698} & \first{0.183} & \secondbest{22.65} & \secondbest{0.818} & \secondbest{0.186} & 28.07 & 0.845 & \secondbest{0.148} \\
Ours          & \first{35.12} & \first{0.976} & \first{0.053} & \first{30.68} & \first{0.961} & \first{0.047} & \first{24.74} & \first{0.699} & \first{0.183} & \first{23.04} & \first{0.824} & \first{0.184} & \first{28.69} & \secondbest{0.853} & \first{0.137} \\
\bottomrule[1pt]
\end{tabular}%
}
\end{table*}
\subsection{Comparisons}

We first evaluate the geometric quality of reconstructed meshes, the primary focus of our method. \cref{tab:exp_geometry_grouped} reports Chamfer-$\mathcal{L}_1$ distances on Glossy Synthetic, where our method achieves the lowest average distance, outperforming MaterialRefGS. On Shiny Blender (\cref{tab:normal_mae}), our method also achieves the lowest mean angular error. 

In \cref{tab:exp_ablation_normal_backbone}, we further investigate whether our clay-guided branch enhances geometry independent of the backbone, and the role of the external normal prior. Adding our clay branch to RGS~\cite{yao2025reflective} reduces the Chamfer distance from $0.0085$ to $0.0074$, with consistent improvements across most scenes. Removing the normal prior from MaterialRefGS while keeping the clay branch (\textit{+ Clay $-$ Normal}) degrades some scenes (\eg, Cat) but improves others (\eg, Bell), where highly curved surfaces lead to noisy normal predictions. Unlike normal supervision, our clay branch operates through an independent rendering path that does not affect the environment map or BRDF parameters, making it more robust to such noise, and supporting its role as both a complementary signal and a standalone source of geometric guidance. Moreover, clay-like objects stay within the pretrained RGB domain of the video backbone, unlike normal or depth maps that require learning a new output modality.

Beyond geometric metrics, we report photometric metrics on synthetic and real-world datasets in \cref{tab:main_compare_avg}. Per-scene scores are provided in \cref{tab:main_compare_synthetic,tab:main_compare_real} in the appendix. For older baselines, we report the original numbers from their papers, while results on Glossy-Real and NeRF-casting are obtained by re-running the most recent reflective splatting methods. On average, our method outperforms the baseline on both synthetic and real datasets.

\begin{figure}[t]
  \centering
  \setlength{\tabcolsep}{2pt}

  \begin{tabular}{>{\centering\arraybackslash}p{0.19\linewidth}
                  >{\centering\arraybackslash}p{0.19\linewidth}
                  >{\centering\arraybackslash}p{0.19\linewidth}
                  >{\centering\arraybackslash}p{0.19\linewidth}
                  >{\centering\arraybackslash}p{0.19\linewidth}}

    \includegraphics[width=\linewidth]{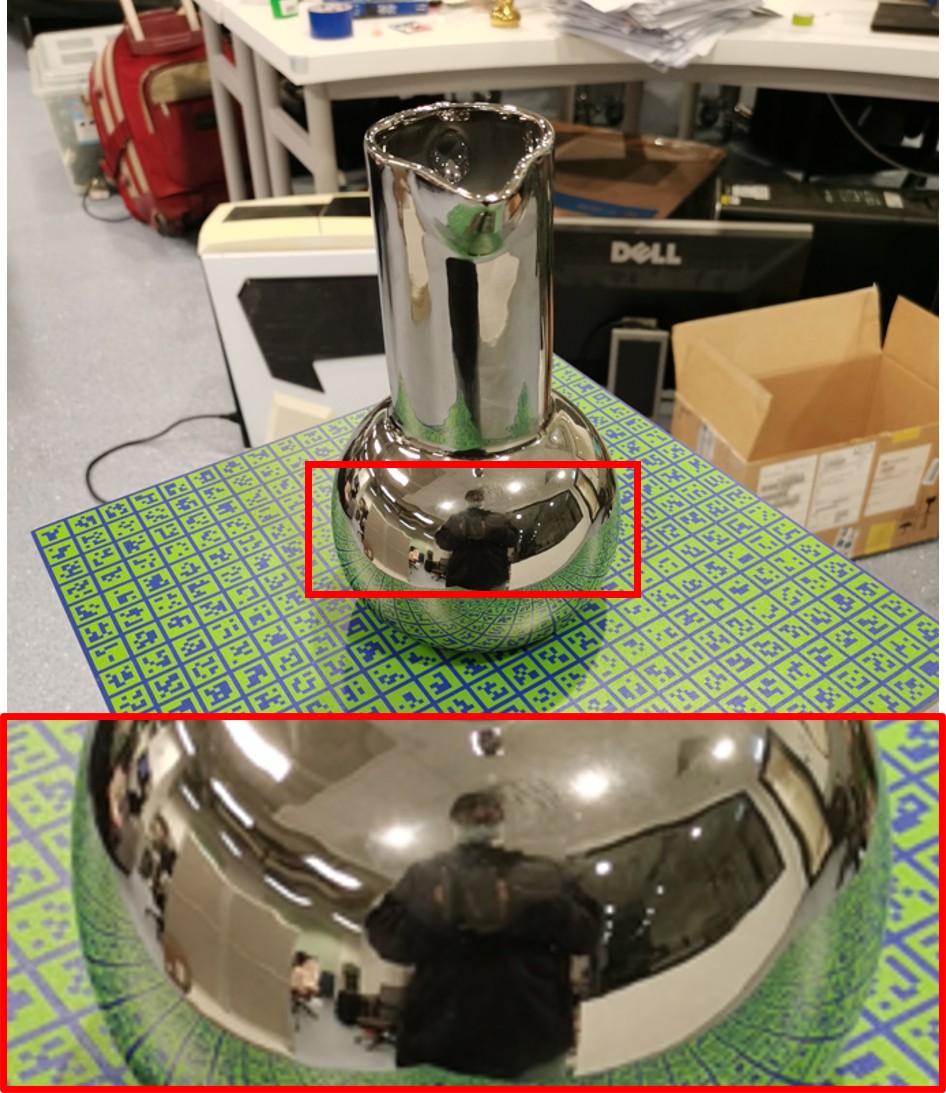} &
    \includegraphics[width=\linewidth]{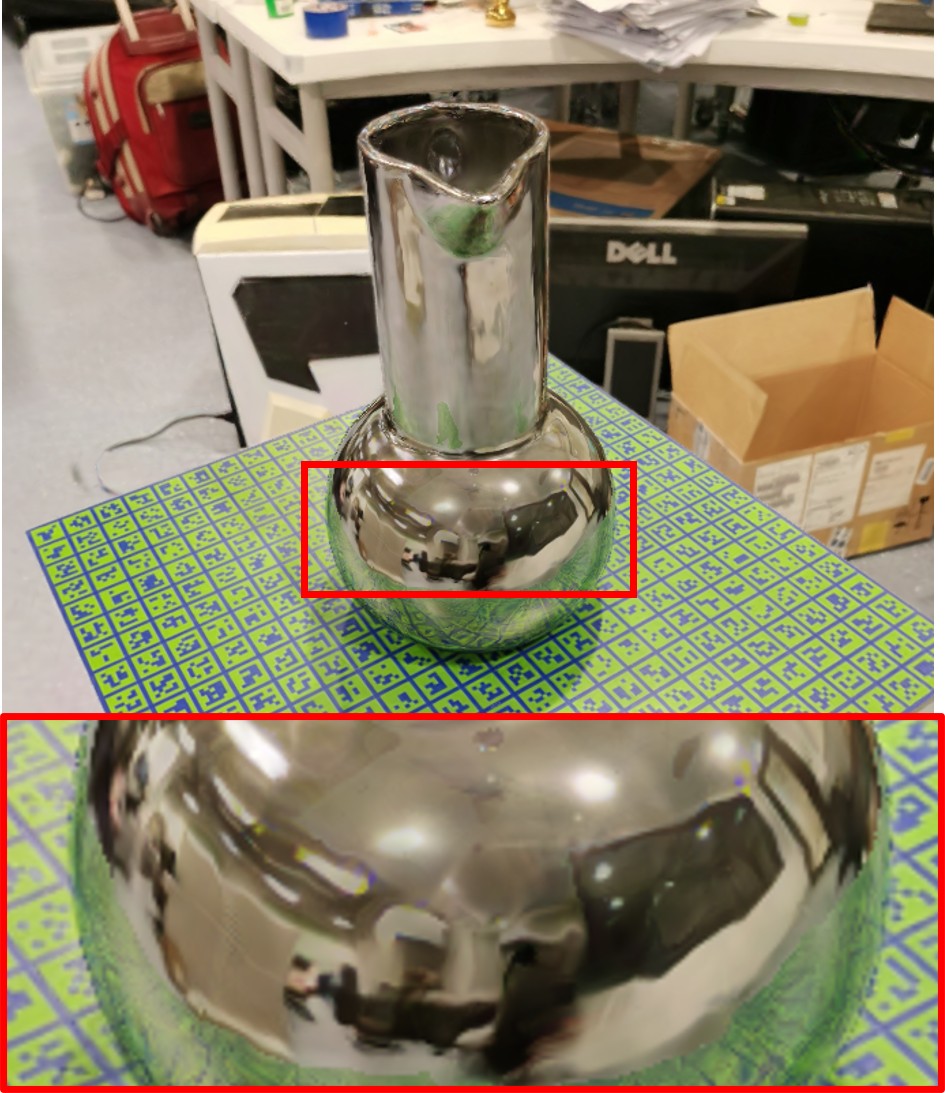} &
    \includegraphics[width=\linewidth]{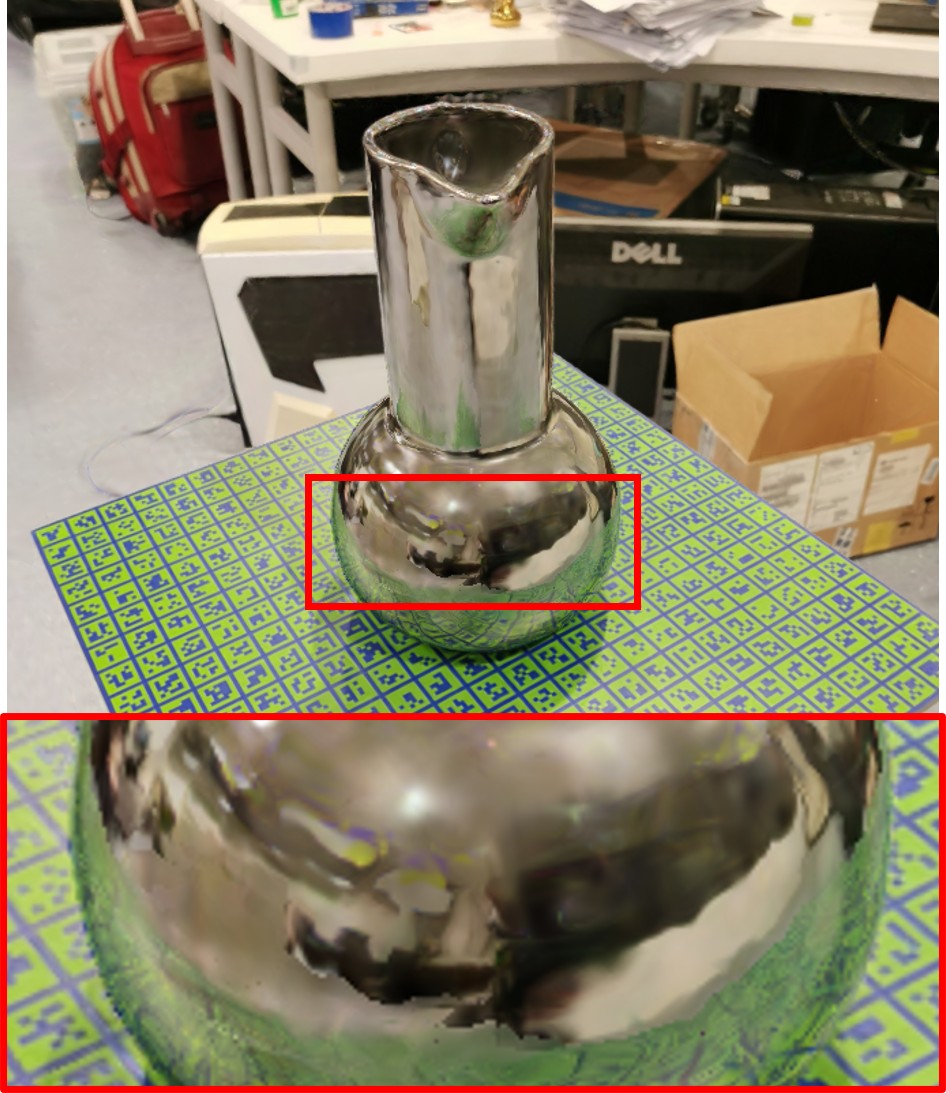} &
    \includegraphics[width=\linewidth]{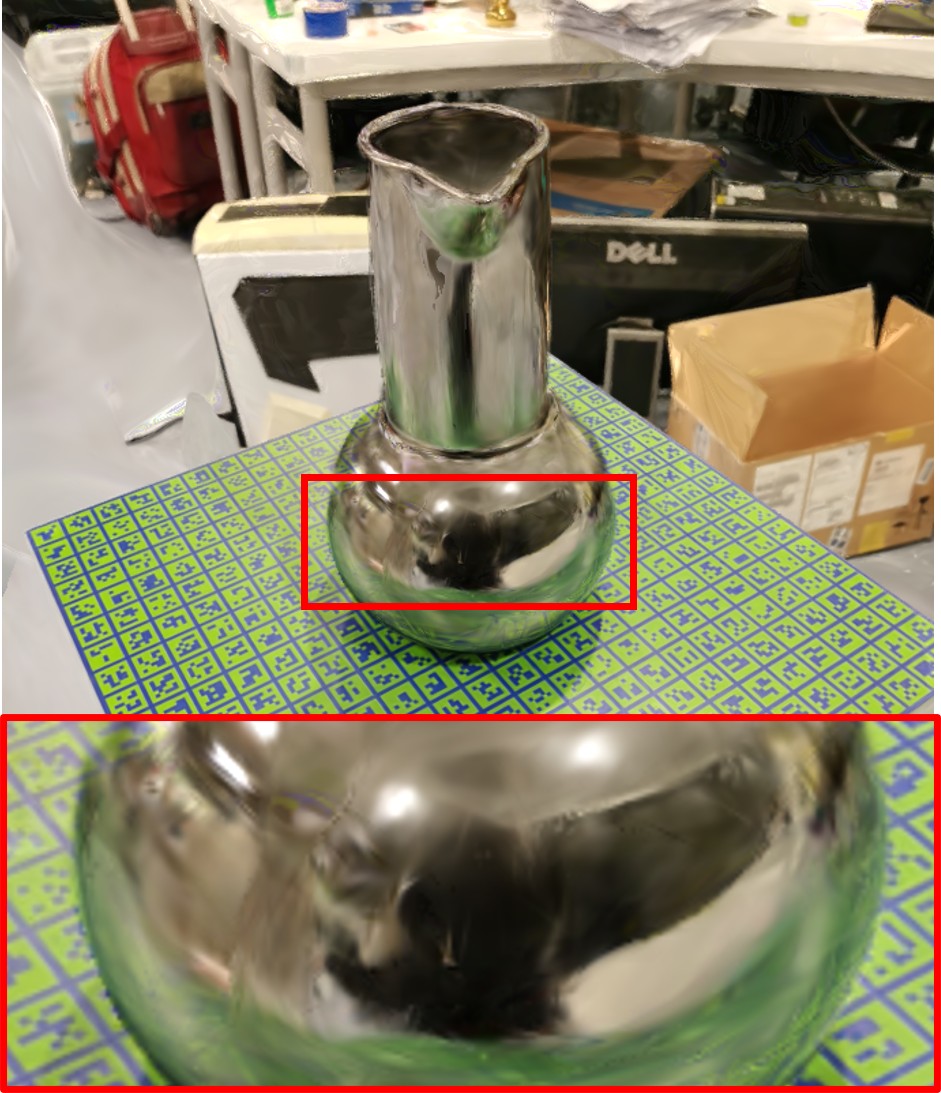} &
    \includegraphics[width=\linewidth]{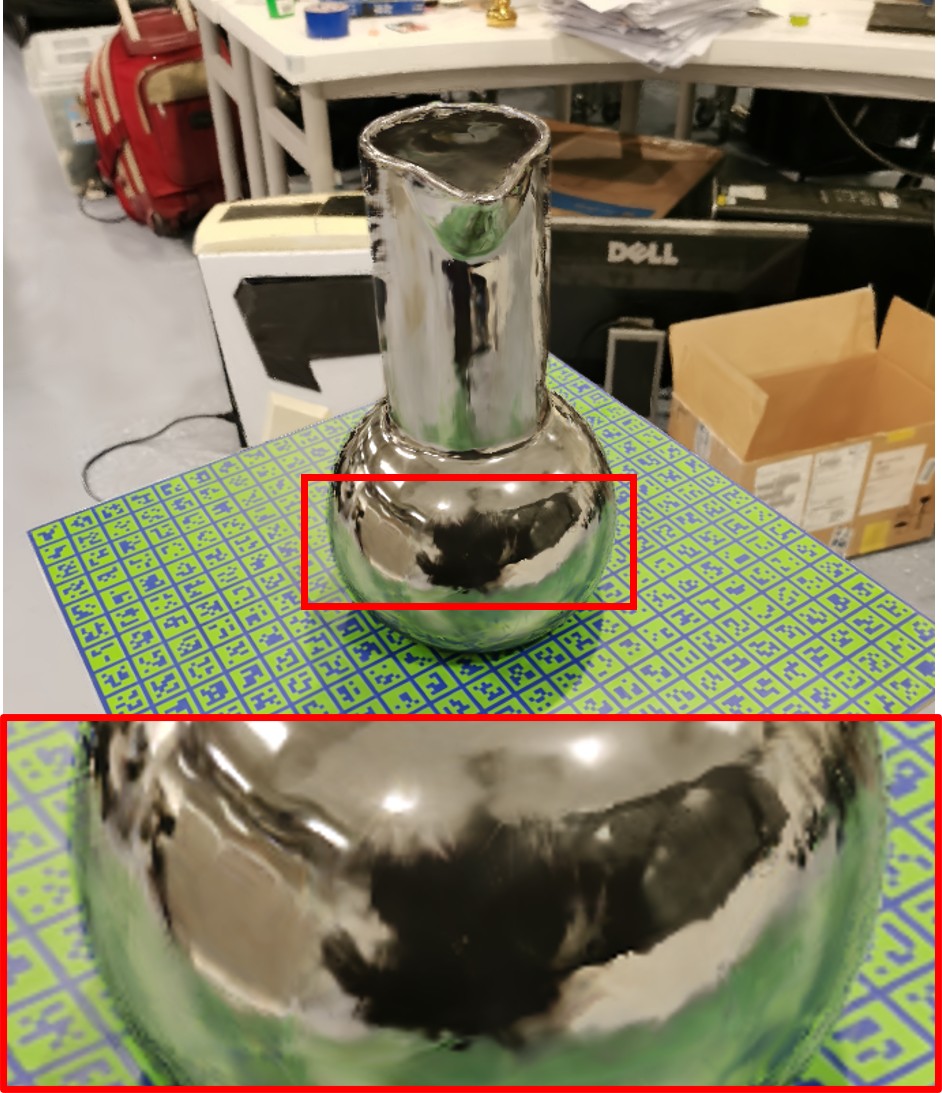} \\[2pt]
    \includegraphics[width=\linewidth]{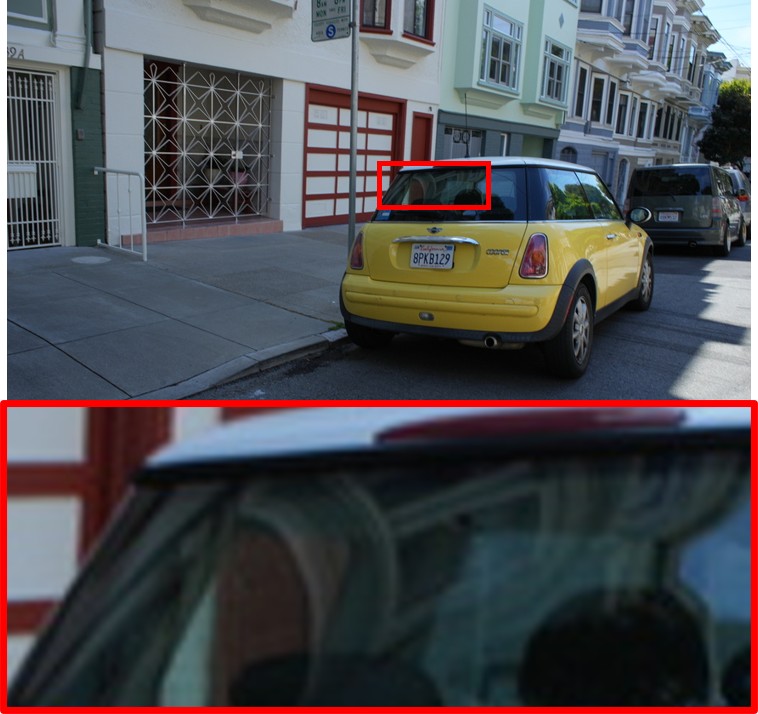} &
    \includegraphics[width=\linewidth]{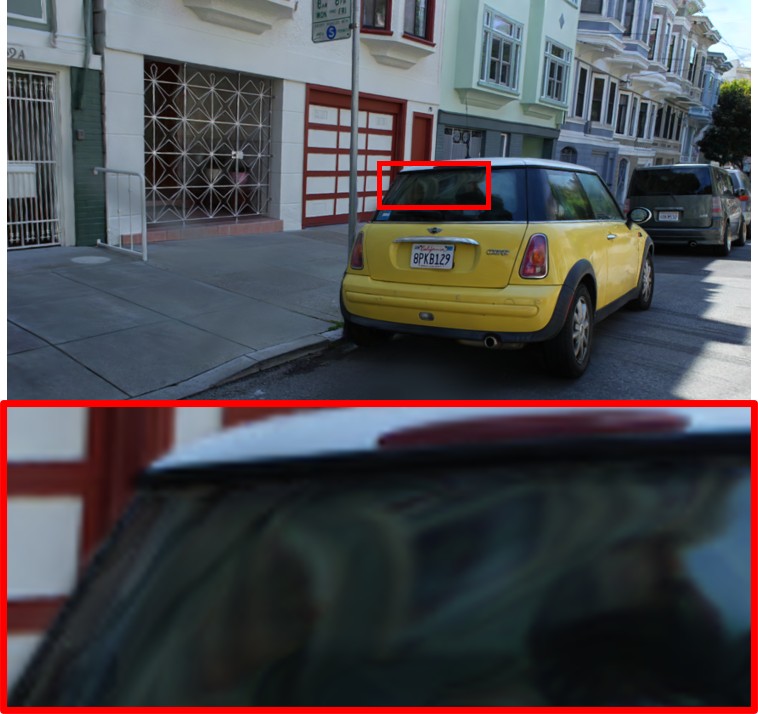} &
    \includegraphics[width=\linewidth]{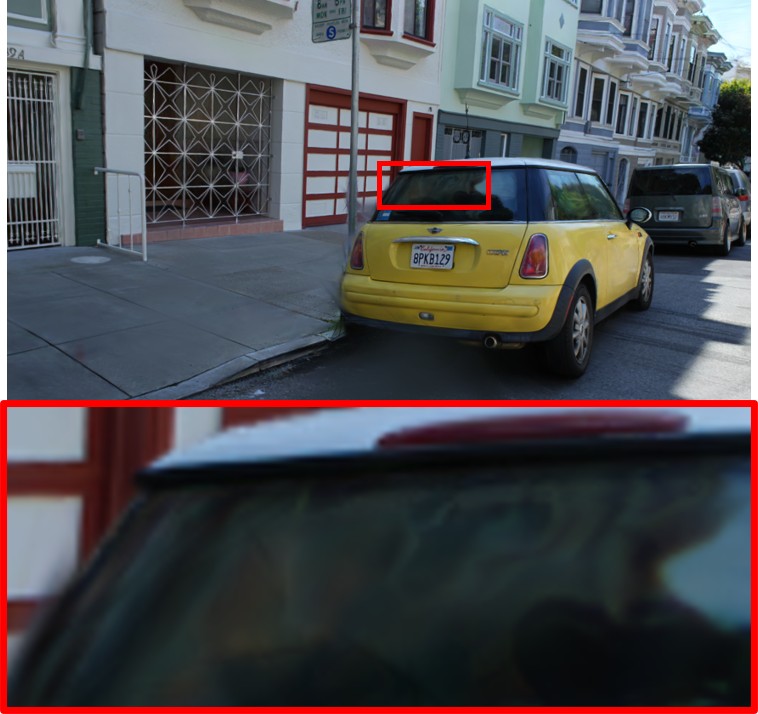} &
    \includegraphics[width=\linewidth]{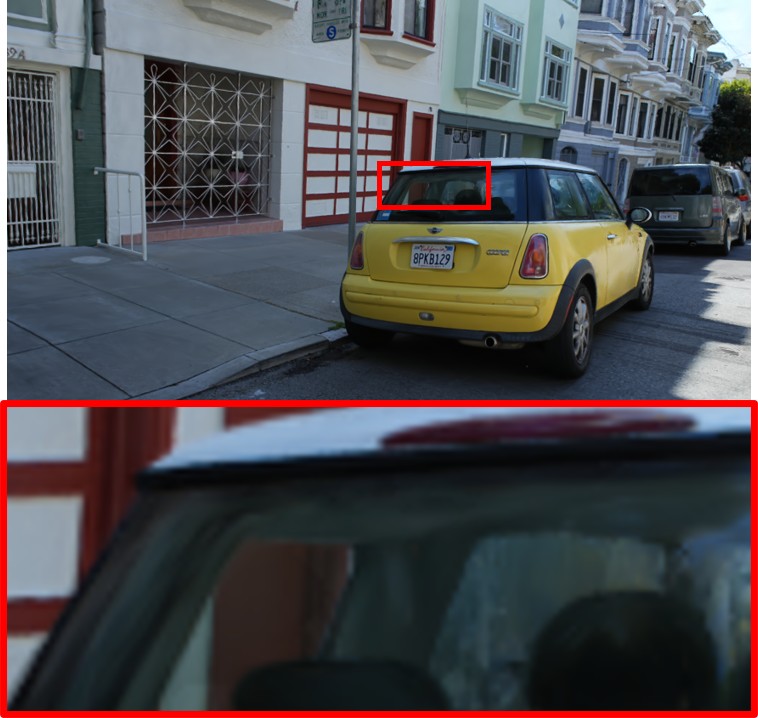} &
    \includegraphics[width=\linewidth]{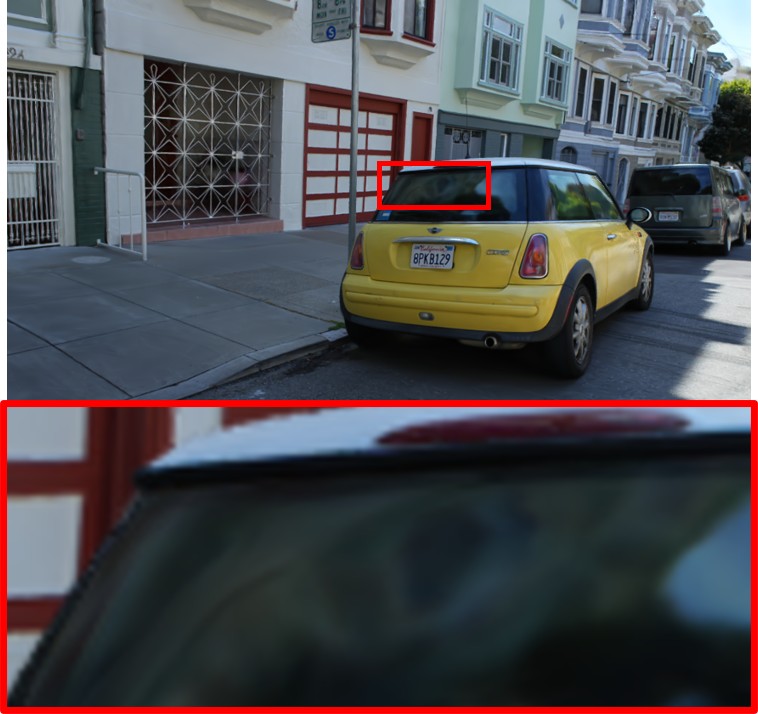} \\[2pt]

    GT & Ours & MaterialRefGS & RGS &  Ref-GS

  \end{tabular}

  \caption{Qualitative comparison of our method with baselines. Our clay-guided supervision yields cleaner geometry, which in turn produces sharper and more faithful reflections.}

  \label{fig:qualitative_rgb}
\end{figure}

\cref{fig:qualitative_rgb} shows qualitative comparisons. In the upper row, the photographer's reflection is a view-dependent artifact that should not be baked into either the environment map or the diffuse color. However, RGS and Ref-GS visibly absorb this \textit{transient} object into the diffuse component, indicating that these methods overfit diffuse color to explain view-dependent appearance. On the car window (bottom), our method clearly reproduces the reflection of the overhead building seen in the GT, whereas baselines blur or fail to capture it. Note that the transmission-like appearance of some baselines arises when the window fails to form a proper surface and is left as a hole, so the diffuse interior behind it becomes directly visible rather than being rendered through a reflective surface. Our method recovers the window surface, so the BRDF model operates on proper geometry and the reflection is correctly reproduced. Additional visualizations are provided in \cref{appx:additional_results}.

Even with monocular normal supervision, the original BRDF path requires substantial training to disentangle geometry from BRDF parameters under a view-inconsistent RGB loss. Our clay branch sidesteps this by learning geometry from view-consistent clay images without specular rendering, similar to 2DGS on non-reflective objects. 
\cref{fig:vis_convergence} shows that this facilitates the separation of specular and diffuse colors and accelerates the learning of the environment map.

\subsection{Ablation studies}
\label{sec:ablation}
\paragraph{Multi-view consistency of clay translation.}
We measure how well our multi-view image-to-clay model maps the same surface point to consistent colors across views using a color \emph{variance ratio}:
\begin{equation}
R = \frac{\mathbb{E}_\mathbf{p}\big[\mathrm{Var}_v(C_v(\mathbf{p}))\big]}{\mathrm{Var}_{\mathbf{p} \in \mathcal{I}}(C(\mathbf{p}))},
\label{eq:variance_ratio}
\end{equation}
where $C_v(\mathbf{p})$ is the color at view $v$ corresponding to surface point $\mathbf{p}$, and $\mathcal{I}$ is the set of all foreground pixels. For each pixel in an anchor view, we use the GT depth to reproject it to its $K=10$ nearest views (selected by camera-center distance) and collect the values observed at the same 3D point. The numerator measures the \emph{cross-view variance} at corresponding surface points, while the denominator measures the \emph{global variance} across all foreground pixels. A low $R$ thus indicates that a method produces expressive outputs overall while keeping the same surface point consistent across views.

\begin{figure*}[t]
\centering
\scriptsize
\setlength{\tabcolsep}{1pt}
\renewcommand{\arraystretch}{1.0}
\begin{tabular}{@{}
    >{\centering\arraybackslash}p{0.115\linewidth}
    >{\centering\arraybackslash}p{0.115\linewidth}@{\hspace{8pt}}
    >{\centering\arraybackslash}p{0.115\linewidth}
    >{\centering\arraybackslash}p{0.115\linewidth}@{\hspace{8pt}}
    >{\centering\arraybackslash}p{0.115\linewidth}
    >{\centering\arraybackslash}p{0.115\linewidth}@{\hspace{8pt}}
    >{\centering\arraybackslash}p{0.115\linewidth}
    >{\centering\arraybackslash}p{0.115\linewidth}@{}}
    \includegraphics[width=\linewidth]{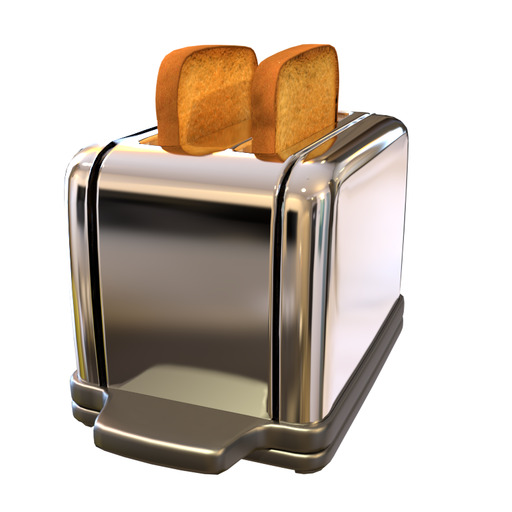} &
    \includegraphics[width=\linewidth]{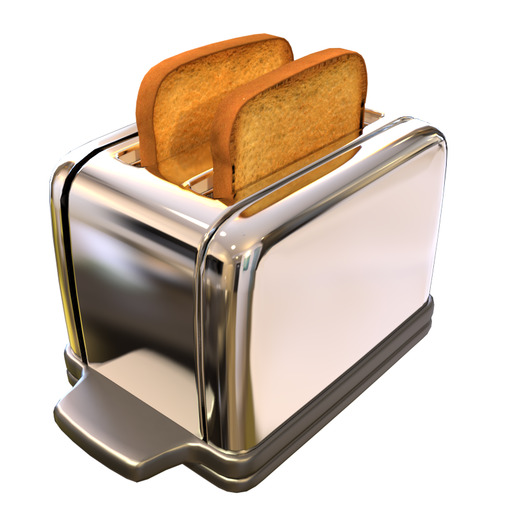} &
    \includegraphics[width=\linewidth]{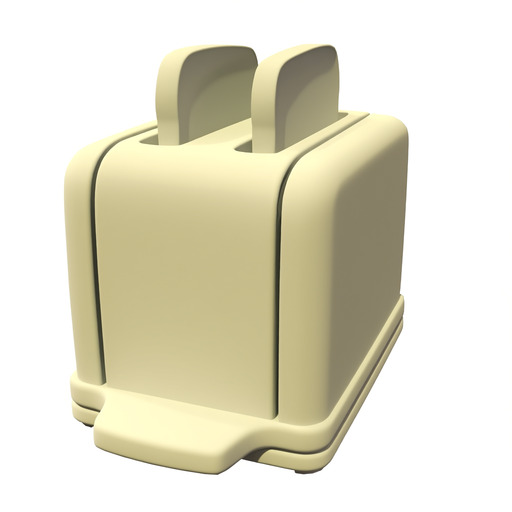} &
    \includegraphics[width=\linewidth]{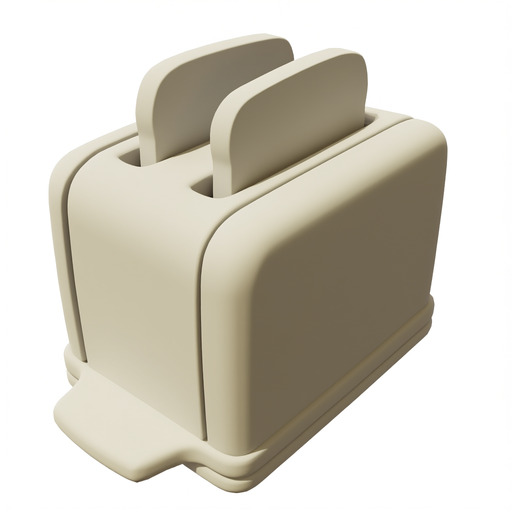} &
    \includegraphics[width=\linewidth]{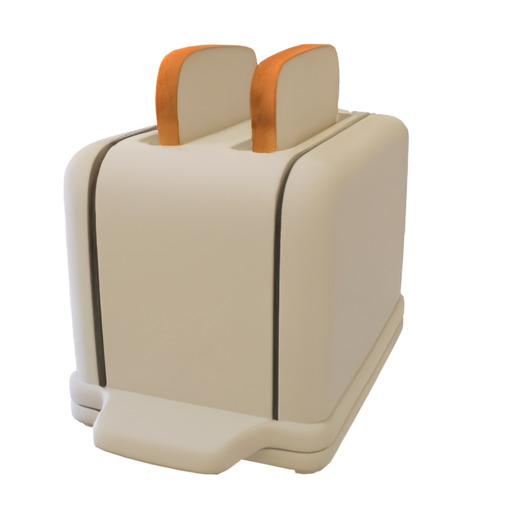} &
    \includegraphics[width=\linewidth]{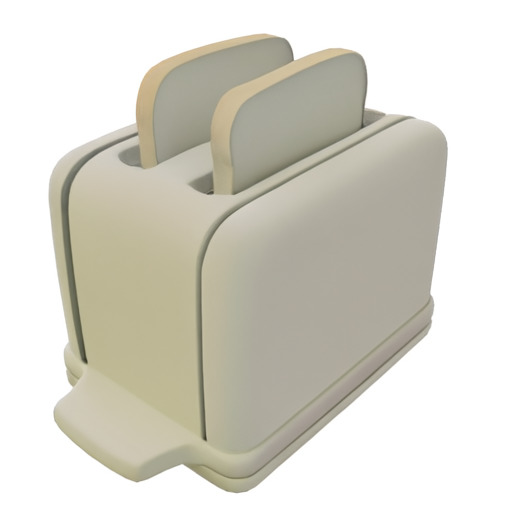} &
    \includegraphics[width=\linewidth]{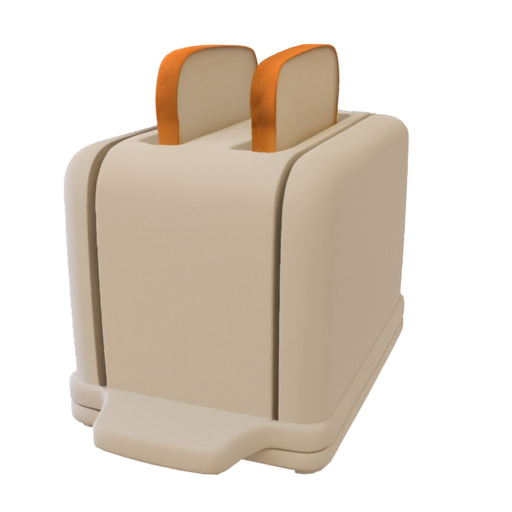} &
    \includegraphics[width=\linewidth]{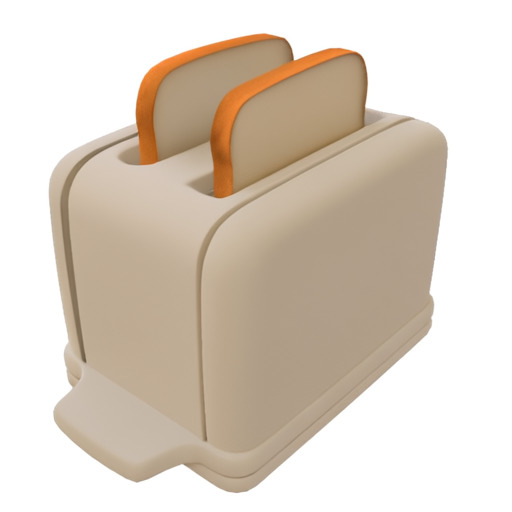} \\[1pt]
    \multicolumn{2}{c}{RGB} &
    \multicolumn{2}{c}{Qwen-Edit} &
    \multicolumn{2}{c}{LTX-2 (no anchor)} &
    \multicolumn{2}{c}{LTX-2 (Ours)}
\end{tabular}
\caption{\textbf{Multi-view image-to-clay translation comparison.}
Qwen-Image-Edit~\cite{wu2025qwen} and LTX-2~\cite{hacohen2026ltx} without
anchors often produce color-inconsistent translations across different views.}
\label{fig:dataset_and_consistency}
\end{figure*}

\begin{table*}[t]
\centering
\scriptsize
\caption{\textbf{Multi-view consistency of clay translation and its downstream
effect.} Left: consistency of translated clay images on Shiny Blender; lower
ratio $R$ indicates more view-consistent supervision relative to its
expressiveness. Right: reconstruction obtained by running the full pipeline
with each translator's clay images, keeping everything else fixed. Best is
bolded.}
\label{tab:translation_consistency}
\setlength{\tabcolsep}{6pt}
\begin{tabular}{@{}lccc|cccc@{}}
\toprule[1pt]
 & \multicolumn{3}{c}{Translation consistency} &
   \multicolumn{4}{c}{Downstream reconstruction} \\
\cmidrule(lr){2-4} \cmidrule(l){5-8}
Method & Cross-view Var $\downarrow$
       & Global Var $\uparrow$
       & Ratio $R$ $\downarrow$
       & MAE $\downarrow$ & PSNR $\uparrow$ & SSIM $\uparrow$ & LPIPS $\downarrow$ \\
\midrule
Qwen-Edit~\cite{wu2025qwen} & 0.0214 & \textbf{0.0555} & 0.3875
    & 1.53 & 34.60 & 0.9732 & 0.057 \\
LTX-2 (sorted)              & 0.0181 & 0.0455 & 0.3966
    & 1.39 & 34.78 & 0.9749 & 0.054 \\
LTX-2 (no anchor)           & \textbf{0.0157} & 0.0398 & 0.4299
    & 1.33 & 34.94 & 0.9750 & 0.054 \\
LTX-2 (\textbf{ours})       & 0.0163 & 0.0436 & \textbf{0.3661}
    & \textbf{1.27} & \textbf{35.12} & \textbf{0.9760} & \textbf{0.053} \\
\bottomrule[1pt]
\end{tabular}
\end{table*}

\begin{figure}[t]
  \centering
  \includegraphics[width=0.163\linewidth]{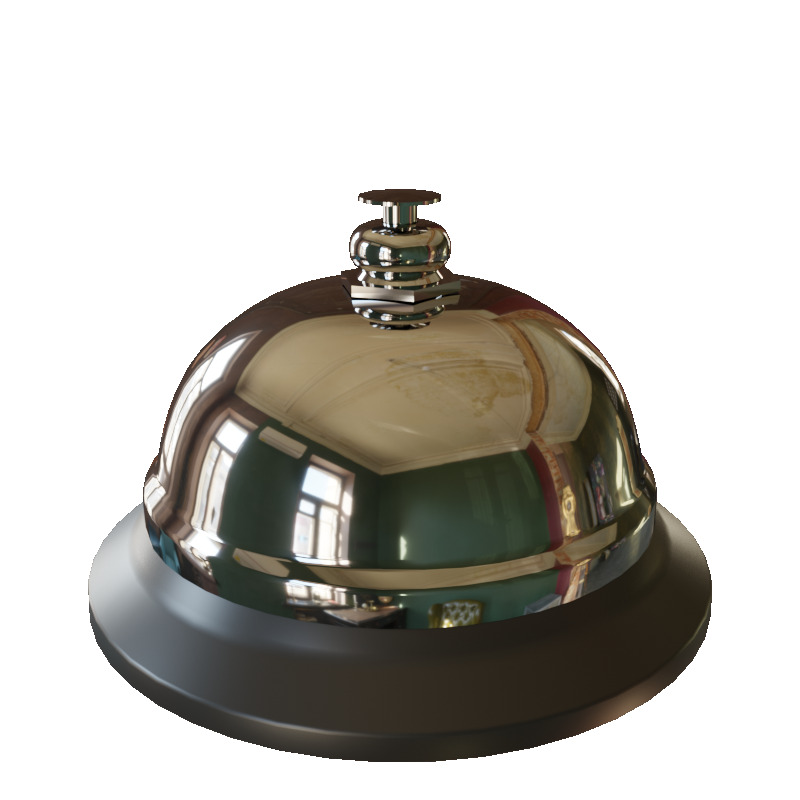}\hfill
  \includegraphics[width=0.163\linewidth]{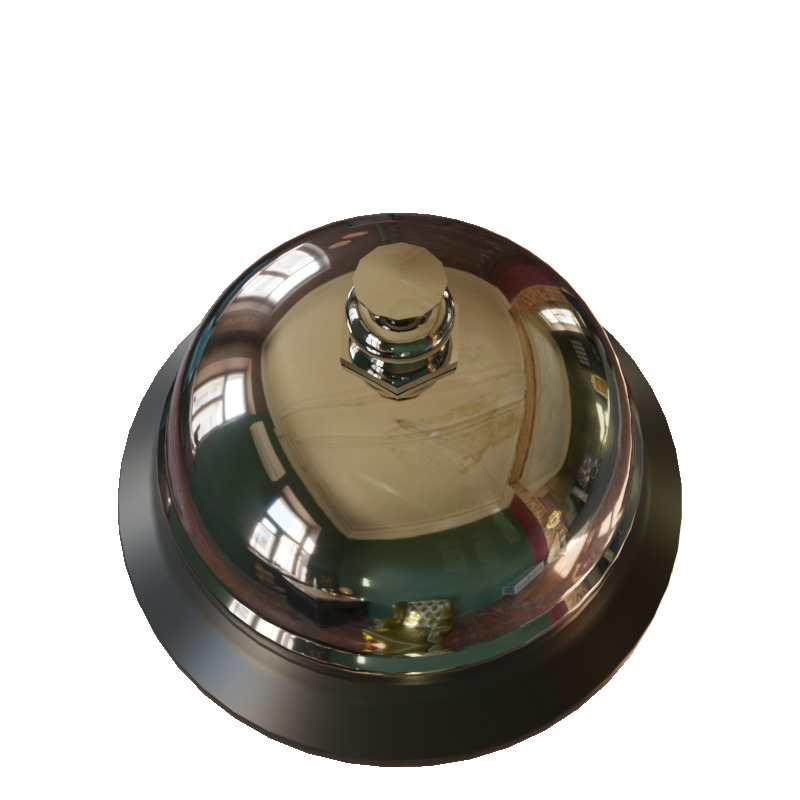}\hfill
  \includegraphics[width=0.163\linewidth]{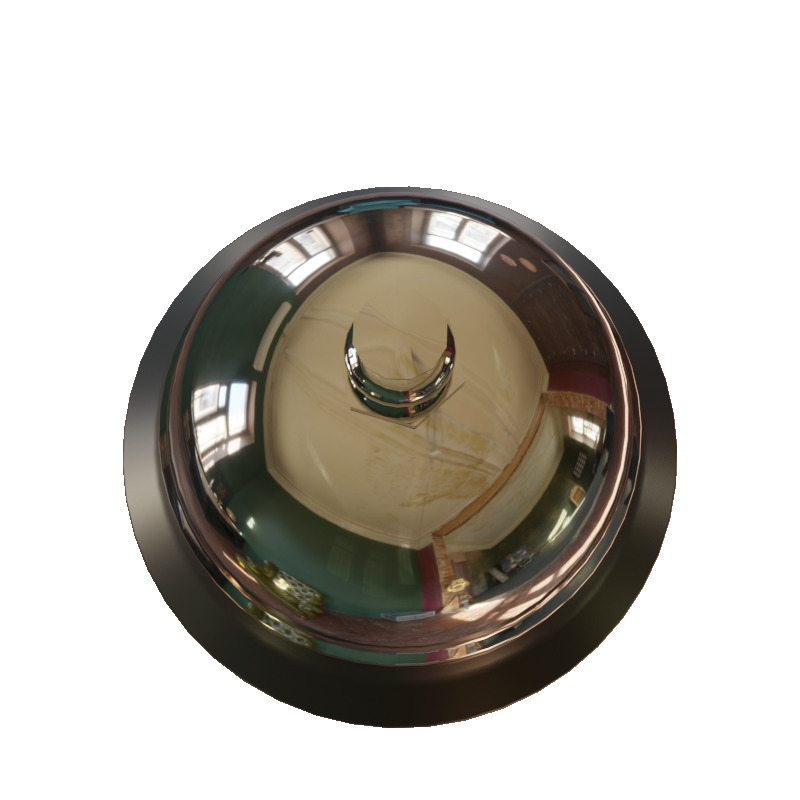}\hfill
  \includegraphics[width=0.163\linewidth]{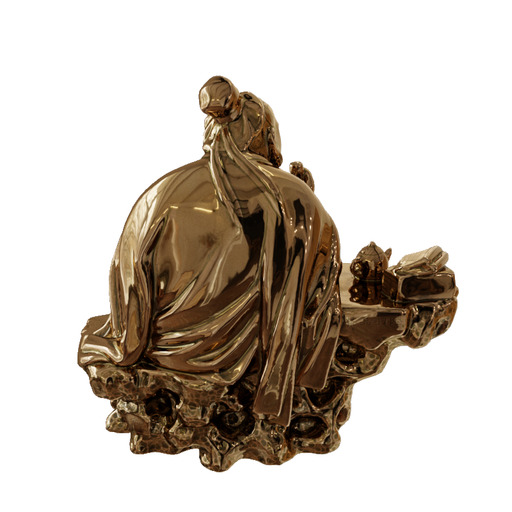}\hfill
  \includegraphics[width=0.163\linewidth]{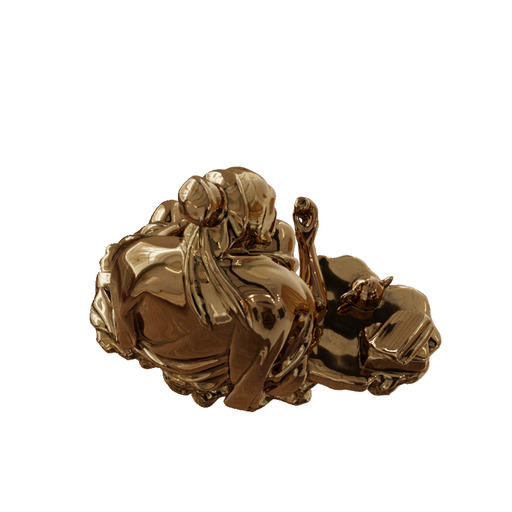}\hfill
  \includegraphics[width=0.163\linewidth]{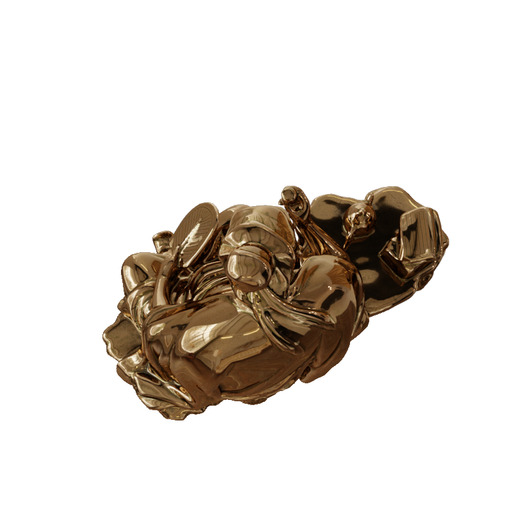}\\[1pt]
  \includegraphics[width=0.163\linewidth]{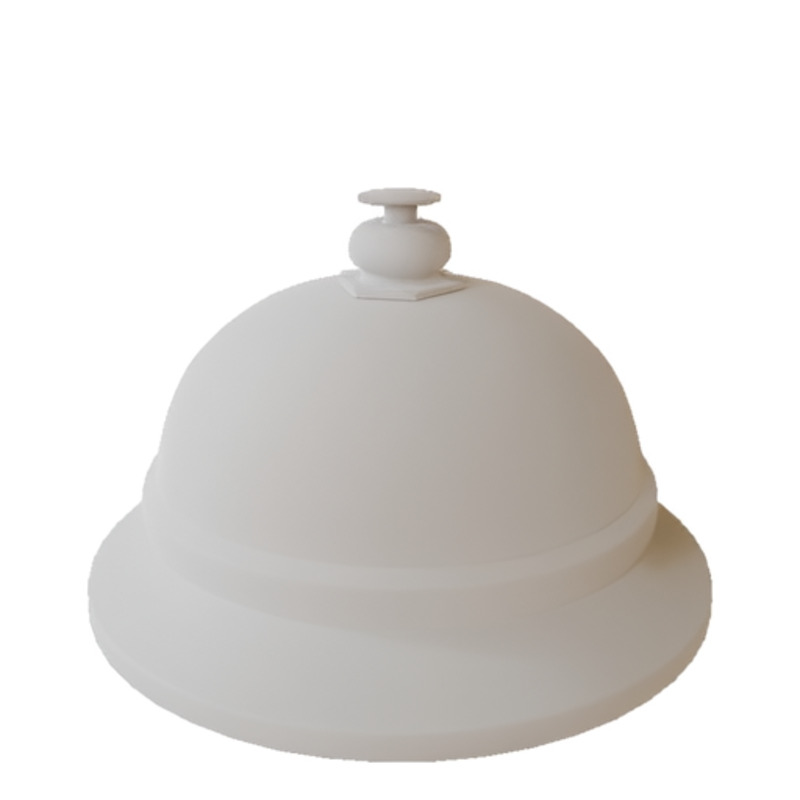}\hfill
  \includegraphics[width=0.163\linewidth]{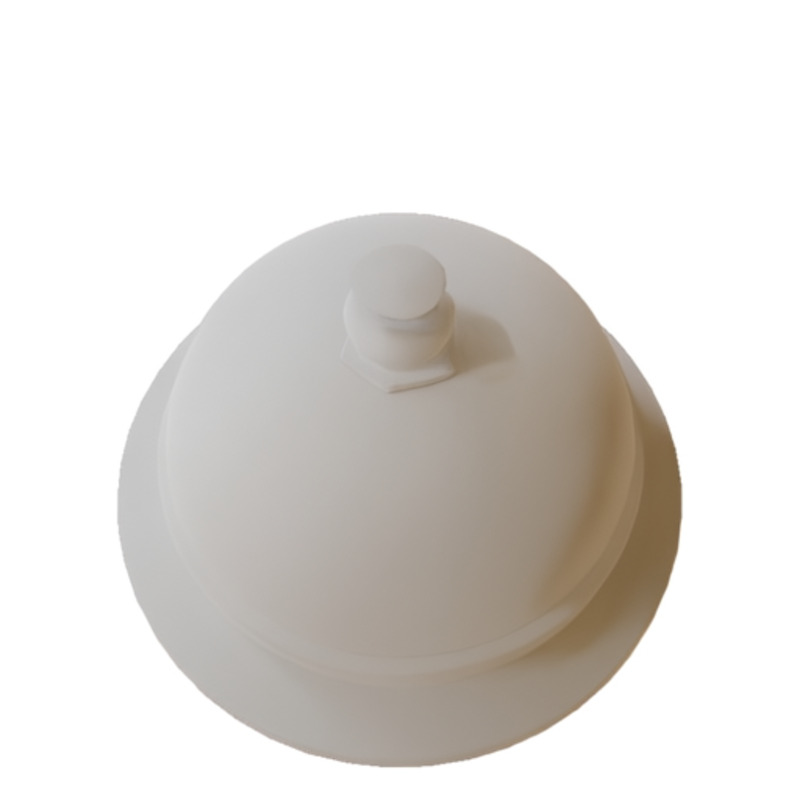}\hfill
  \includegraphics[width=0.163\linewidth]{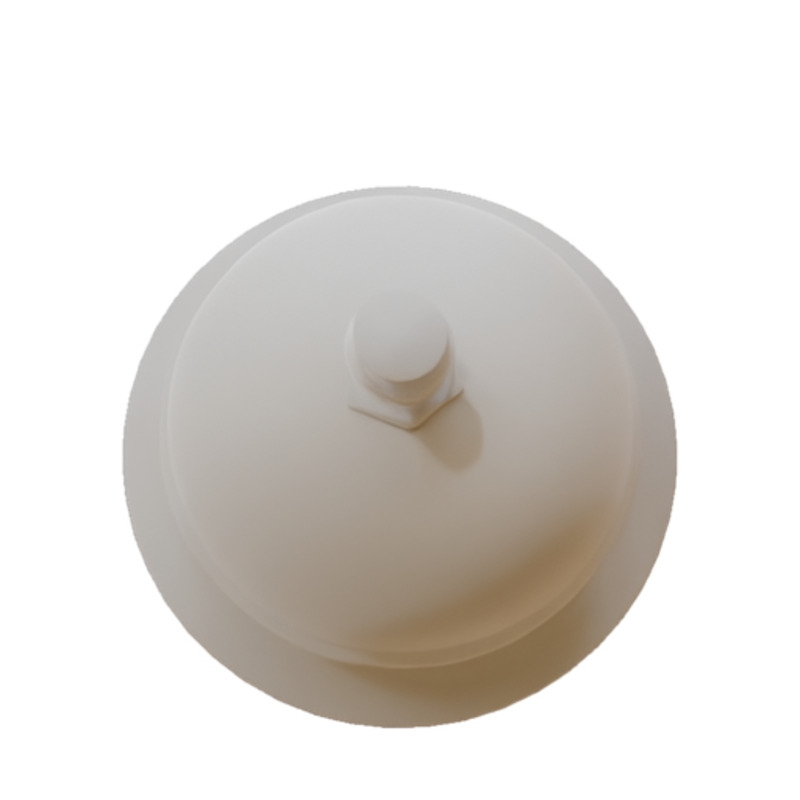}\hfill
  \includegraphics[width=0.163\linewidth]{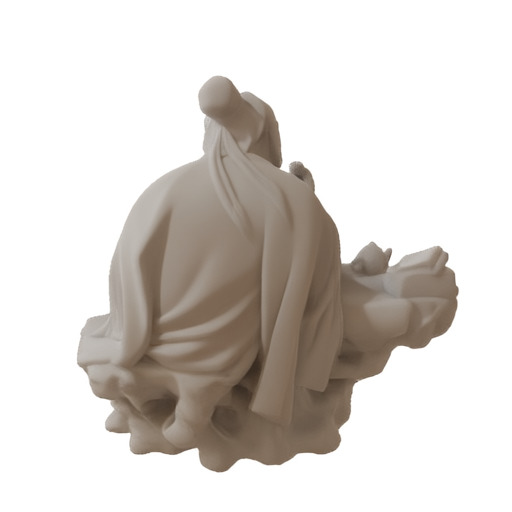}\hfill
  \includegraphics[width=0.163\linewidth]{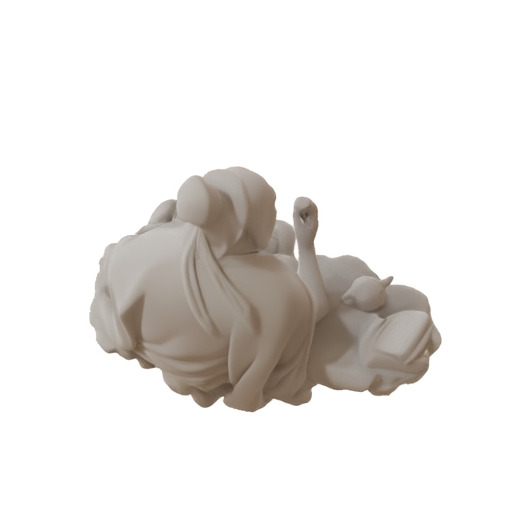}\hfill
  \includegraphics[width=0.163\linewidth]{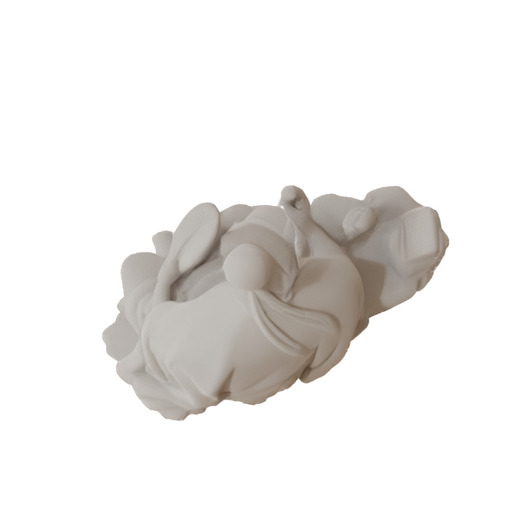}
  \caption{\textbf{Multi-view image-to-clay translation results on synthetic
  scenes.} Top: input RGB views from Glossy Synthetic; bottom: translated clay
  images. Regions reflecting the same environment lighting are translated into
  consistent tones across views. Real-world results are provided in
  \cref{fig:clayoutput}.}
  \label{fig:clayoutput_syn}
\end{figure}

\cref{tab:translation_consistency} compares per-image translation with Qwen-Image-Edit~\cite{wu2025qwen} and our LTX-2-based translator under three view-arrangement strategies. Qwen-Image-Edit yields strong shading contrast (high global variance) but lacks any mechanism for cross-view coherence, producing visibly inconsistent colors across views (\cref{fig:dataset_and_consistency}). LTX-2 without anchors (\cref{anchor}) stays consistent within each generation group but drifts across groups, implicitly suppressing overall variance to compensate. LTX-2 with sorted ordering groups views by camera proximity, so each group only sees a partial portion of the object, raising cross-view variance at points observed from distant groups. Our random-input strategy with anchor conditioning achieves the lowest $R$, indicating the most view-consistent translation. \cref{fig:clayoutput_syn} visualizes our translation on synthetic scenes, where regions reflecting the same environment lighting are mapped to consistent tones across views; real-world results are shown in \cref{fig:clayoutput}.

The rightmost columns of \cref{tab:translation_consistency} further link this
consistency to final reconstruction quality. Our translator achieves the lowest variance ratio and the best downstream reconstruction metrics among the evaluated variants. 

\paragraph{Selective gradient routing.}
We rely on selective gradient routing to make the early training stage learn geometry primarily from the clay loss. \cref{tab:ablation_routing} shows our ablation. Applying the clay loss without routing already helps, but routing it to both position and opacity yields the best PSNR. Maintaining a weak residual gradient from the RGB loss is also beneficial, particularly on textureless or flat regions where clay images provide limited cues. We additionally ablate the lock duration: locking for around $40\%$ of total iterations is most helpful, while longer durations leave the BRDF path with insufficient time to refine the geometry, leading to performance degradation.

\section{Conclusion}
We presented the Pygmalion Effect in Vision, a framework that reconstructs reflective objects by reducing specular ambiguity through clay-guided supervision. Combining a BRDF-based reflective branch with a clay-guided branch using synthesized clay images enables reliable disentanglement of reflection and geometry. Beyond practical gains, our results suggest that neutralizing specular cues offers a strong inductive bias for learning geometry, pointing toward new directions for challenging materials and appearance modeling.

\begin{ack}
We sincerely thank our colleagues at NAVER AI Lab, especially Sangdoo Yun, for their valuable support and insightful feedback. This work was supported by NAVER Corporation and NAVER Cloud Corporation through research funding and computing resources. Gayoung Lee, Junho Kim, and Jin-Hwa Kim are employees of NAVER Corporation.
\end{ack}

{
    \small
    \bibliographystyle{unsrt}
    \bibliography{main}
}

\appendix
\renewcommand\thesection{A.\arabic{section}}
\renewcommand\thesubsection{A.\arabic{section}.\arabic{subsection}}
\renewcommand\thefigure{A\arabic{figure}}
\renewcommand\thetable{A\arabic{table}}
\setcounter{figure}{0}
\setcounter{table}{0}

\newpage
\section*{Technical Appendices and Supplementary Material}
\section{Analogy to the Pygmalion effect}
\label{appx:analogy}
The Pygmalion effect, originally a psychological phenomenon describing how expectations can shape outcomes, finds an intriguing parallel in our study.
In the myth, the sculptor \textit{Pygmalion} creates a statue so lifelike that his belief in its beauty brings it to life.
Similarly, our model is trained on datasets curated through the expectations of ``what a correct reflective object should look like.''
These expectations, embedded in the proposed image-to-clay translation, recursively reinforce the model’s perception of the object.

In our work, this metaphor captures the essence of geometry-aware generative reconstruction:
our learning framework not only reconstructs the observed world but also projects the internal geometric belief of the image-to-clay translation model back into the image domain, closing a loop between seeing and shaping.
By acknowledging this Pygmalion effect in vision, we emphasize the importance of critical evaluation, ensuring that the visual worlds our models bring to life remain faithful to reality rather than to their own internal expectations through evaluation benchmarks.

\section{Multi-view dataset for image-to-clay translation}
\label{appx:data-prep-i2c}
To train our multi-view image-to-clay translation model, we construct a dataset of paired multi-view renderings — each sample contains the same object rendered with a reflective material and with a diffuse clay material, sharing identical camera poses. Our dataset comprises $7{,}996$ objects sourced from HY3D-Bench~\cite{hunyuan3d2026hy3dbench} and an additional $154$ objects from Objaverse++~\cite{lin2025objaverseplusplus}, all rendered using our pipeline described below.

\paragraph{Mesh sources.}
We primarily use HY3D-Bench~\cite{hunyuan3d2026hy3dbench}, a curated collection of $252$K+ watertight, manifold meshes that have been cleaned and normalized through a professional preprocessing pipeline. Compared to raw Objaverse~\cite{objaverse,objaverseXL} meshes, which often contain non-manifold geometry, holes, and inconsistent normal orientations, HY3D-Bench provides geometry that is reliably renderable across diverse camera poses without artifacts such as backface leakage or inconsistent shading. This is particularly important for our paired-rendering setup, where any geometric inconsistency between the reflective and clay passes would inject spurious supervision signal. We additionally include $154$ objects from Objaverse++~\cite{lin2025objaverseplusplus}, a quality-annotated subset of Objaverse, to retain a small portion of original surface textures.

\paragraph{Synthetic rendering pipeline.}
We render each mesh using Blender's Cycles path tracer at $768 \times 768$ resolution. Each mesh is first normalized so that its bounding-box maximum dimension fits within a unit cube centered at the origin, with consistent up-axis orientation. For each object, we sample one HDRI environment map from PolyHaven~\cite{polyhaven} as the background, applied through the world shader without making the film transparent — preserving realistic illumination across all views.

\paragraph{Multi-view camera sampling.}
We generate $24$ camera poses per object using Fibonacci-sphere sampling with stochastic perturbation. Camera elevations are distributed in $[15^\circ, 100^\circ]$ from the top, covering the upper, side, and slightly lower regions of the object. Azimuthal angles follow the golden-ratio sequence and are jittered by $\pm 0.15$ rad in azimuth and $\pm 0.1$ rad in elevation, with the camera distance and look-at point also randomly perturbed. This yields well-distributed yet diverse views per object, avoiding the regularity of a fixed sampling pattern that the model could overfit to.

\paragraph{Paired material rendering.}
For each object and camera pose, we render two passes that share identical geometry, illumination, and viewpoint:

\textit{Reflective pass.} With probability $0.5$ we assign a metallic appearance, choosing one of ten preset metal colors (e.g., gold, silver, copper, gunmetal) with metallic factor $1.0$ and roughness sampled uniformly from $[0.0, 0.15]$. Otherwise, we assign a dielectric appearance with random HSV color, metallic factor $0$, roughness in $[0.0, 0.1]$, specular IOR in $[0.8, 1.0]$, and a random clearcoat layer in $[0, 1]$. This split exposes the model to both pure-metal and glossy-painted surface types observed in real-world reflective objects.

\textit{Clay pass.} The same scene is re-rendered with all materials overridden to a neutral diffuse appearance: base color $(0.3, 0.3, 0.3)$, metallic $0$, roughness $1.0$, and specular $0$. The result approximates a Lambertian gray-clay rendering that conveys geometric structure without specular cues.

\begin{figure}[t]
  \centering
  \setlength{\tabcolsep}{2pt} 
  \begin{tabular}{c c c c c c}
    \includegraphics[width=0.155\linewidth]{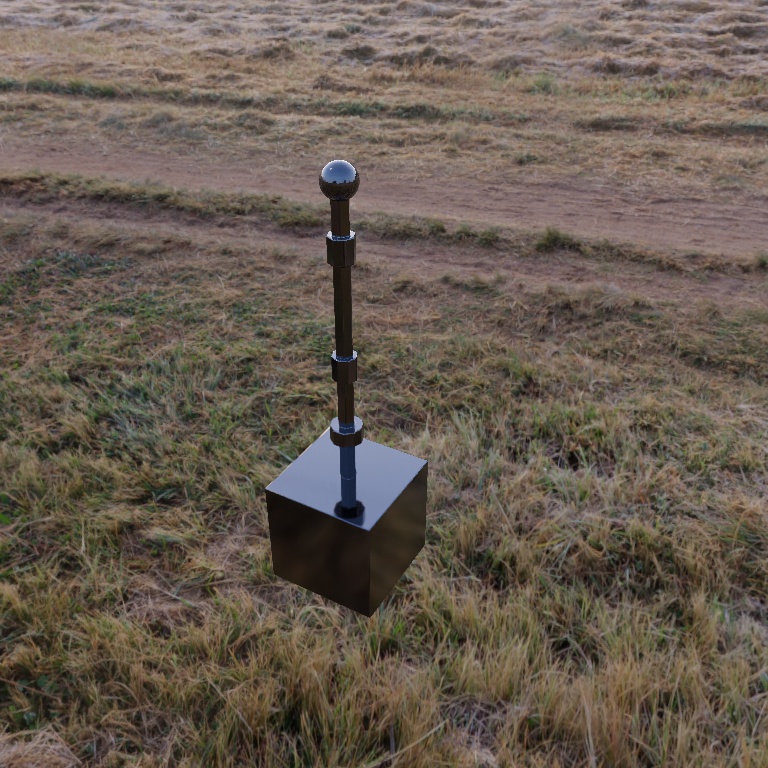} &
    \includegraphics[width=0.155\linewidth]{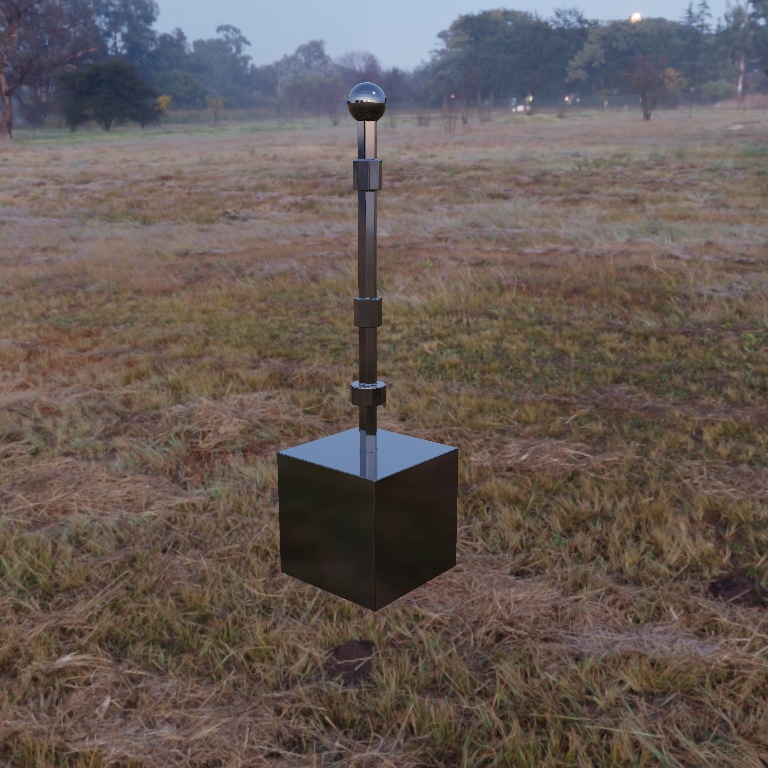} &
    \includegraphics[width=0.155\linewidth]{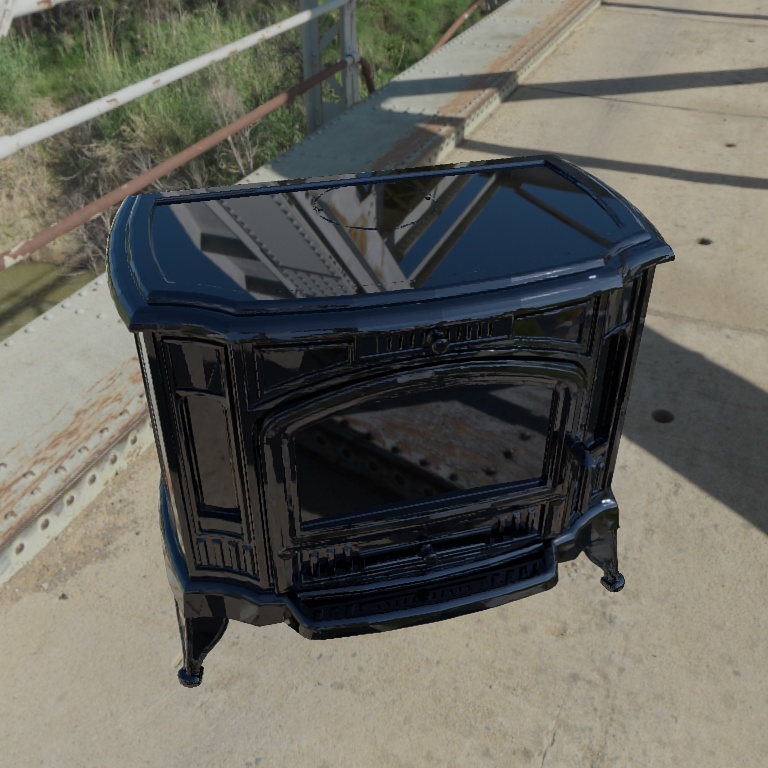} &
    \includegraphics[width=0.155\linewidth]{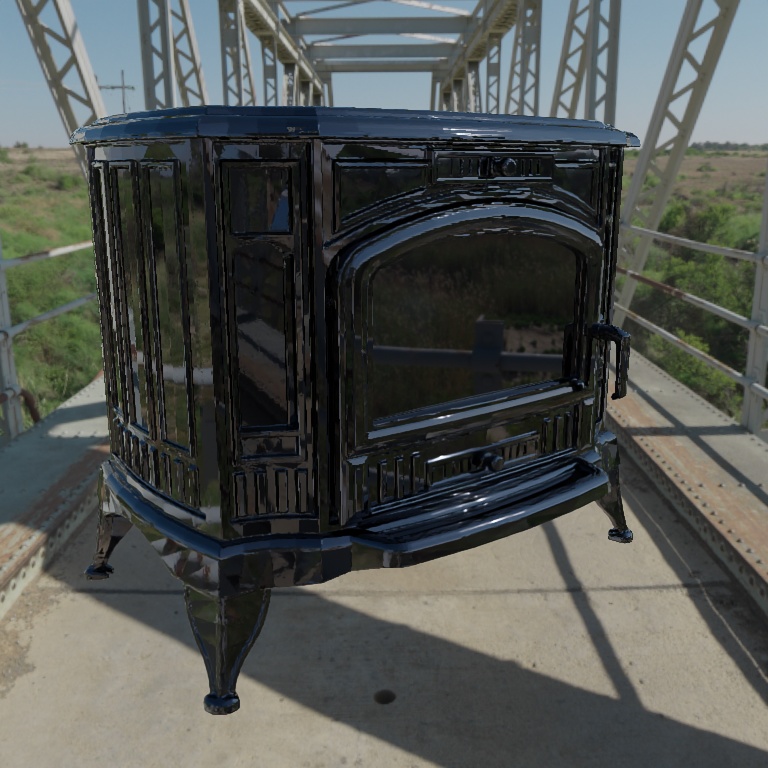} &
    \includegraphics[width=0.155\linewidth]{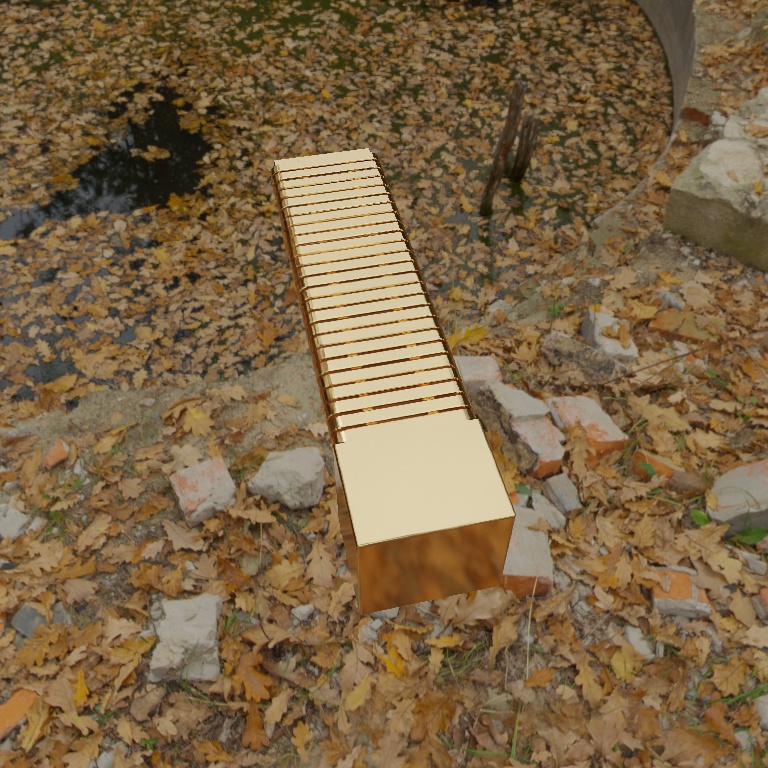} &
    \includegraphics[width=0.155\linewidth]{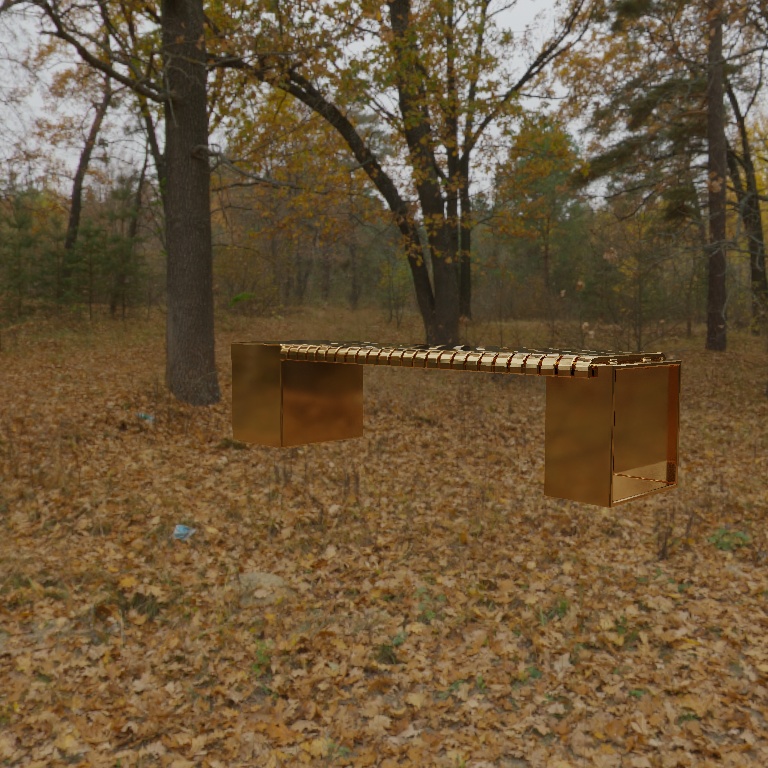} \\
    \includegraphics[width=0.155\linewidth]{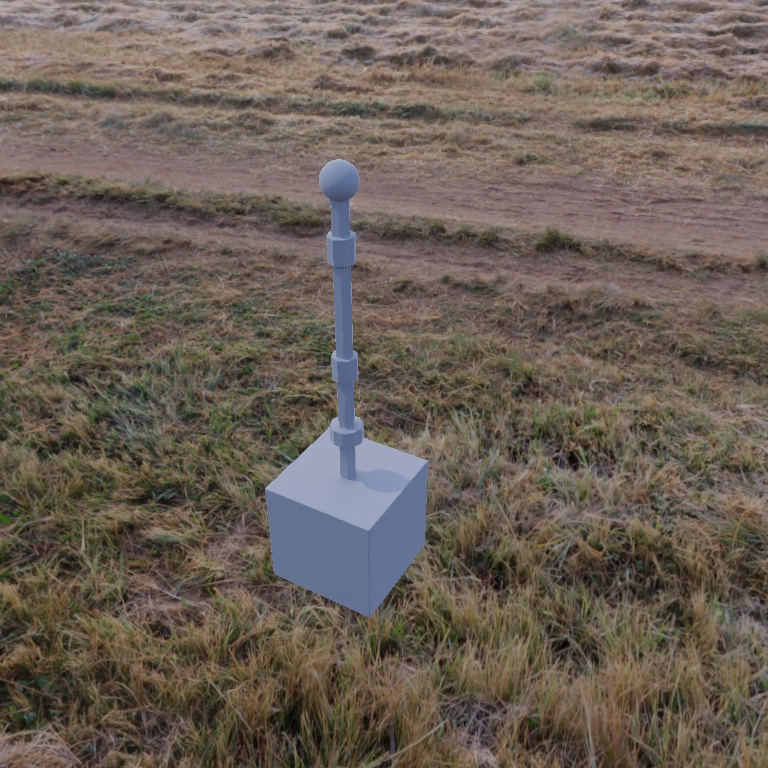} &
    \includegraphics[width=0.155\linewidth]{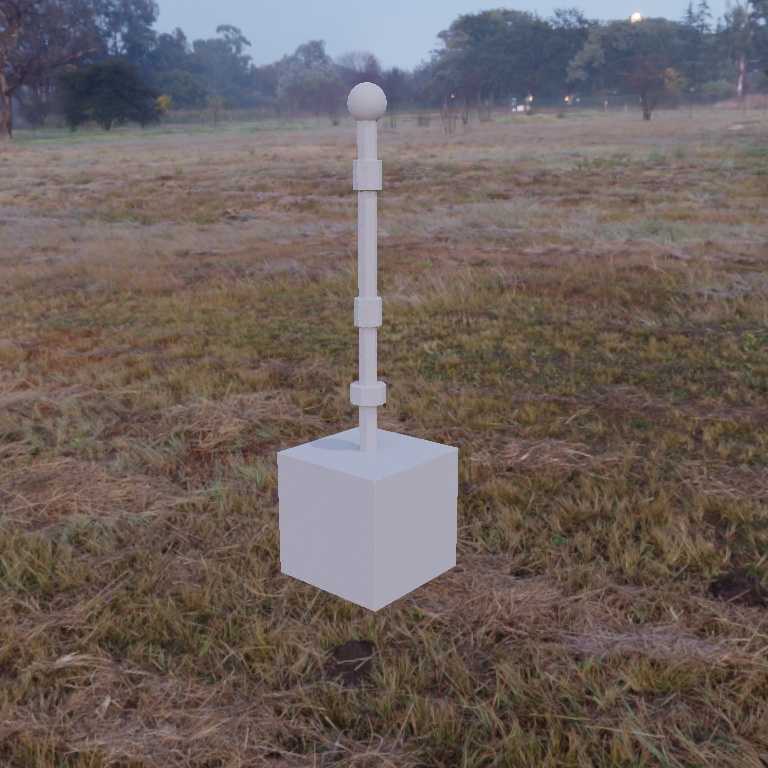} &
    \includegraphics[width=0.155\linewidth]{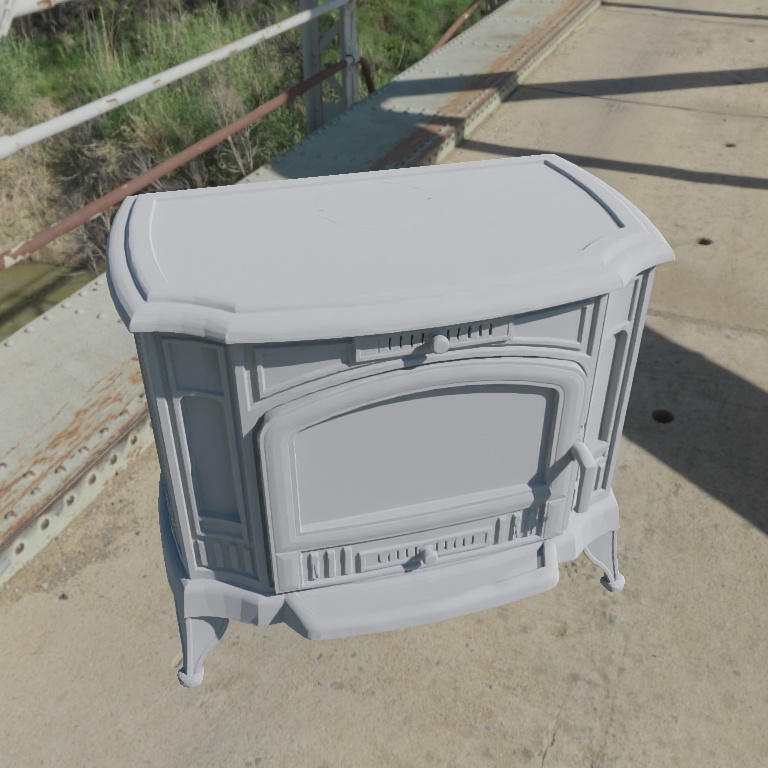} &
    \includegraphics[width=0.155\linewidth]{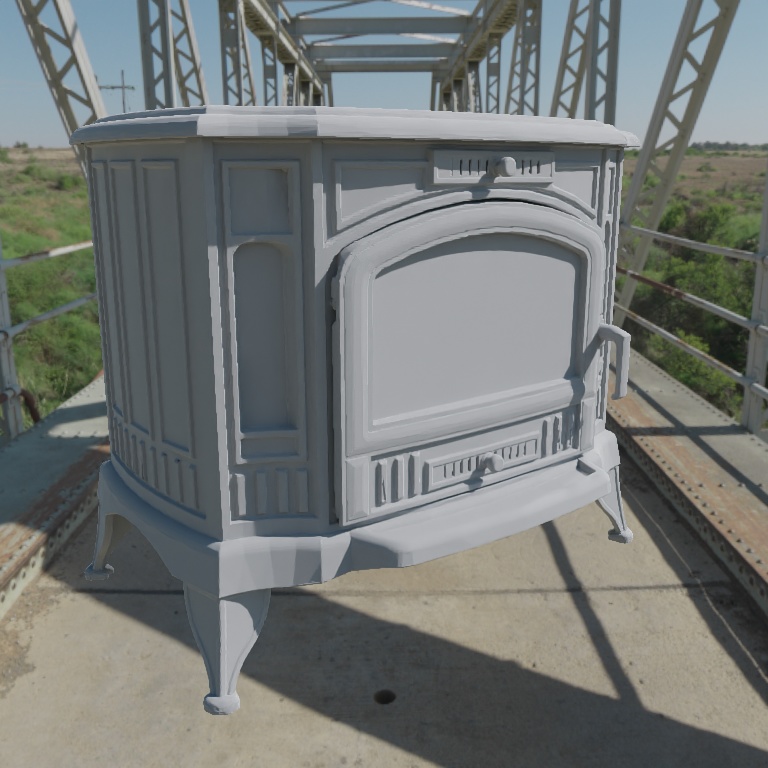} &
    \includegraphics[width=0.155\linewidth]{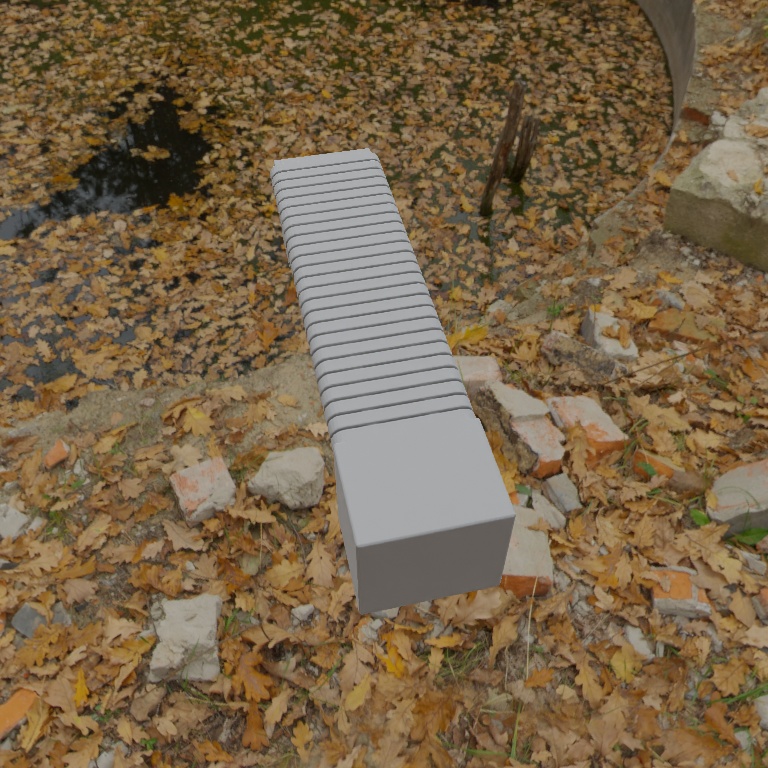} &
    \includegraphics[width=0.155\linewidth]{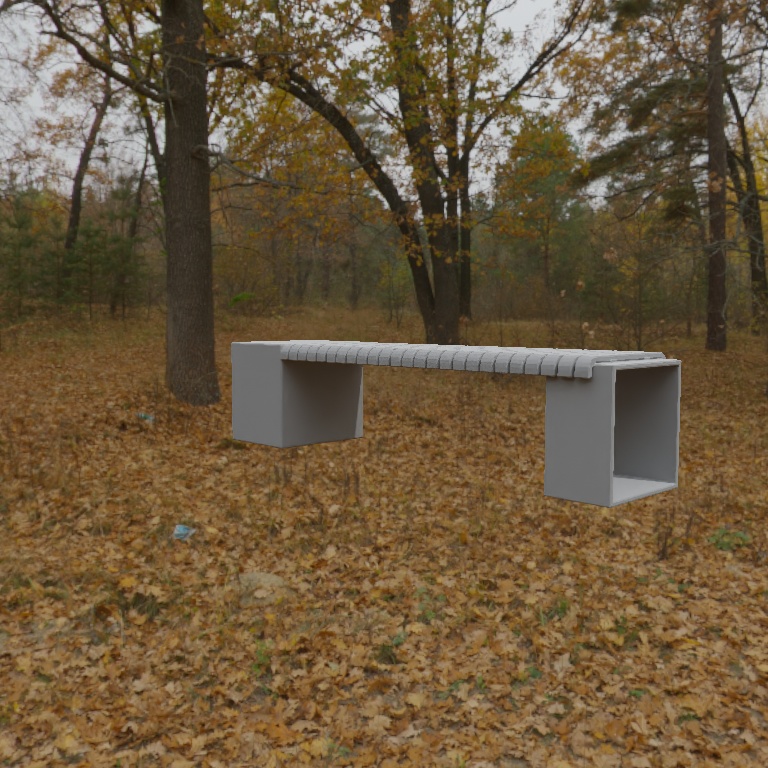} 
  \end{tabular}
  \caption{Visualization of our multi-view dataset. Top: rendered reflective objects, where each pair of columns shows two views of the same object. Bottom: corresponding clay renderings under the same camera and lighting.}
  \label{fig:blender_data_vis}
\end{figure}

Because both passes share the same camera trajectory and HDRI, the resulting $24$-view paired sequence forms a clean multi-view supervision signal for our pseudo-video training (\cref{sec:im2clay_model}). Object masks are additionally rendered via Blender's compositor to support background-free training setups. Camera intrinsics and extrinsics are stored in a NeRF-style \texttt{transforms.json} per object. Example renderings are shown in \cref{fig:blender_data_vis}.

\section{Preliminary: LTX-2 and In-Context LoRA}
\label{appx:ltx_preliminary}
LTX-2~\cite{hacohen2026ltx} is a latent video diffusion transformer trained with rectified flow matching. Video frames are compressed into compact latent tokens by a 3D causal VAE, then processed by a DiT-style backbone. Spatiotemporal positions are injected via 3D rotary position embeddings (RoPE), and the model accepts per-token timesteps, allowing clean and noisy tokens to coexist within a single sequence.

In-Context LoRA (IC-LoRA)~\cite{huang2024iclora} is a parameter-efficient adaptation scheme that recasts conditional generation as in-context inpainting on a frozen diffusion transformer. Conditioning and target tokens are concatenated into a single sequence, with the conditioning portion kept clean and the loss applied only to the noisy target tokens. 
While originally proposed for image diffusion transformers, IC-LoRA has been extended to video diffusion models for controlled editing and generation~\cite{wang2024videoincontext}.
Since only a small set of LoRA adapters is trained, the pretrained joint-attention prior is preserved while the adapters learn the conditioning-to-target mapping.

\section{Detailed formulation of reflective 2D Gaussian Splatting}
\label{appx:2dgs}

2D Gaussian Splatting (2DGS)~\cite{huang20242d} improves upon Gaussian-based representations 
by defining oriented 2D disks instead of volumetric Gaussians. 
Each Gaussian is represented on a local tangent plane, parameterized by its 
center $\mathbf{p}$, two tangential directions $\mathbf{t}_u, \mathbf{t}_v$, 
and scaling factors $s_u, s_v$. 
A surface point on the disk is expressed as:
\begin{equation}
    \mathbf{P}(u, v) = \mathbf{p} + s_u \mathbf{t}_u u + s_v \mathbf{t}_v v.
\end{equation}
The Gaussian weight over this plane is:
\begin{equation}
    \mathcal{G}(u, v) = \exp\!\left(-\frac{u^2 + v^2}{2}\right).
\end{equation}
During rendering, a pixel $\mathbf{x}$ is projected onto the local plane of each Gaussian,
yielding the corresponding local coordinates $\mathbf{u}(\mathbf{x}) = [u(\mathbf{x}), v(\mathbf{x})]^\top$
through the inverse projection of $\mathbf{P}(u,v)$.
The color at $\mathbf{x}$ is then obtained by the standard alpha-compositing rule:
\begin{equation}
    \mathbf{c}(\mathbf{x}) =
    \sum_{i=1}^{N} 
    \mathbf{c}_i\, \alpha_i\, \hat{\mathcal{G}}_i(\mathbf{u}(\mathbf{x}))
    \prod_{j=1}^{i-1} \left( 1 - \alpha_j\, \hat{\mathcal{G}}_j(\mathbf{u}(\mathbf{x})) \right),
\end{equation}
where $\mathbf{c}_i$ is the Gaussian color, $\alpha$ is the Gaussian opacity, and $\hat{\mathcal{G}}$ is the visibility-corrected Gaussian weight of $\mathcal{G}$.

We build upon MaterialRefGS~\cite{zhangmaterialrefgs}, which is based on Reflective Gaussian Splatting (\reflectgs{})~\cite{yao2025reflective}, where each Gaussian is associated with both geometric and BRDF parameters. Each Gaussian primitive is associated with a feature vector 
$\vb_i = [\lambda_i, m_i, r_i, \mathbf{n}_i]$, 
where $\lambda_i$, $m_i$, $r_i$, and $\mathbf{n}_i$ denote albedo, metallic, roughness, 
and surface normal, respectively, for the $i$-th of $N$ Gaussians. 
These attributes are composited using the same alpha blending as in 2DGS:
\begin{equation}
    \mathbf{B}(\vx) =
    \sum_{i=1}^{N} 
    \vb_i\, \alpha_i\, \hat{\mathcal{G}}_i(\mathbf{u}(\vx))
    \prod_{j=1}^{i-1} \left( 1 - \alpha_j\, \hat{\mathcal{G}}_j(\mathbf{u}(\vx)) \right),
    \label{eq:blending_function_appx}
\end{equation}
where $\mathbf{B} = [\Lambda, M, R, \mN]$ represents the accumulated material features, while $\alpha$ and $\hat{\mathcal{G}}$ denote opacity and 2D Gaussian evaluated at the local tangent coordinate $\mathbf{u}(\vx)$, respectively.

The final outgoing radiance $L(\omega_o)$ at the viewing direction $\omega_o$ 
is computed using the physically-based rendering equation:
\begin{equation}
    L(\omega_o) = \int_{\Omega} L_i(\omega_i)\, 
    f(\omega_i, \omega_o)\, (\omega_i \cdot \mN)\, d\omega_i,
\end{equation}
where $L_i(\omega_i)$ denotes the incoming radiance from direction $\omega_i$, 
and $f(\omega_i, \omega_o)$ is the bidirectional reflectance distribution function (BRDF).

The BRDF is decomposed into diffuse and specular terms $f = f_d + f_s$. While the diffuse term $f_d$ is straightforward to integrate, the specular term follows the GGX microfacet model~\cite{walter2007microfacet}:
\begin{equation}
    f_s(\omega_i, \omega_o) =
    \frac{D\, G\, F}{4(\omega_o \cdot \mN)(\omega_i \cdot \mN)},
\end{equation}
where $D$, $G$, and $F$ denote the normal distribution, 
geometry, and Fresnel terms, respectively.
To efficiently evaluate the specular reflection, we compute the specular radiance $L_s(\omega_o)$ using the precomputed split-sum formulation:
\begin{equation}
\begin{split}
    \label{eqn:split-sum}
    L_s(\omega_o) \approx &
    \left( \int_{\Omega} f_s(\omega_i, \omega_o)
    (\omega_i \cdot \mN)\, d\omega_i \right) \\
    & \cdot
    \left( \int_{\Omega} L_i(\omega_i)
    D(\omega_i, \omega_o)
    (\omega_i \cdot \mN)\, d\omega_i \right),
\end{split}
\end{equation}
where the first term depends only on the view direction and roughness and can therefore be precomputed and stored in a 2D lookup texture, 
while the second term is efficiently evaluated from prefiltered environment maps.

Following the previous work~\cite{yao2025reflective}, we further separate the lighting into direct $L_\text{dir}$, the second term in \cref{eqn:split-sum}, and indirect $L_\text{ind}$ components to account for inter-reflection:
\begin{equation}
\begin{split}
    L_s'(\omega_o) \approx &
    \left( \int_{\Omega} f_s(\omega_i, \omega_o)
    (\omega_i \cdot \mN)\, d\omega_i \right) \\
    & \cdot 
    \big( L_{\text{dir}} \cdot \chi_v + L_{\text{ind}} \cdot (1 - \chi_v) \big),
\end{split}
\end{equation}
where $\chi_v \in \{0,1\}$ is a binary visibility indicator computed by ray tracing on the extracted mesh surface. The indirect light $L_\text{ind}$ is defined as follows:
\begin{equation}
\begin{split}
    L_{\text{ind}} =
    \sum_{i=1}^{N} 
    \vl_{\text{ind},i}\, \alpha_i
    \prod_{j=1}^{i-1} (1 - \alpha_j),
\end{split}
\end{equation}
where $\vl_{\text{ind},i}$ denotes an additional view-dependent color evaluated along the reflected direction and modeled using spherical harmonics~\cite{muller1966spherical}.


\section{Additional implementation details}
\label{appx:imple_details}
We fine-tune LTX-2~\cite{hacohen2026ltx} for $50{,}000$ iterations on $8 \times$ NVIDIA H100 GPUs 
with a batch size of $1$ per GPU, where each sample contains all $24$ multi-view 
images. For each sample, the $24$ views are randomly resampled 
with replacement and augmented with random cropping. We adapt the model with 
LoRA~\cite{hu2022lora} modules of rank $32$ and use AdamW with a 
learning rate of $2 \times 10^{-4}$. All training samples share a unified text 
prompt: \texttt{"A 3D rendered object with diffuse white clay-like material 
and surface properties"}. To enable the anchor-based inference scheme described 
in \cref{sec:im2clay_model}, with probability $0.5$ we expose four randomly 
selected target views as clean anchors.

For the reflective object reconstruction stage, we use a single H100 GPU. Following the baselines, we train for $50{,}000$ iterations on synthetic data and $30{,}000$ iterations on real-world data, except for the NeRF-Casting scenes, which we train longer for $50{,}000$ iterations. Selective gradient routing is applied during the first $20{,}000$ iterations for synthetic data and the first $6{,}000$ iterations for real data. For the mask loss, the normal prior schedule, and other regularization terms, we follow the hyperparameters provided in the official MaterialRefGS~\cite{zhangmaterialrefgs} codebase. Following the official recommendation, we estimate monocular normal maps using the official implementation of Metric3D v2~\cite{hu2024metric3d} and use them as normal supervision. 

Since the MaterialRefGS codebase provides no configuration for the Glossy-Real and NeRF-Casting datasets, we adopt its Ref-Real setting as the default and adjust a few options so that the unbounded backgrounds remain intact: we set the distortion loss weight to zero on both datasets, and for Glossy-Real we additionally disable the normal propagation phase, whose periodic perturbation, designed under the assumption of object-centric synthetic data, introduces background floaters in unbounded real scenes. These adjusted settings are applied identically to the MaterialRefGS baseline and to our clay-guided variant for a fair comparison.

\paragraph{Timing breakdown.}
\cref{tab:timing} reports the wall-clock timing of our full pipeline, measured on the \textit{angel} scene of Glossy Synthetic ($112$ views) using a single NVIDIA H100 GPU. Clay translation accounts for only $3.5\%$ of the total pipeline time ($4.4\%$ including the one-time model loading), and is a one-time pre-processing cost per scene, independent of the number of optimization iterations. The clay translation stage is therefore far from a runtime bottleneck.

\begin{table*}[h]
\centering
\small
\caption{\textbf{Wall-clock timing breakdown.} Measured on the \textit{angel} scene of Glossy Synthetic ($112$ views) with a single NVIDIA H100 GPU.}
\label{tab:timing}
\begin{tabular}{lr}
\toprule[1pt]
Stage & Time \\
\midrule
LTX-2 model loading (one-time) & 1m 16s \\
Clay translation of all $112$ training views & 4m 59s \\
MaterialRefGS optimization ($50{,}000$ iterations) & 2h 15m 7s \\
\midrule
Total & 2h 21m 22s \\
\bottomrule[1pt]
\end{tabular}
\end{table*}

\section{Additional visualization and results}
\label{appx:additional_results}

In this section, we provide additional visualizations of the image-to-clay transformation results and our reconstructed outputs.

\subsection{Multi-view image-to-clay model outputs}
\label{appx:additional_clayvis}
\begin{figure}[t]
  \centering
  \begin{minipage}[c]{0.46\linewidth}
    \centering
    \includegraphics[width=0.32\linewidth]{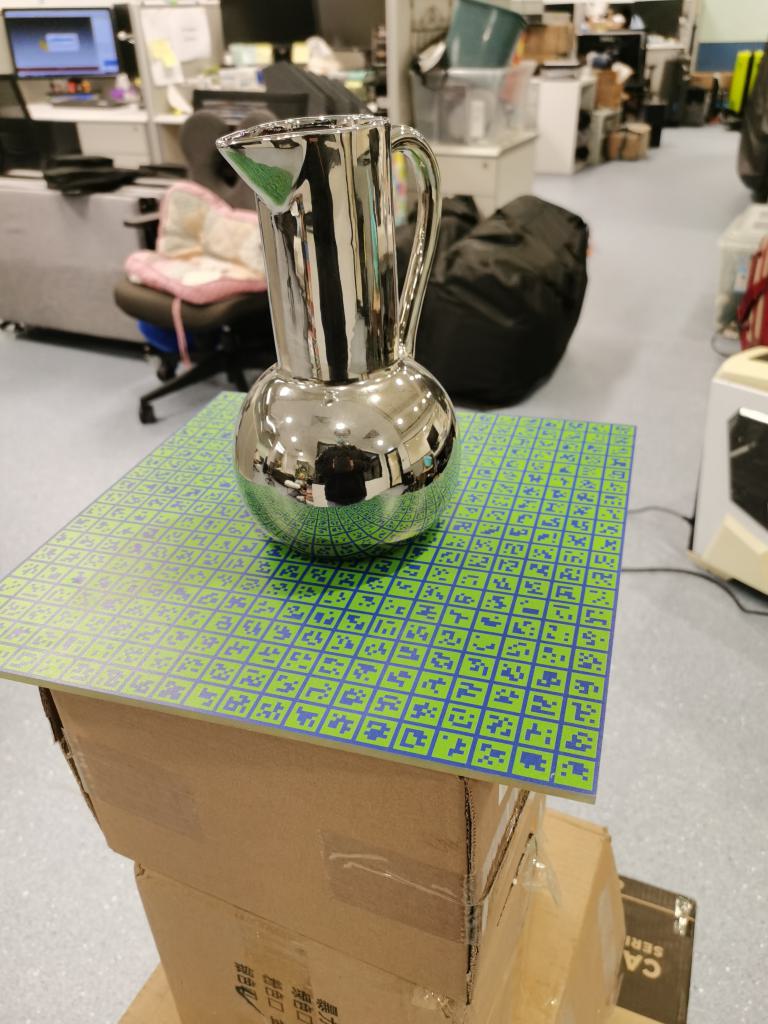}
    \includegraphics[width=0.32\linewidth]{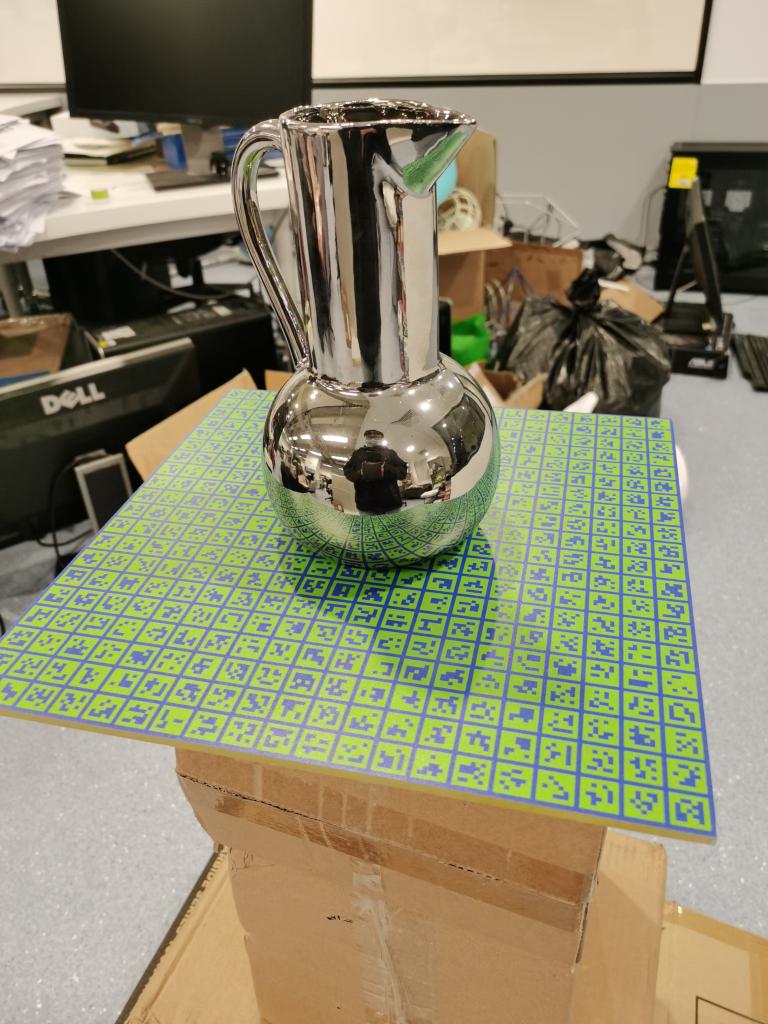}
    \includegraphics[width=0.32\linewidth]{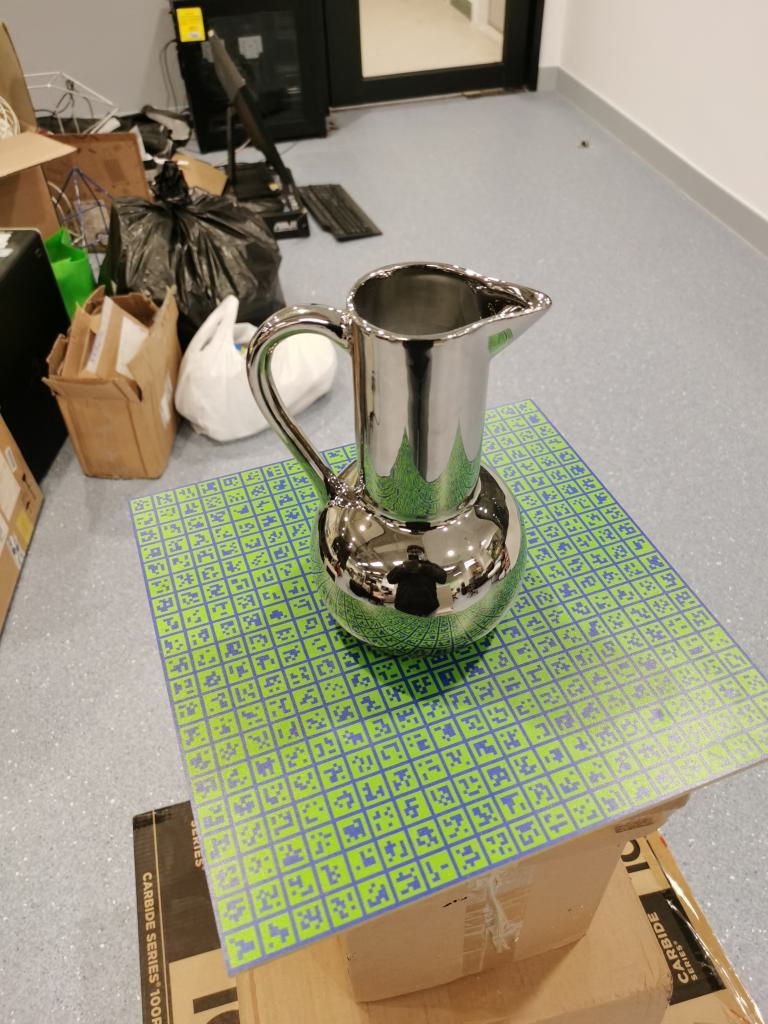}
  \end{minipage}
  \begin{minipage}[c]{0.46\linewidth}
    \centering
    \includegraphics[width=0.32\linewidth]{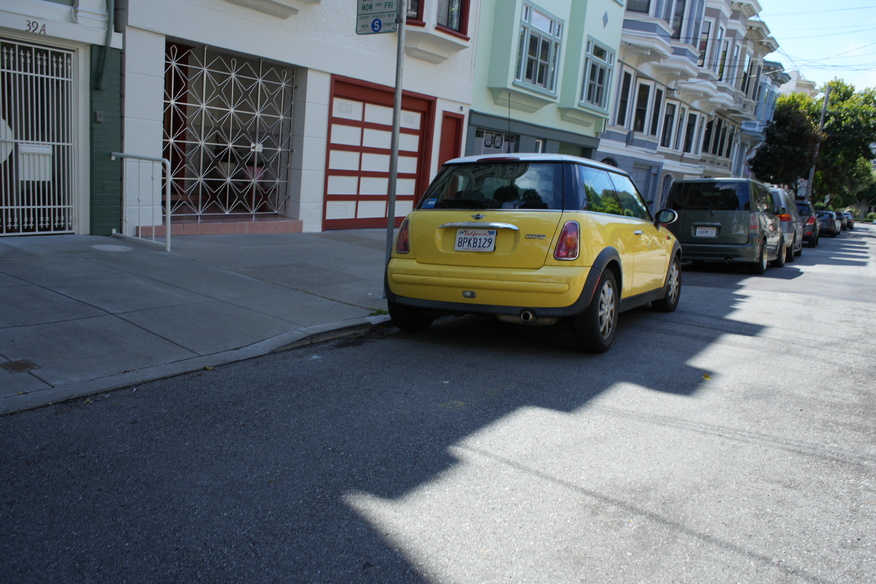}
    \includegraphics[width=0.32\linewidth]{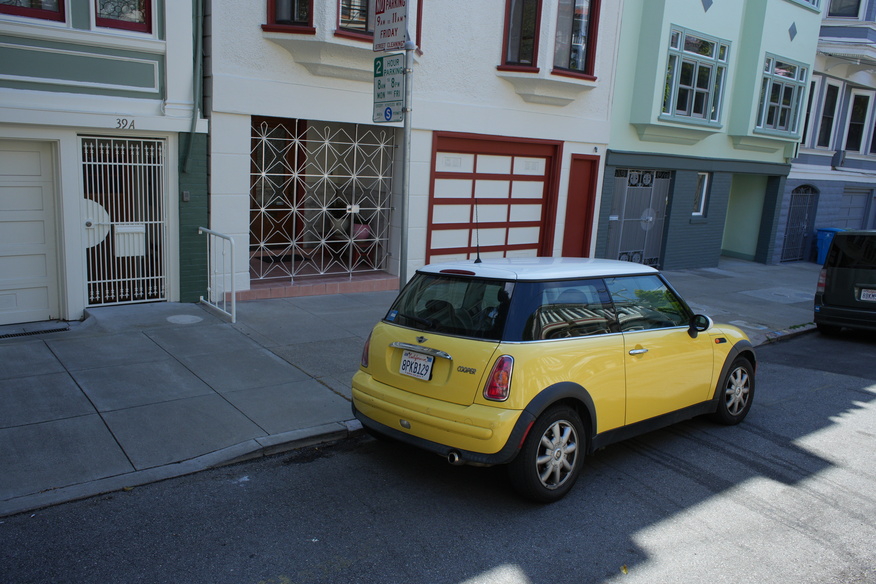}
    \includegraphics[width=0.32\linewidth]{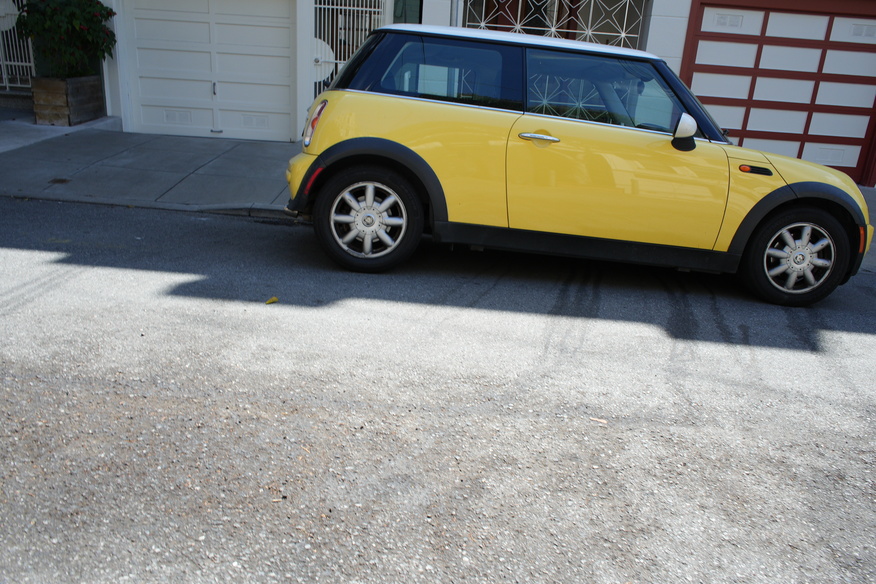}\\[1pt]
    \includegraphics[width=0.32\linewidth]{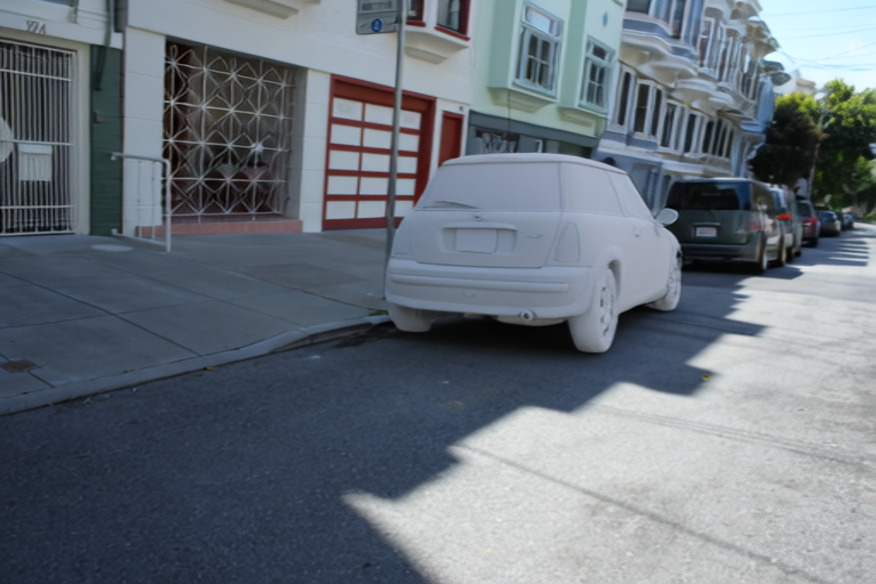}
    \includegraphics[width=0.32\linewidth]{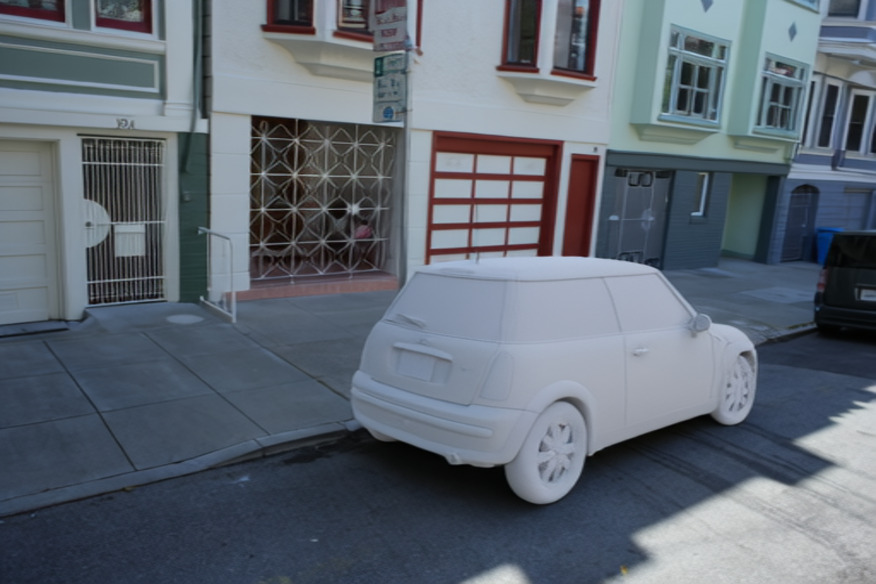}
    \includegraphics[width=0.32\linewidth]{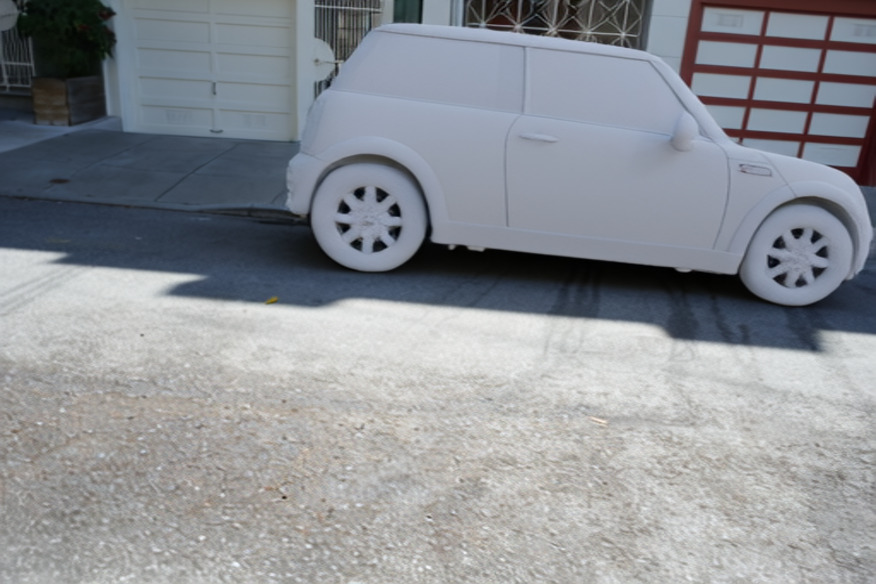}
  \end{minipage}\\[1pt]
  
  \begin{minipage}[c]{0.46\linewidth}
    \centering
    \includegraphics[width=0.32\linewidth]{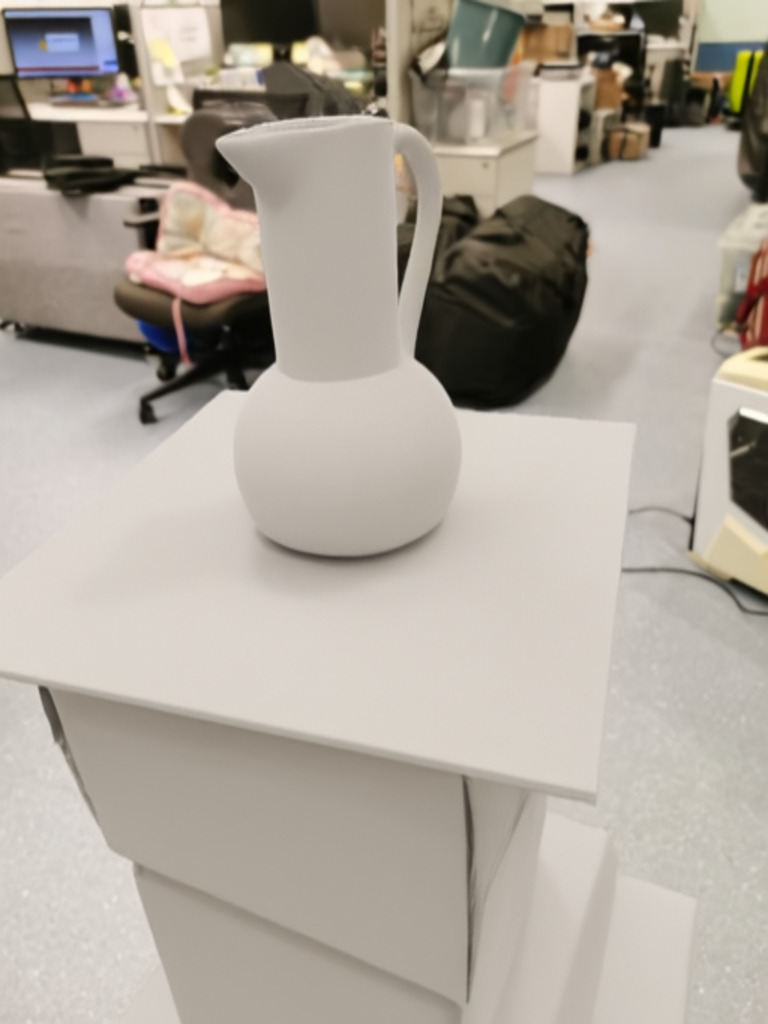}
    \includegraphics[width=0.32\linewidth]{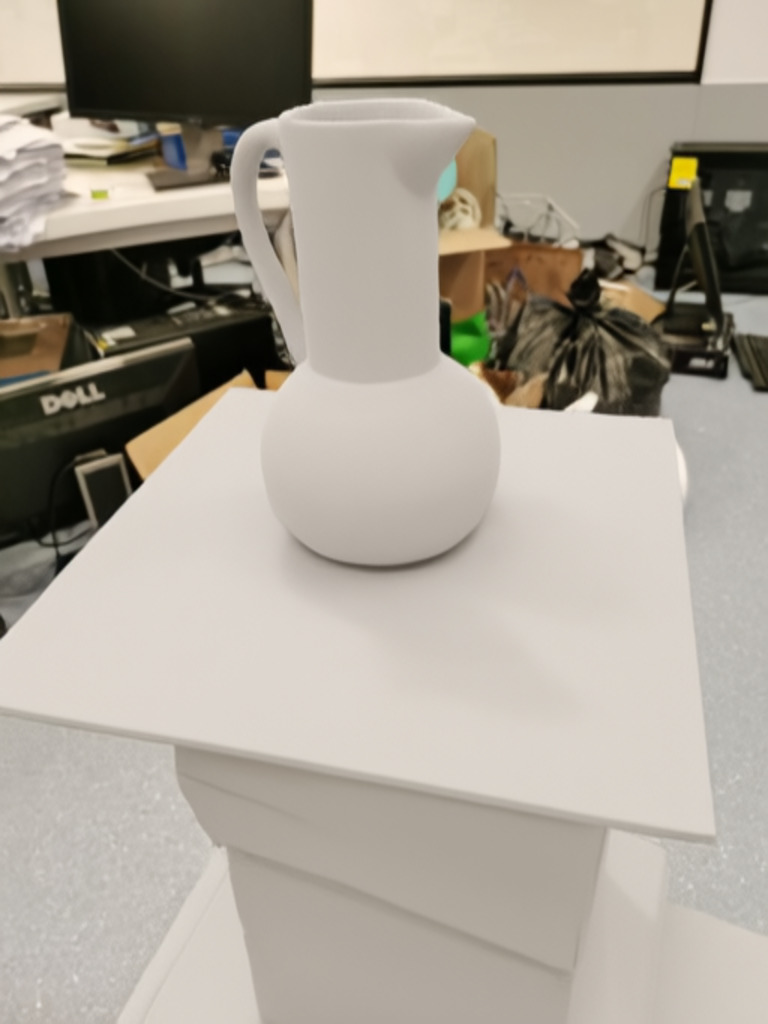}
    \includegraphics[width=0.32\linewidth]{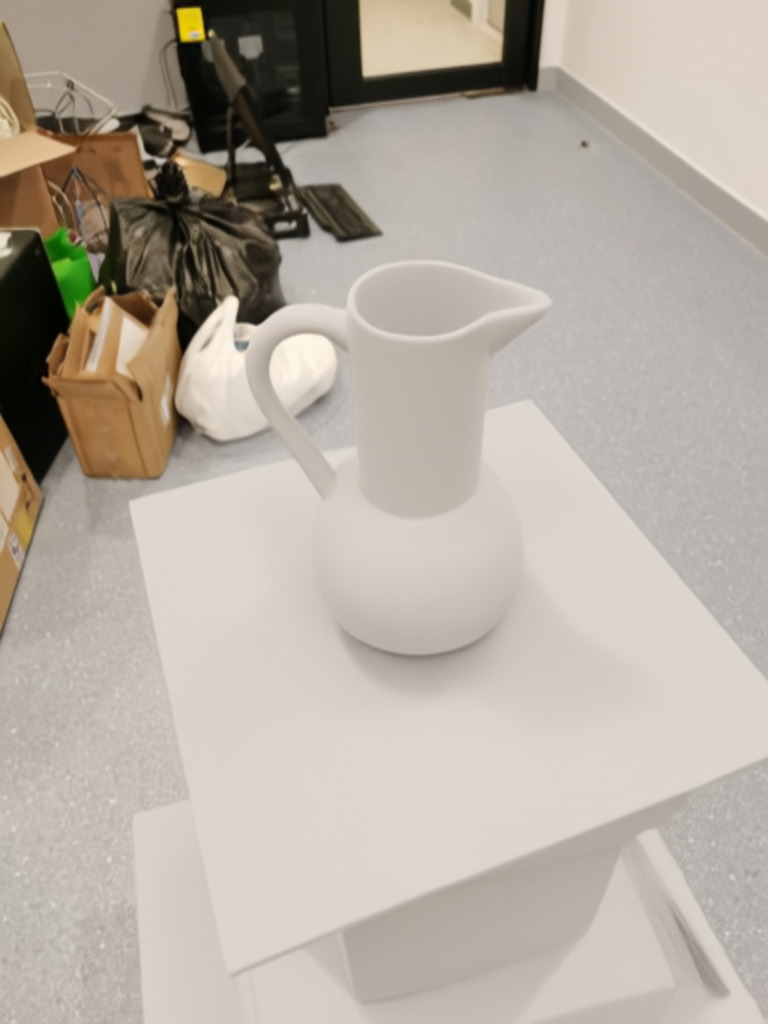}
  \end{minipage}
  \begin{minipage}[c]{0.46\linewidth}
    \centering
    \includegraphics[width=0.32\linewidth]{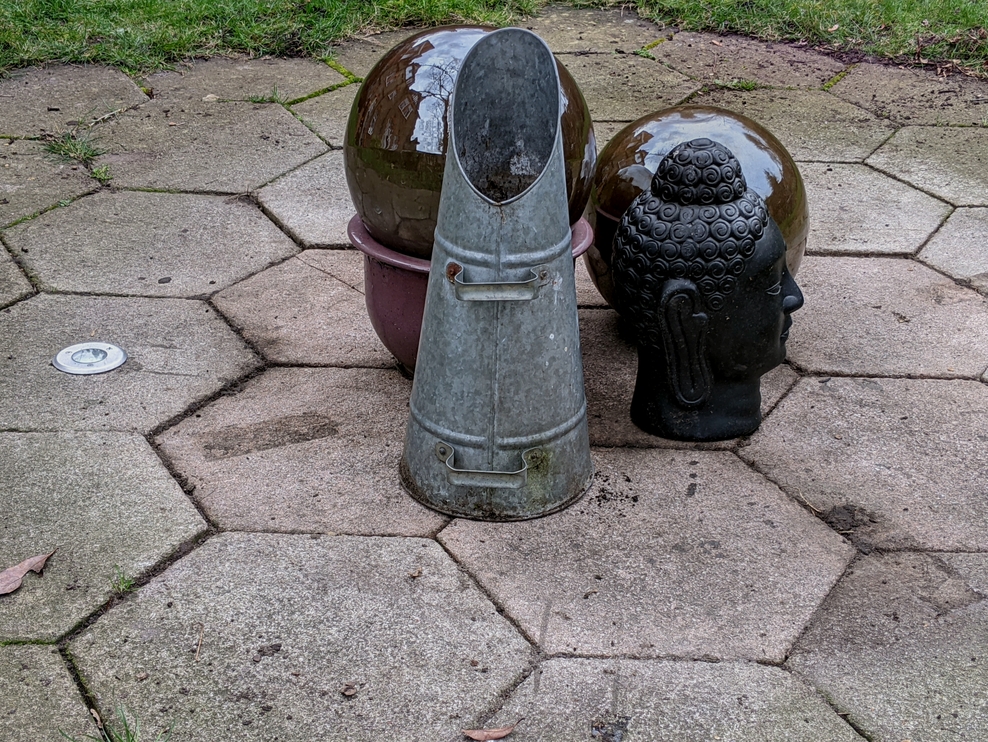}
    \includegraphics[width=0.32\linewidth]{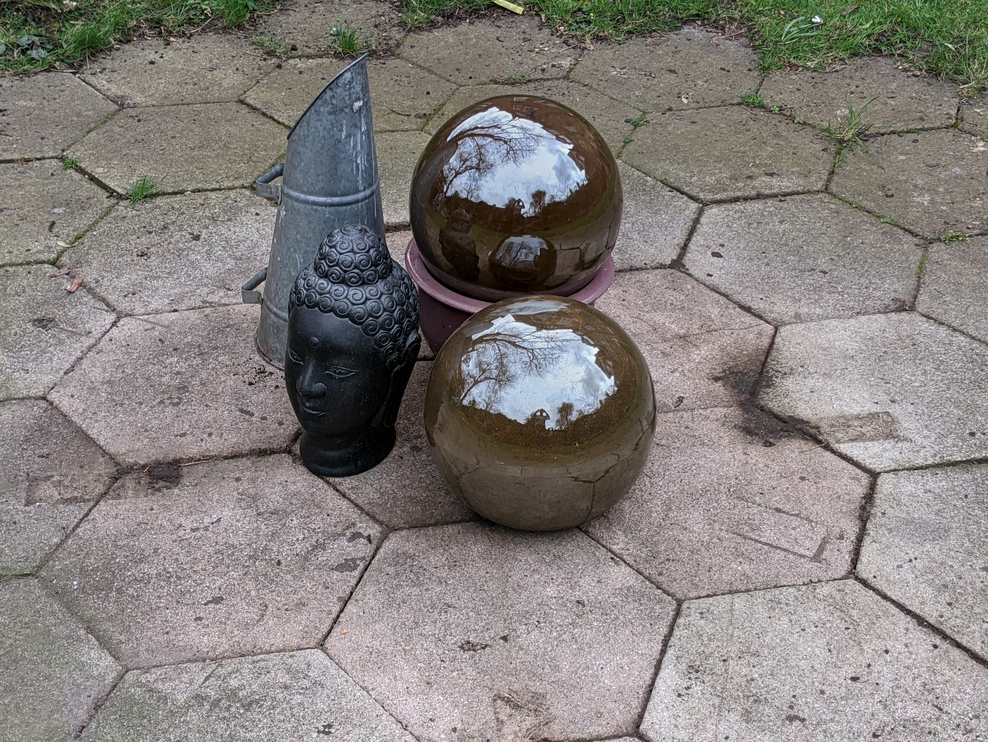}
    \includegraphics[width=0.32\linewidth]{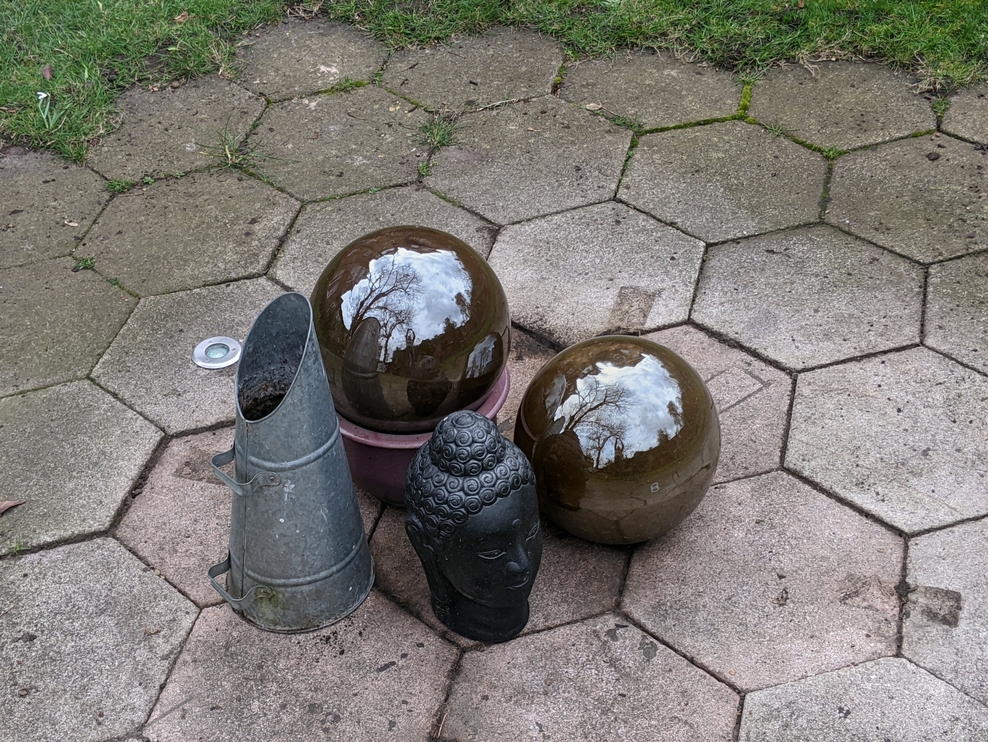}\\[1pt]
    \includegraphics[width=0.32\linewidth]{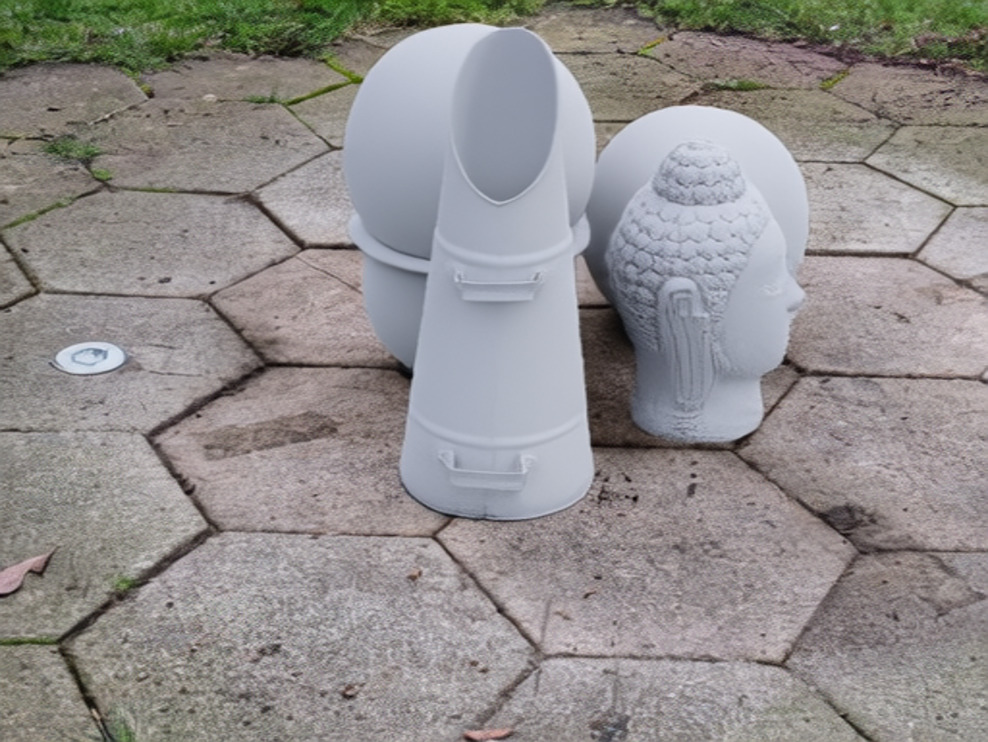}
    \includegraphics[width=0.32\linewidth]{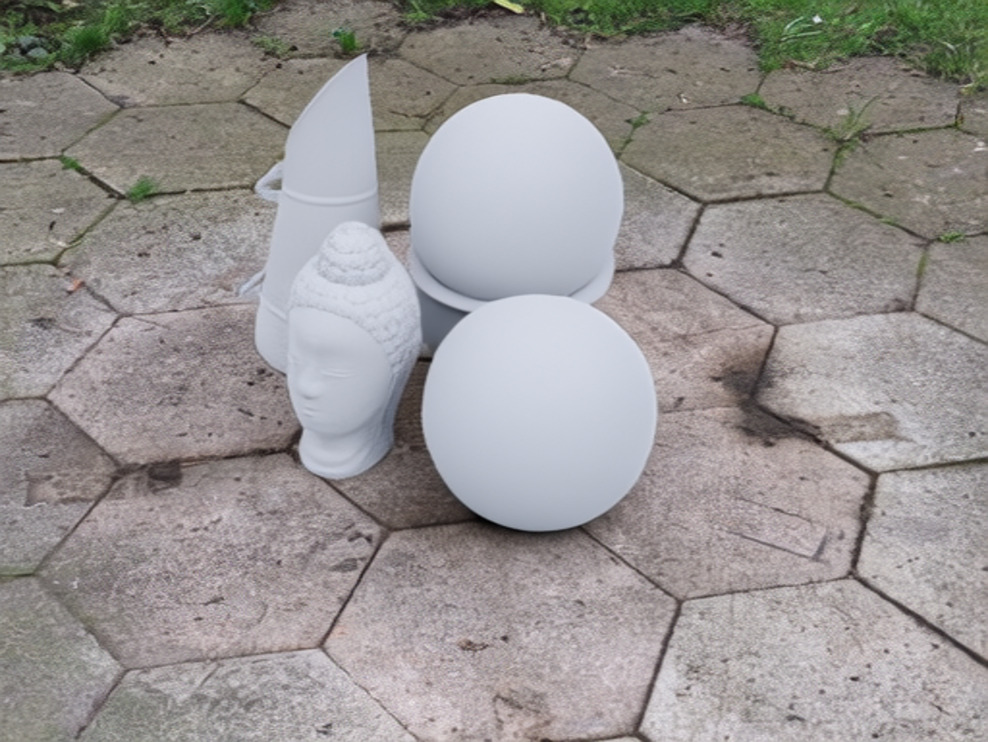}
    \includegraphics[width=0.32\linewidth]{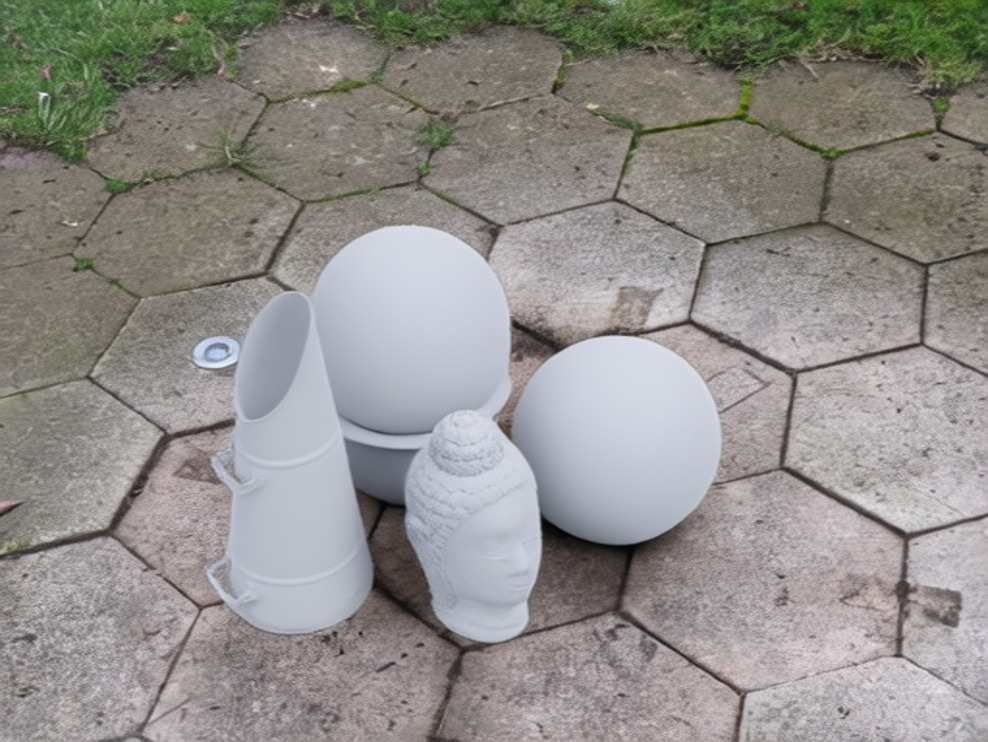}
  \end{minipage}\\[1pt]

\caption{Visualization of our multi-view image-to-clay outputs on real-world
scenes. Synthetic results are shown in \cref{fig:clayoutput_syn}.}
  \label{fig:clayoutput}
\end{figure}

\Cref{fig:clayoutput} shows our image-to-clay translation results on real-world reflective objects. Results on synthetic scenes are visualized in \cref{fig:clayoutput_syn} of the main paper. Despite being trained only on synthetic data, our model generalizes well to real-world scenes thanks to the strong prior of LTX-2. Notably, for object-centric scenes, the model also translates auxiliary structures into clay, but since the underlying geometry remains aligned with the RGB image, this does not affect downstream training. For car windows—a region where monocular normal estimators frequently fail—our model produces clean matte surfaces. Although our synthetic training pairs contain only a single material per object, objects with spatially varying materials (\eg, the car in \cref{fig:clayoutput}, where glass, metal, and rubber coexist) are translated reliably without noticeable artifacts at material boundaries, which we attribute to the material diversity of our training set and the strong object prior of the video backbone.

\begin{table*}[t]
\centering
\small
\caption{Per-scene image quality comparison on synthetic datasets. The intensity of the red color signifies a better result. The Avg.\ column is bolded for readability.}
\resizebox{\textwidth}{!}{%
\begin{tabular}{lcccccccccccccccc}
\toprule[1pt]
\textbf{Datasets} 
         & \multicolumn{7}{c}{\textbf{Shiny Blender}~\citep{verbin2024refnerf}} 
         & \multicolumn{9}{c}{\textbf{Glossy Synthetic}~\citep{liu2023nero}} \\
\textbf{Scenes}   
         & \textbf{Ball} & \textbf{Car} & \textbf{Coffee} & \textbf{Helmet} & \textbf{Teapot} & \textbf{Toaster} & \textbf{Avg.}
         & \textbf{Angel} & \textbf{Bell} & \textbf{Cat} & \textbf{Horse} & \textbf{Luyu} & \textbf{Potion} & \textbf{Tbell} & \textbf{Teapot} & \textbf{Avg.} \\
\midrule
\multicolumn{17}{c}{\textbf{PSNR} $\uparrow$} \\
\midrule
2DGS          & 25.97 & 26.38 & 32.31 & 27.42 & 44.97 & 20.42 & 29.58 & 26.95 & 24.79 & 30.65 & 25.18 & 26.89 & 29.50 & 23.28 & 21.29 & 26.07 \\
GShader       & 30.99 & 27.96 & 32.39 & 28.32 & 45.86 & 26.28 & 31.97 & 25.08 & 28.07 & 31.81 & 26.56 & 27.18 & 30.09 & 24.48 & 23.58 & 27.11 \\
R3DG          & 23.64 & 25.92 & 30.10 & 25.01 & 43.15 & 18.80 & 27.77 & 24.90 & 23.51 & 27.59 & 23.37 & 24.68 & 27.29 & 21.25 & 20.47 & 24.13 \\
3DGS-DR       & 33.43 & 30.48 & \secondbest{34.53} & 31.44 & \secondbest{47.04} & 26.76 & 33.95 & 29.07 & 30.60 & 32.59 & 26.17 & 28.96 & 32.65 & 29.03 & 25.77 & 29.36 \\
Ref-GS        & \secondbest{36.10} & 30.94 & 34.38 & \first{33.40} & 46.69 & 27.28 & \thirdbest{34.80} & 30.34 & \thirdbest{31.70} & \thirdbest{33.15} & \thirdbest{27.45} & 29.46 & 32.64 & \secondbest{30.08} & 26.47 & 30.16 \\
\reflectgs{}  & \first{37.01} & \thirdbest{31.04} & \first{34.63} & 32.32 & \first{47.16} & \thirdbest{28.05} & \secondbest{35.04} & \thirdbest{30.38} & \first{32.86} & 33.01 & 27.05 & \thirdbest{30.04} & \thirdbest{33.07} & 29.84 & \thirdbest{26.68} & 30.37 \\
MaterialRefGS & 35.38 & \secondbest{31.27} & 34.34 & \thirdbest{32.87} & 44.97 & \secondbest{29.25} & 34.68 & \secondbest{31.42} & 31.40 & \first{33.57} & \secondbest{27.52} & \secondbest{30.25} & \first{33.47} & \first{30.21} & \secondbest{26.99} & \secondbest{30.60} \\
Ours          & \thirdbest{35.53} & \first{31.31} & \thirdbest{34.45} & \secondbest{33.08} & \thirdbest{47.00} & \first{29.35} & \first{35.12} & \first{31.53} & \secondbest{31.93} & \secondbest{33.46} & \first{27.62} & \first{30.34} & \secondbest{33.45} & \thirdbest{30.05} & \first{27.04} & \first{30.68} \\
\midrule
\multicolumn{17}{c}{\textbf{SSIM} $\uparrow$} \\
\midrule
2DGS          & 0.934 & 0.930 & 0.972 & 0.953 & \thirdbest{0.997} & 0.892 & 0.946 & 0.918 & 0.911 & 0.958 & 0.909 & 0.918 & 0.939 & 0.902 & 0.886 & 0.918 \\
GShader       & 0.966 & 0.932 & 0.971 & 0.951 & 0.996 & 0.929 & 0.958 & 0.914 & 0.919 & 0.961 & 0.933 & 0.914 & 0.936 & 0.898 & 0.901 & 0.922 \\
R3DG          & 0.888 & 0.922 & 0.963 & 0.931 & 0.995 & 0.858 & 0.926 & 0.894 & 0.888 & 0.934 & 0.878 & 0.889 & 0.911 & 0.875 & 0.869 & 0.892 \\
3DGS-DR       & 0.979 & 0.963 & \first{0.976} & 0.971 & \thirdbest{0.997} & 0.942 & 0.971 & 0.942 & 0.959 & \thirdbest{0.973} & 0.933 & 0.943 & 0.959 & 0.958 & 0.942 & 0.951 \\
Ref-GS        & \thirdbest{0.981} & 0.961 & 0.973 & \first{0.975} & \thirdbest{0.997} & \thirdbest{0.950} & 0.973 & \thirdbest{0.955} & \secondbest{0.965} & \thirdbest{0.973} & \thirdbest{0.948} & 0.946 & 0.957 & 0.956 & 0.944 & 0.956 \\
\reflectgs{}  & \thirdbest{0.981} & \thirdbest{0.964} & \first{0.976} & 0.971 & \first{0.998} & 0.948 & 0.973 & 0.954 & \first{0.969} & \thirdbest{0.973} & 0.944 & \thirdbest{0.952} & \thirdbest{0.963} & \thirdbest{0.962} & \thirdbest{0.947} & \thirdbest{0.958} \\
MaterialRefGS & \first{0.984} & \secondbest{0.967} & \thirdbest{0.976} & \thirdbest{0.973} & 0.997 & \secondbest{0.954} & \secondbest{0.975} & \secondbest{0.960} & 0.959 & \first{0.977} & \secondbest{0.949} & \secondbest{0.955} & \first{0.965} & \first{0.965} & \secondbest{0.953} & \secondbest{0.960} \\
Ours          & \first{0.984} & \first{0.967} & \thirdbest{0.976} & \secondbest{0.974} & \secondbest{0.997} & \first{0.955} & \first{0.976} & \first{0.960} & \thirdbest{0.963} & \secondbest{0.977} & \first{0.950} & \first{0.955} & \secondbest{0.966} & \secondbest{0.964} & \first{0.953} & \first{0.961} \\
\midrule
\multicolumn{17}{c}{\textbf{LPIPS} $\downarrow$} \\
\midrule
2DGS          & 0.156 & 0.052 & 0.079 & 0.079 & 0.008 & 0.127 & 0.084 & 0.072 & 0.109 & 0.060 & 0.071 & 0.066 & 0.097 & 0.125 & 0.101 & 0.088 \\
GShader       & 0.121 & 0.044 & 0.078 & 0.074 & 0.007 & 0.079 & 0.067 & 0.082 & 0.098 & 0.056 & 0.056 & 0.064 & 0.088 & 0.091 & 0.122 & 0.083 \\
R3DG          & 0.214 & 0.058 & 0.090 & 0.125 & 0.013 & 0.170 & 0.112 & 0.085 & 0.125 & 0.089 & 0.081 & 0.080 & 0.117 & 0.156 & 0.115 & 0.106 \\
3DGS-DR       & 0.105 & \thirdbest{0.033} & \secondbest{0.076} & 0.050 & \first{0.006} & 0.082 & 0.059 & 0.052 & 0.050 & 0.042 & 0.057 & 0.048 & 0.068 & 0.059 & 0.060 & 0.054 \\
Ref-GS        & \thirdbest{0.098} & 0.034 & 0.082 & \first{0.045} & \first{0.006} & \thirdbest{0.070} & \thirdbest{0.056} & \thirdbest{0.042} & \thirdbest{0.049} & 0.041 & \thirdbest{0.046} & 0.046 & 0.076 & 0.073 & 0.064 & 0.055 \\
\reflectgs{}  & \thirdbest{0.098} & \thirdbest{0.033} & \secondbest{0.076} & 0.049 & \first{0.006} & 0.074 & 0.056 & \thirdbest{0.042} & \first{0.040} & \thirdbest{0.040} & 0.048 & \thirdbest{0.043} & \thirdbest{0.064} & \thirdbest{0.058} & \thirdbest{0.058} & \thirdbest{0.049} \\
MaterialRefGS & \secondbest{0.090} & \secondbest{0.032} & 0.076 & \thirdbest{0.048} & 0.009 & \secondbest{0.067} & \secondbest{0.054} & \secondbest{0.038} & 0.052 & \secondbest{0.036} & \secondbest{0.045} & \secondbest{0.042} & \secondbest{0.061} & \first{0.057} & \secondbest{0.053} & \secondbest{0.048} \\
Ours          & \first{0.089} & \first{0.031} & \first{0.076} & \secondbest{0.046} & 0.006 & \first{0.066} & \first{0.053} & \first{0.038} & \secondbest{0.047} & \first{0.036} & \first{0.044} & \first{0.041} & \first{0.061} & \secondbest{0.056} & \first{0.052} & \first{0.047} \\
\bottomrule[1pt]
\end{tabular}%
}
\label{tab:main_compare_synthetic}
\end{table*}
\begin{table*}[t]
\centering
\small
\caption{Per-scene image quality comparison on real-world datasets. ``Ours'' uses MaterialRefGS as the reflective Gaussian splatting backbone. The intensity of the red color signifies a better result.}
\resizebox{\textwidth}{!}{%
\begin{tabular}{lccccccccccccccc}
\toprule[1pt]
\textbf{Datasets} 
         & \multicolumn{4}{c}{\textbf{Ref-Real}~\citep{verbin2024refnerf}} 
         & \multicolumn{6}{c}{\textbf{Glossy-Real}~\citep{liu2023nero}} 
         & \multicolumn{5}{c}{\textbf{NeRF-casting}~\citep{verbin2024nerfcasting}} \\
\textbf{Scenes}   
         & \textbf{Garden} & \textbf{Sedan} & \textbf{Toycar} & \textbf{Avg.}
         & \textbf{Bear} & \textbf{Bunny} & \textbf{Coral} & \textbf{Maneki} & \textbf{Vase} & \textbf{Avg.}
         & \textbf{Compact} & \textbf{Grinder} & \textbf{Hatchback} & \textbf{Toaster} & \textbf{Avg.} \\
\midrule
\multicolumn{16}{c}{\textbf{PSNR} $\uparrow$} \\
\midrule
2DGS                      & 22.53 & \thirdbest{26.23} & 23.70 & 24.15 & 19.47 & 23.59 & 18.94 & 22.37 & 21.61 & 21.20 & 23.38 & 25.33 & 23.96 & 25.99 & 24.66 \\
Ref-GS                    & 22.48 & \first{26.63} & 24.20 & 24.44 & 17.17 & \first{26.02} & \thirdbest{19.09} & \secondbest{24.65} & \first{24.70} & \thirdbest{22.33} & \secondbest{28.19} & 30.46 & 25.27 & \secondbest{28.45} & \thirdbest{28.09} \\
\reflectgs{}              & \thirdbest{22.97} & \secondbest{26.60} & \thirdbest{24.27} & \thirdbest{24.61} & \thirdbest{19.89} & 24.56 & 18.83 & 23.70 & 23.01 & 22.00 & \first{28.32} & \thirdbest{30.67} & \secondbest{26.12} & \thirdbest{28.28} & \secondbest{28.35} \\
MaterialRefGS             & \first{23.29} & 26.04 & \secondbest{24.86} & \secondbest{24.73} & \secondbest{20.38} & \thirdbest{25.59} & \secondbest{19.32} & \thirdbest{24.26} & \thirdbest{23.70} & \secondbest{22.65} & 27.47 & \secondbest{30.89} & \thirdbest{25.77} & 28.16 & 28.07 \\
Ours                      & \secondbest{23.21} & 26.13 & \first{24.89} & \first{24.74} & \first{20.79} & \secondbest{25.72} & \first{19.66} & \first{24.66} & \secondbest{24.39} & \first{23.04} & \thirdbest{27.66} & \first{31.56} & \first{26.20} & \first{29.34} & \first{28.69} \\
\midrule
\multicolumn{16}{c}{\textbf{SSIM} $\uparrow$} \\
\midrule
2DGS                      & 0.609 & \secondbest{0.778} & 0.597 & 0.661 & 0.750 & 0.861 & \first{0.762} & 0.842 & 0.806 & 0.804 & 0.828 & 0.787 & 0.799 & 0.871 & 0.822 \\
Ref-GS                    & 0.607 & \first{0.783} & 0.656 & 0.682 & 0.704 & \first{0.892} & 0.750 & \first{0.875} & \first{0.851} & \thirdbest{0.814} & \first{0.863} & \secondbest{0.822} & \first{0.833} & 0.902 & \first{0.855} \\
\reflectgs{}              & \thirdbest{0.617} & \thirdbest{0.777} & \thirdbest{0.660} & \thirdbest{0.685} & \thirdbest{0.752} & 0.861 & 0.743 & 0.846 & 0.808 & 0.802 & \secondbest{0.857} & 0.819 & \thirdbest{0.819} & \thirdbest{0.903} & \thirdbest{0.849} \\
MaterialRefGS             & \first{0.645} & 0.765 & \first{0.685} & \secondbest{0.698} & \secondbest{0.768} & \thirdbest{0.880} & \thirdbest{0.754} & \thirdbest{0.858} & \thirdbest{0.830} & \secondbest{0.818} & 0.843 & \thirdbest{0.821} & 0.812 & \secondbest{0.903} & 0.845 \\
Ours                      & \secondbest{0.642} & 0.770 & \secondbest{0.684} & \first{0.699} & \first{0.769} & \secondbest{0.885} & \secondbest{0.757} & \secondbest{0.865} & \secondbest{0.843} & \first{0.824} & \thirdbest{0.847} & \first{0.829} & \secondbest{0.823} & \first{0.915} & \secondbest{0.853} \\
\midrule
\multicolumn{16}{c}{\textbf{LPIPS} $\downarrow$} \\
\midrule
2DGS                      & 0.254 & 0.225 & 0.396 & 0.292 & 0.275 & 0.203 & \thirdbest{0.252} & 0.236 & 0.248 & 0.243 & 0.235 & 0.254 & 0.244 & 0.166 & 0.225 \\
Ref-GS                    & \thirdbest{0.242} & \thirdbest{0.196} & \thirdbest{0.236} & \thirdbest{0.225} & 0.332 & \thirdbest{0.164} & 0.263 & \thirdbest{0.190} & \thirdbest{0.189} & \thirdbest{0.228} & \thirdbest{0.178} & \thirdbest{0.221} & \thirdbest{0.187} & \thirdbest{0.154} & \thirdbest{0.185} \\
\reflectgs{}              & 0.256 & 0.245 & 0.256 & 0.252 & \thirdbest{0.273} & 0.218 & 0.276 & 0.243 & 0.261 & 0.254 & 0.215 & 0.244 & 0.230 & 0.157 & 0.211 \\
MaterialRefGS             & \first{0.180} & \secondbest{0.179} & \first{0.191} & \first{0.183} & \secondbest{0.233} & \secondbest{0.133} & \first{0.212} & \secondbest{0.173} & \secondbest{0.178} & \secondbest{0.186} & \secondbest{0.159} & \secondbest{0.156} & \secondbest{0.167} & \secondbest{0.110} & \secondbest{0.148} \\
Ours                      & \secondbest{0.184} & \first{0.174} & \first{0.191} & \first{0.183} & \first{0.232} & \first{0.130} & \secondbest{0.233} & \first{0.164} & \first{0.161} & \first{0.184} & \first{0.152} & \first{0.143} & \first{0.154} & \first{0.099} & \first{0.137} \\
\bottomrule[1pt]
\end{tabular}%
}
\label{tab:main_compare_real}
\end{table*}

\begin{table*}[t]
\caption{Mean Angular Error of normals on Shiny Blender dataset. Lower is better. The intensity of the red color signifies a better result.}
\scriptsize
\centering
\begin{tabular}{lccccccc}
\toprule
 & \textbf{Car} & \textbf{Ball} & \textbf{Helmet} & \textbf{Teapot} & \textbf{Toaster} & \textbf{Coffee} & \textbf{Average} \\
\midrule
GS-IR
    & 28.31 & 25.79 & 25.58 & 15.35 & 33.51 & 15.38 & 23.99 \\
GShader
    & 14.05 & 7.03 & 9.33 & 7.17 & 13.08 & 14.93 & 10.93 \\
3DGS-DR
    & 2.32 & \thirdbest{0.85} & \thirdbest{1.67} & \thirdbest{0.53} & 6.99 & \thirdbest{2.21} & 2.43 \\
Ref-GS
    & 2.02 & 1.05 & 1.99 & 0.69 & 3.92 & 3.61 & 2.21 \\
\reflectgs{}
    & \thirdbest{1.77} & 1.17 & 2.07 & 0.70 & \thirdbest{3.90} & 3.07 & \thirdbest{2.11} \\
MaterialRefGS
    & \secondbest{1.55} & \secondbest{0.65} & \first{1.18} & \secondbest{0.43} & \secondbest{2.51} & \secondbest{1.72} & \secondbest{1.34} \\
Ours
    & \first{1.48} & \first{0.63} & \secondbest{1.19} & \first{0.38} & \first{2.25} & \first{1.70} & \first{1.27} \\    
\bottomrule[1pt]
\end{tabular}
\label{tab:normal_mae}
\end{table*}

\subsection{Clay-guided reflective object reconstruction results}
\begin{figure}[t]
  \centering
  \setlength{\tabcolsep}{2pt}

  \begin{tabular}{>{\centering\arraybackslash}p{0.16\linewidth}
                  >{\centering\arraybackslash}p{0.16\linewidth}
                  >{\centering\arraybackslash}p{0.16\linewidth}
                  >{\centering\arraybackslash}p{0.16\linewidth}
                  >{\centering\arraybackslash}p{0.16\linewidth}}

    \includegraphics[width=\linewidth]{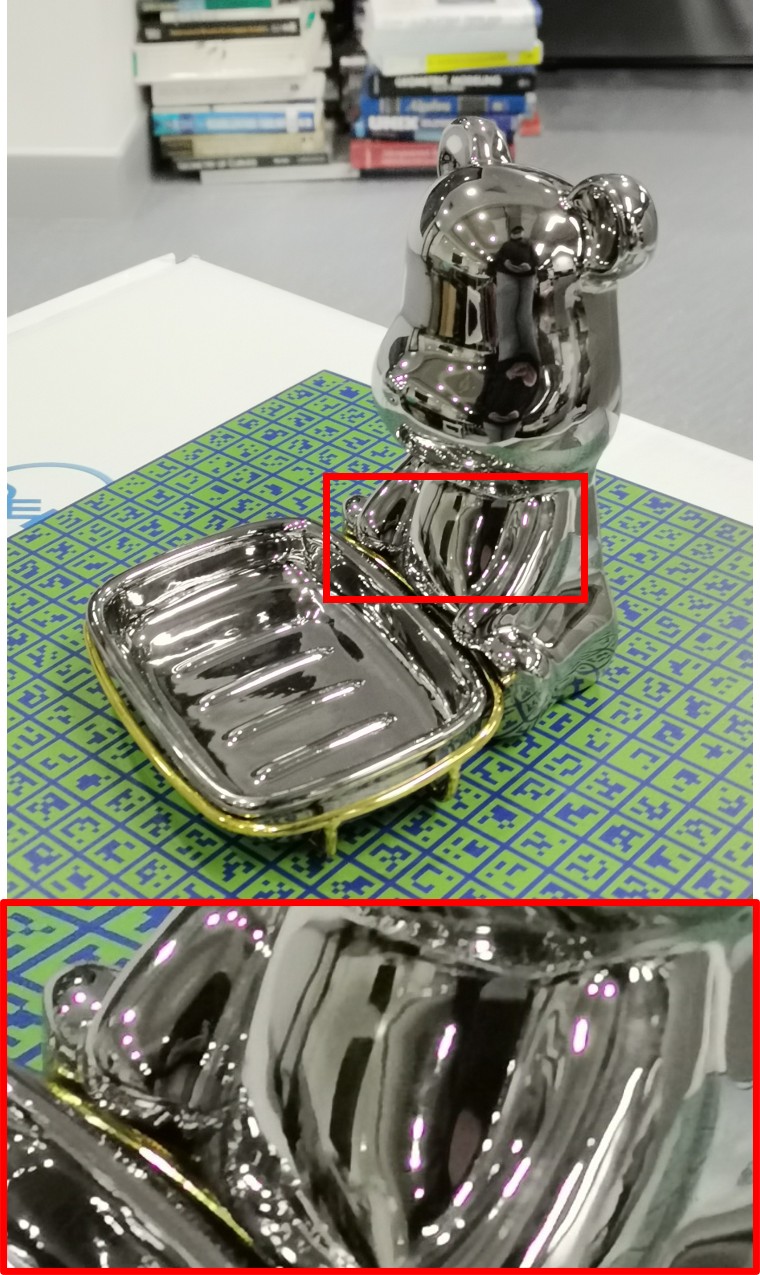} &
    \includegraphics[width=\linewidth]{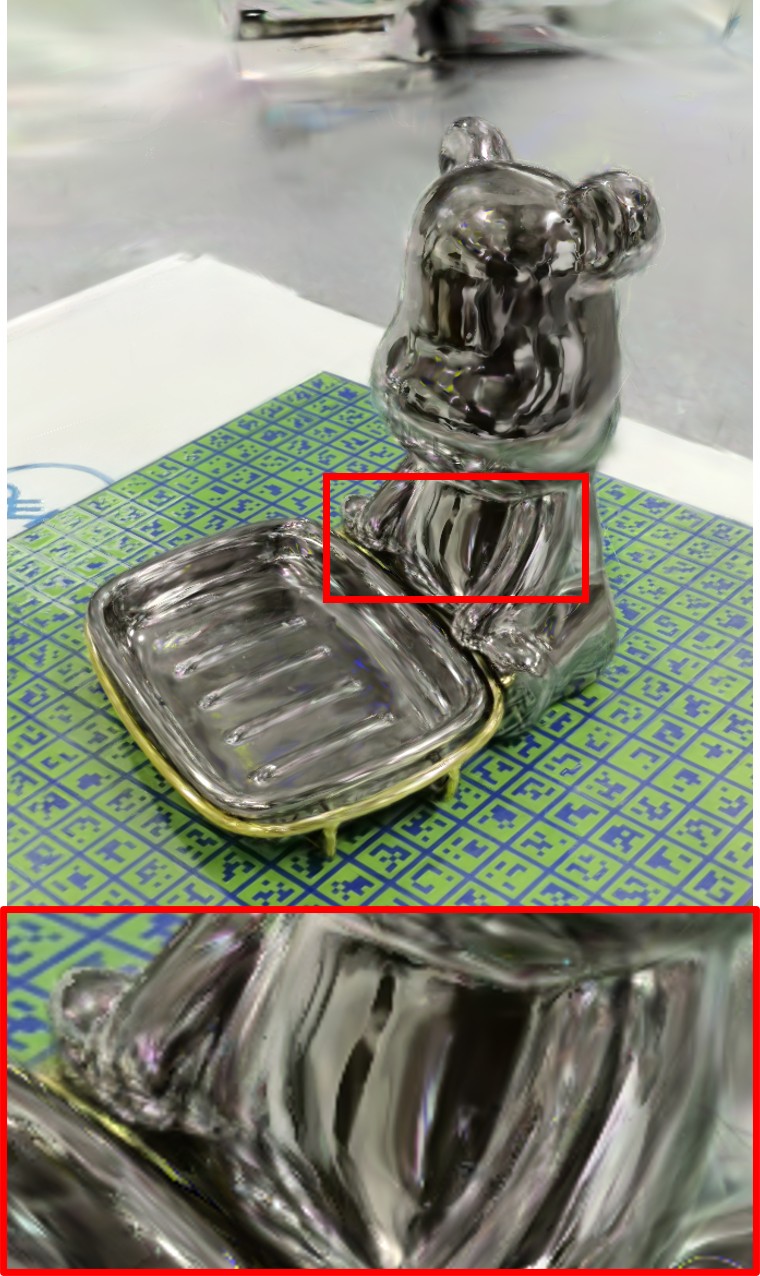} &
    \includegraphics[width=\linewidth]{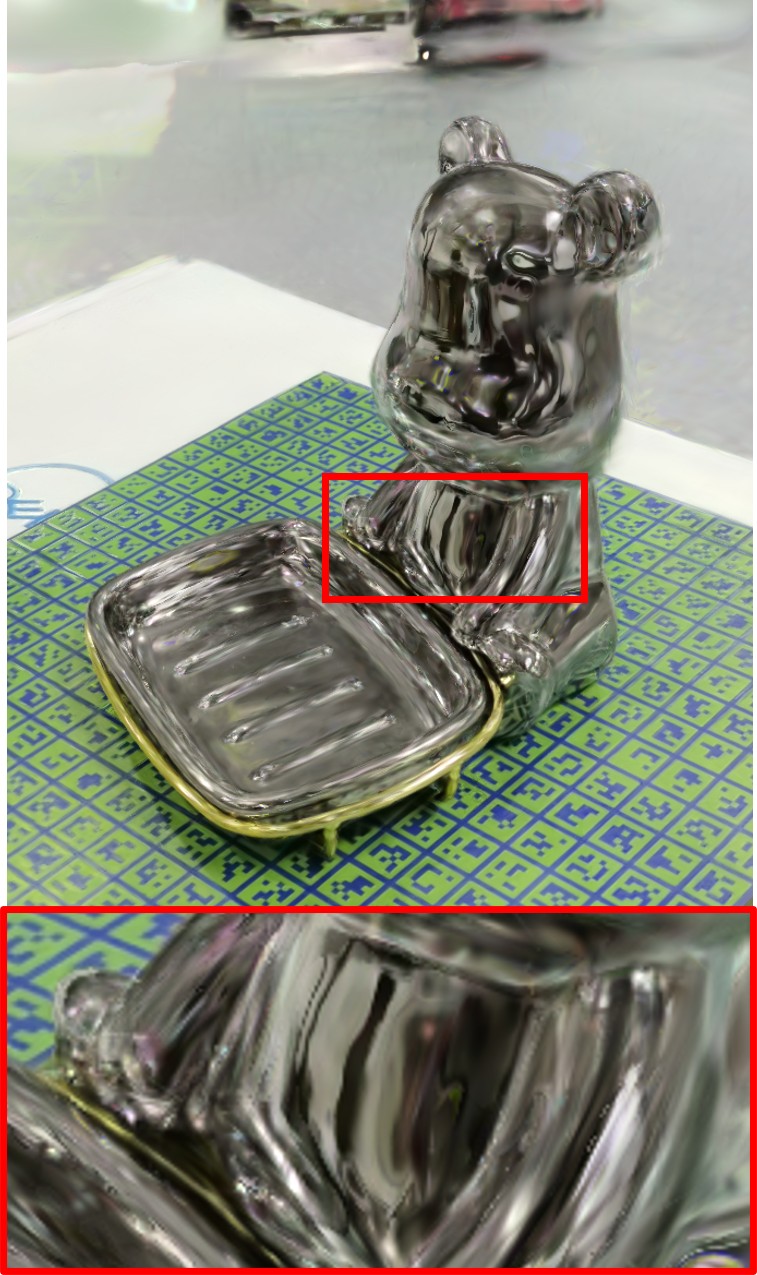} &
    \includegraphics[width=\linewidth]{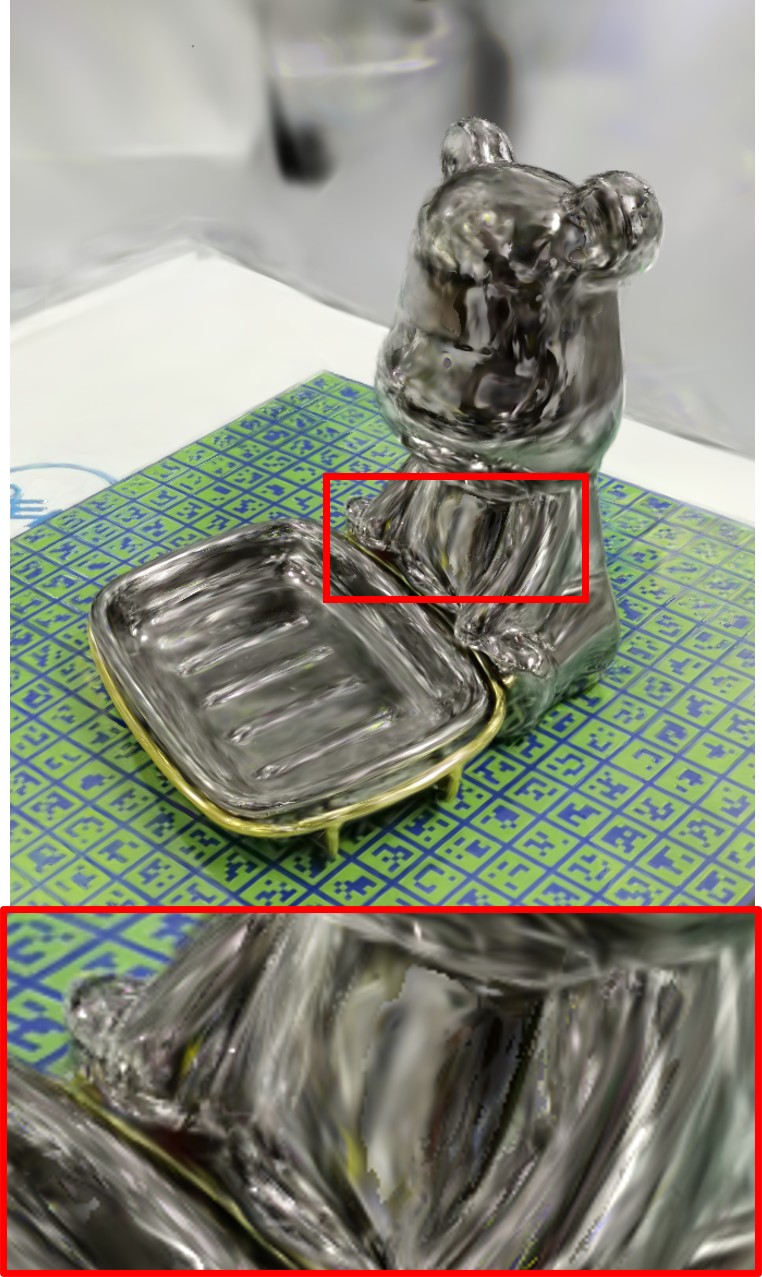} &
    \includegraphics[width=\linewidth]{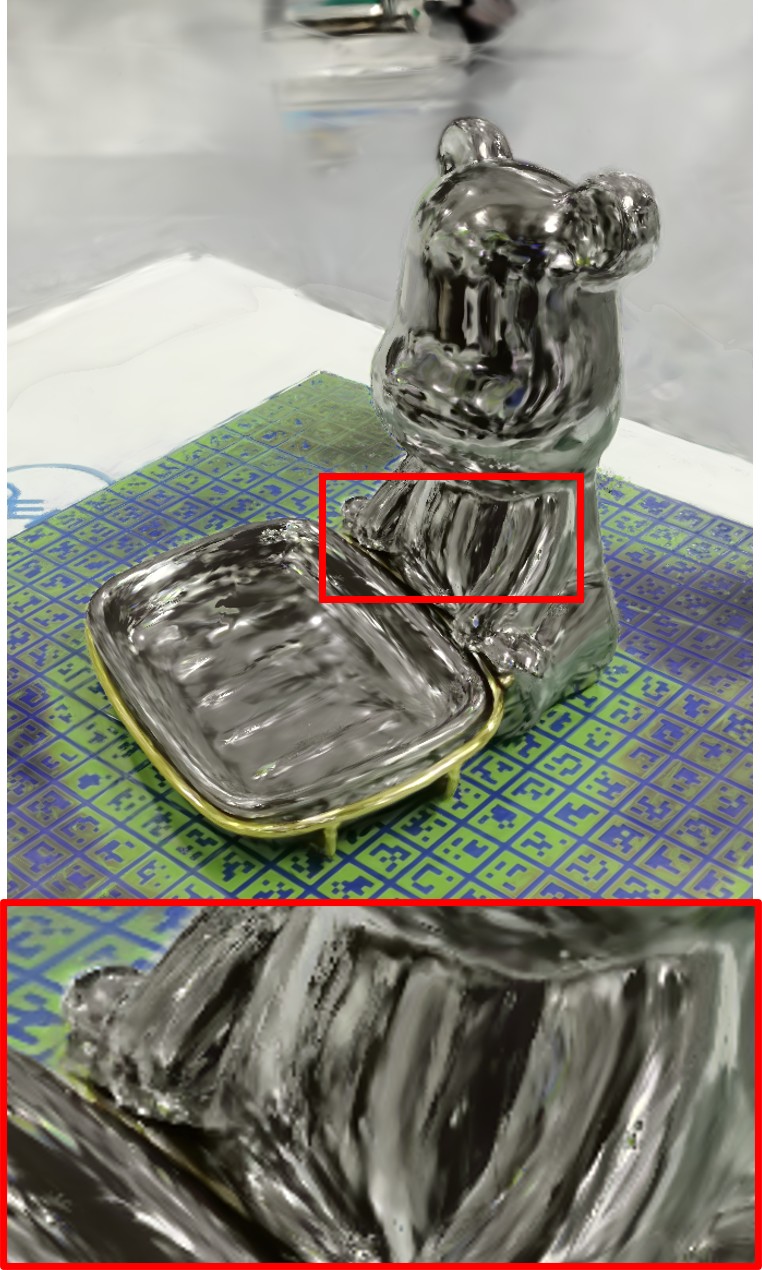} \\[2pt]

    \includegraphics[width=\linewidth]{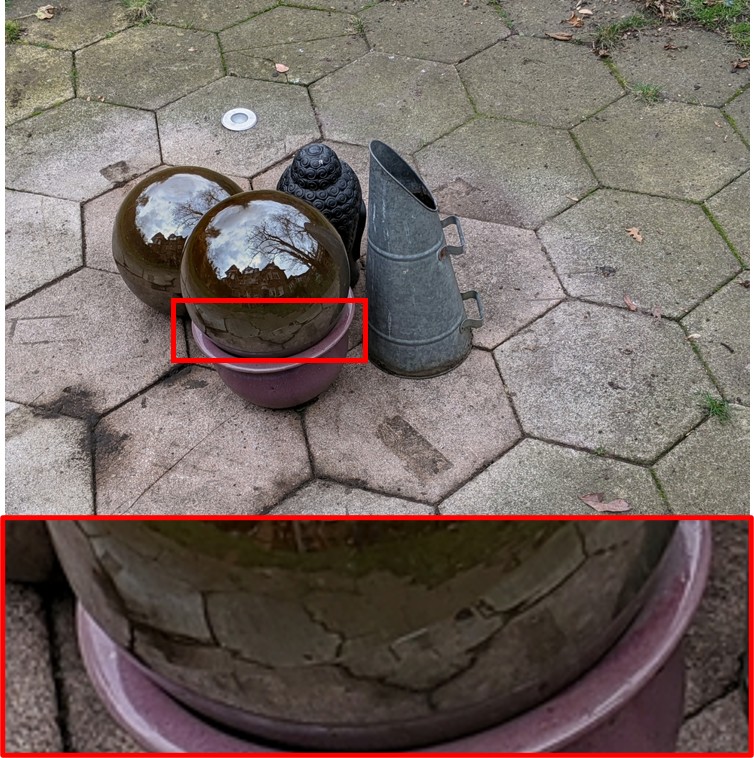} &
    \includegraphics[width=\linewidth]{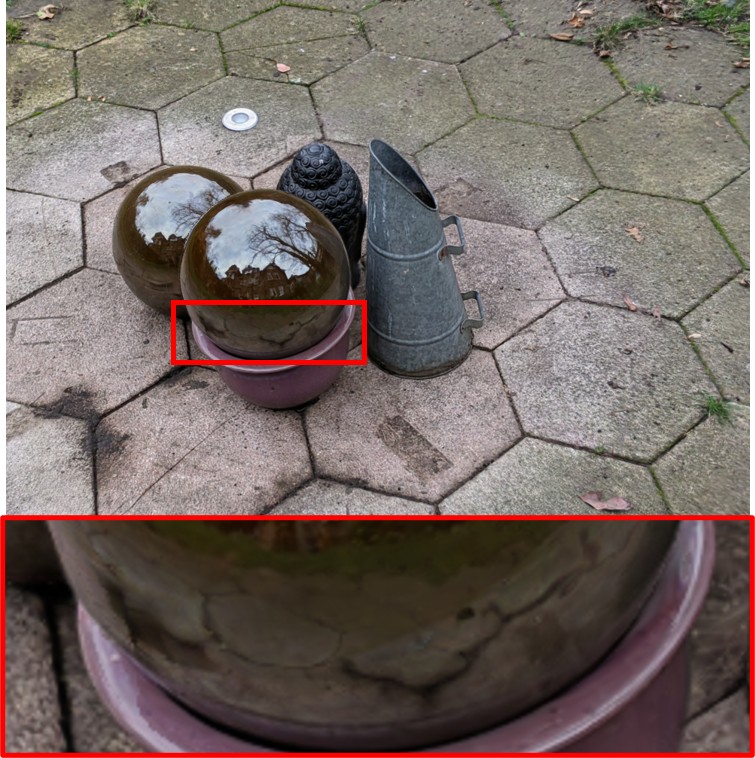} &
    \includegraphics[width=\linewidth]{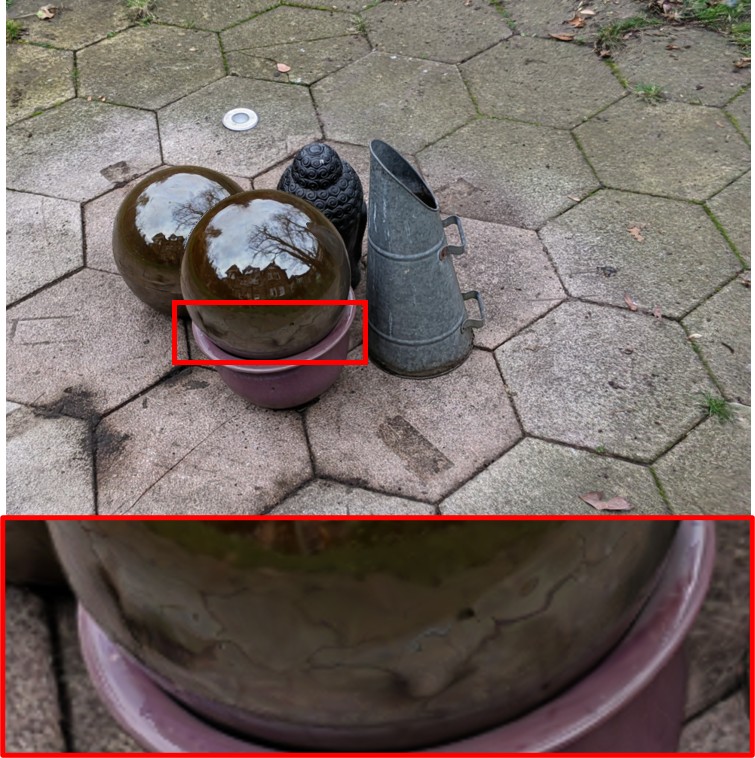} &
    \includegraphics[width=\linewidth]{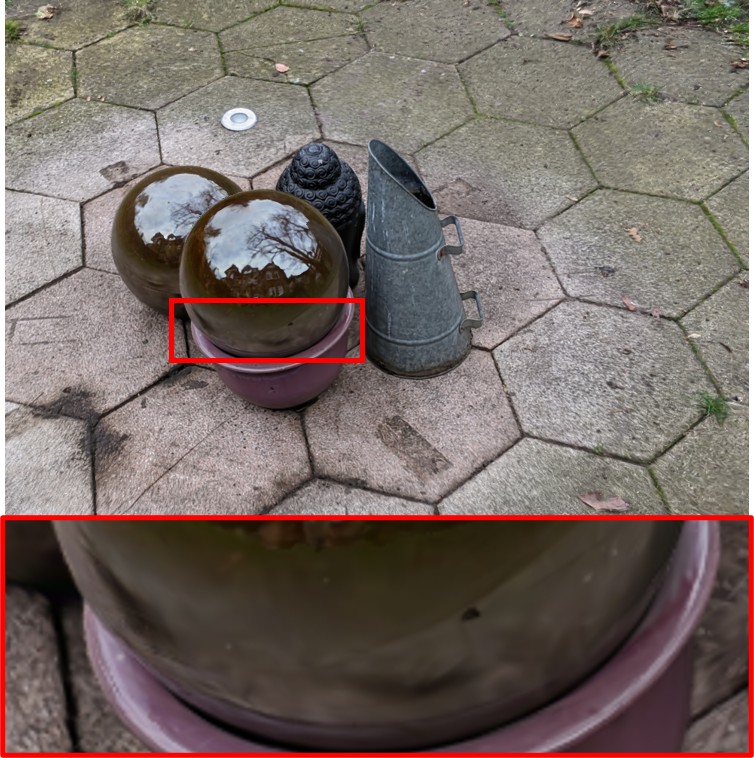} &
    \includegraphics[width=\linewidth]{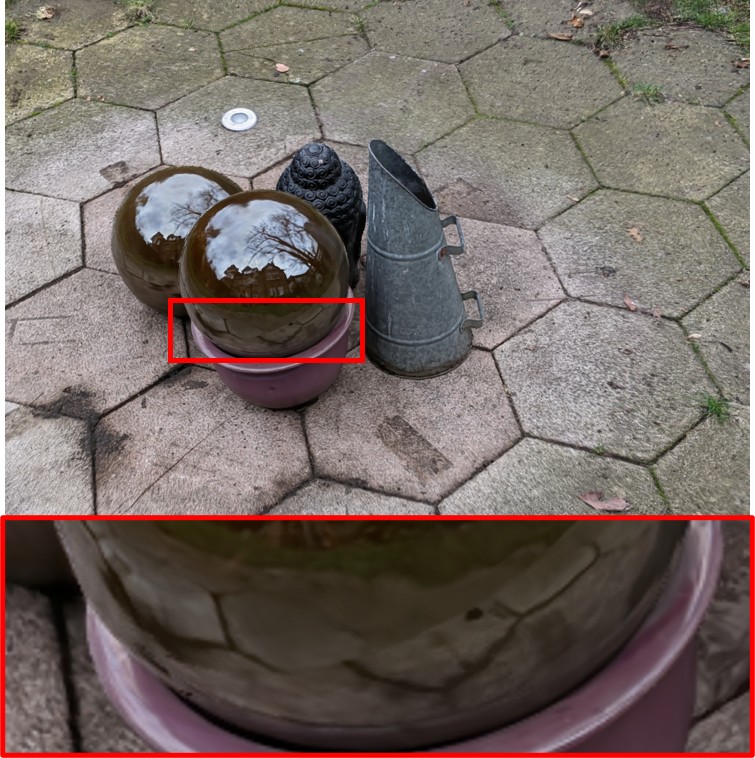} \\[2pt]
    \includegraphics[width=\linewidth]{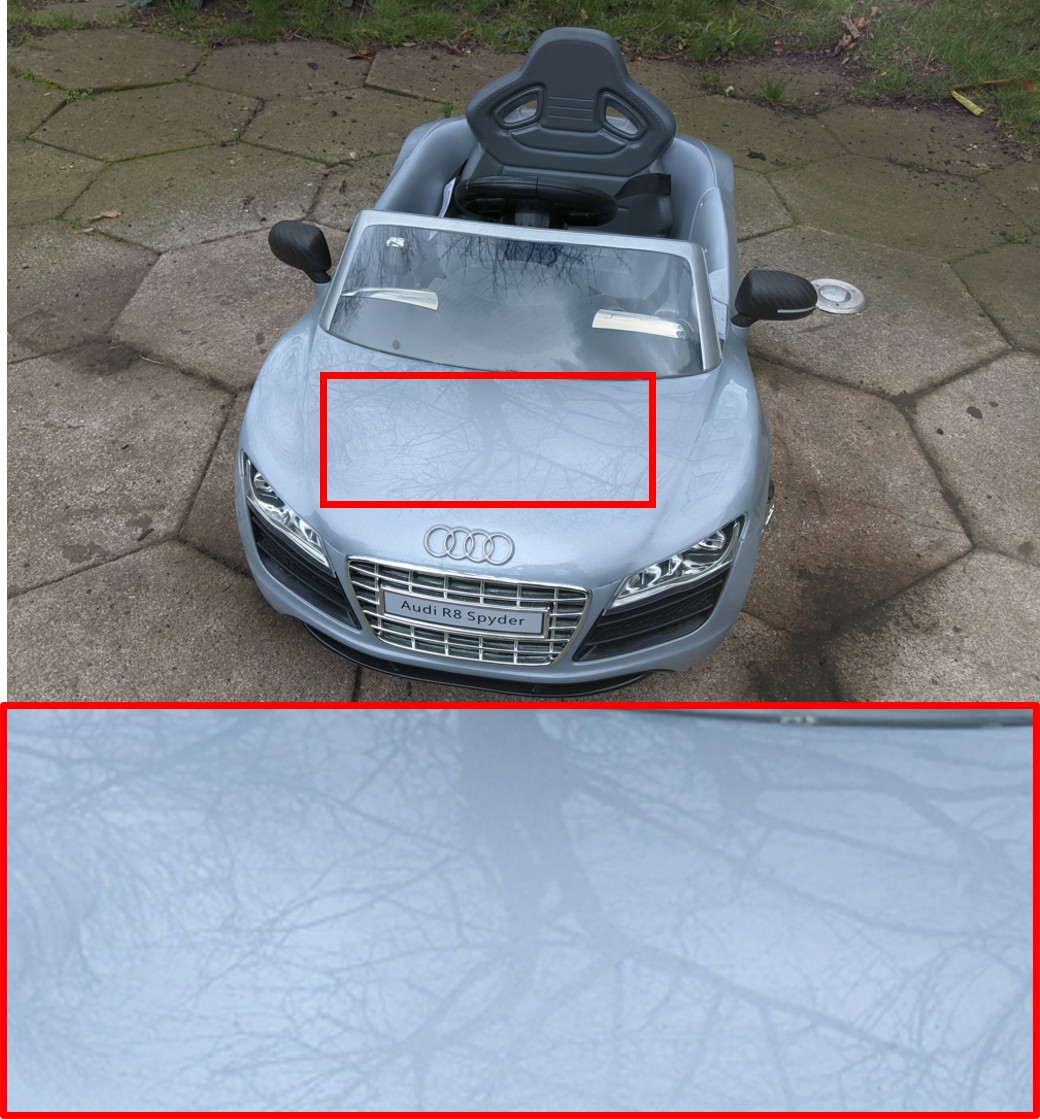} &
    \includegraphics[width=\linewidth]{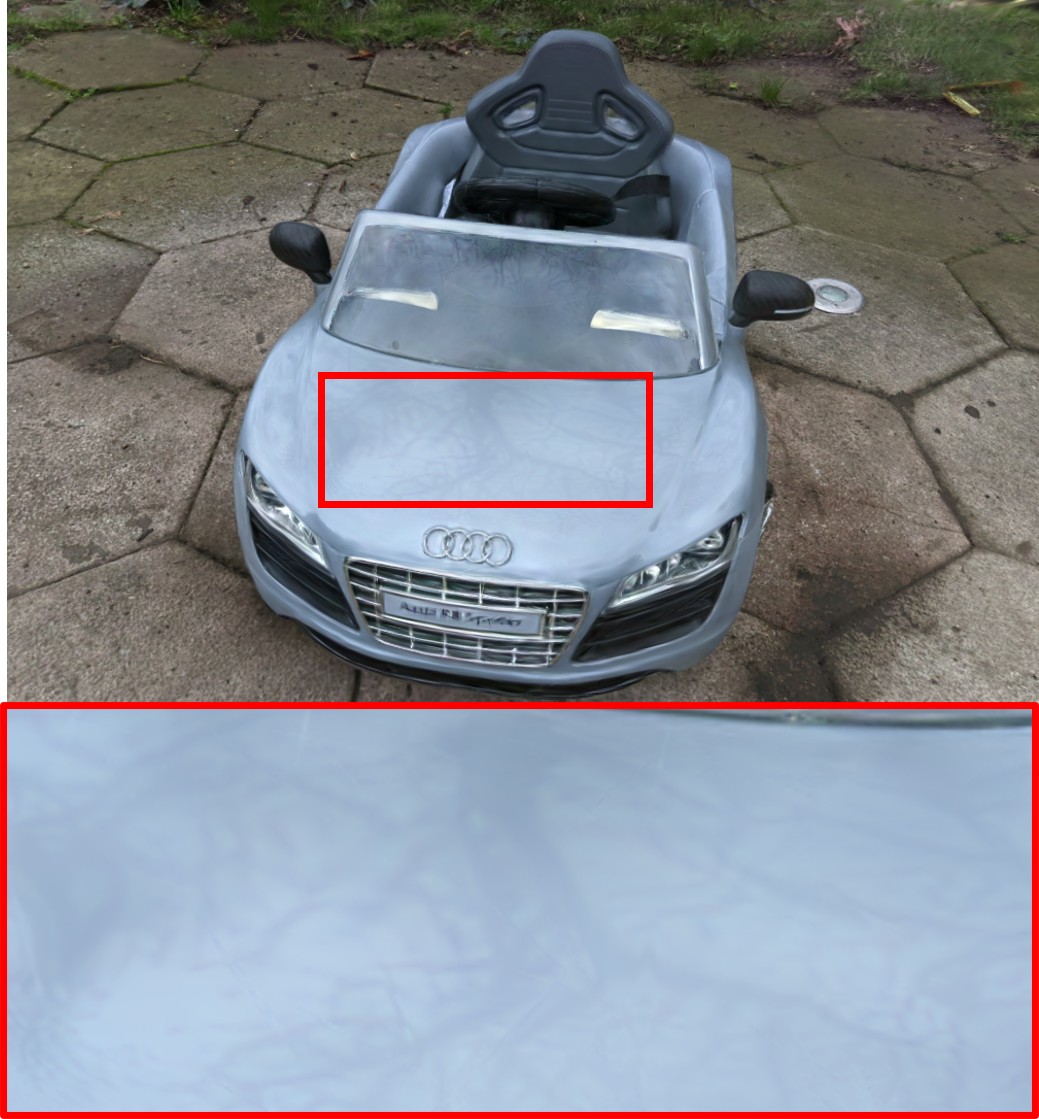} &
    \includegraphics[width=\linewidth]{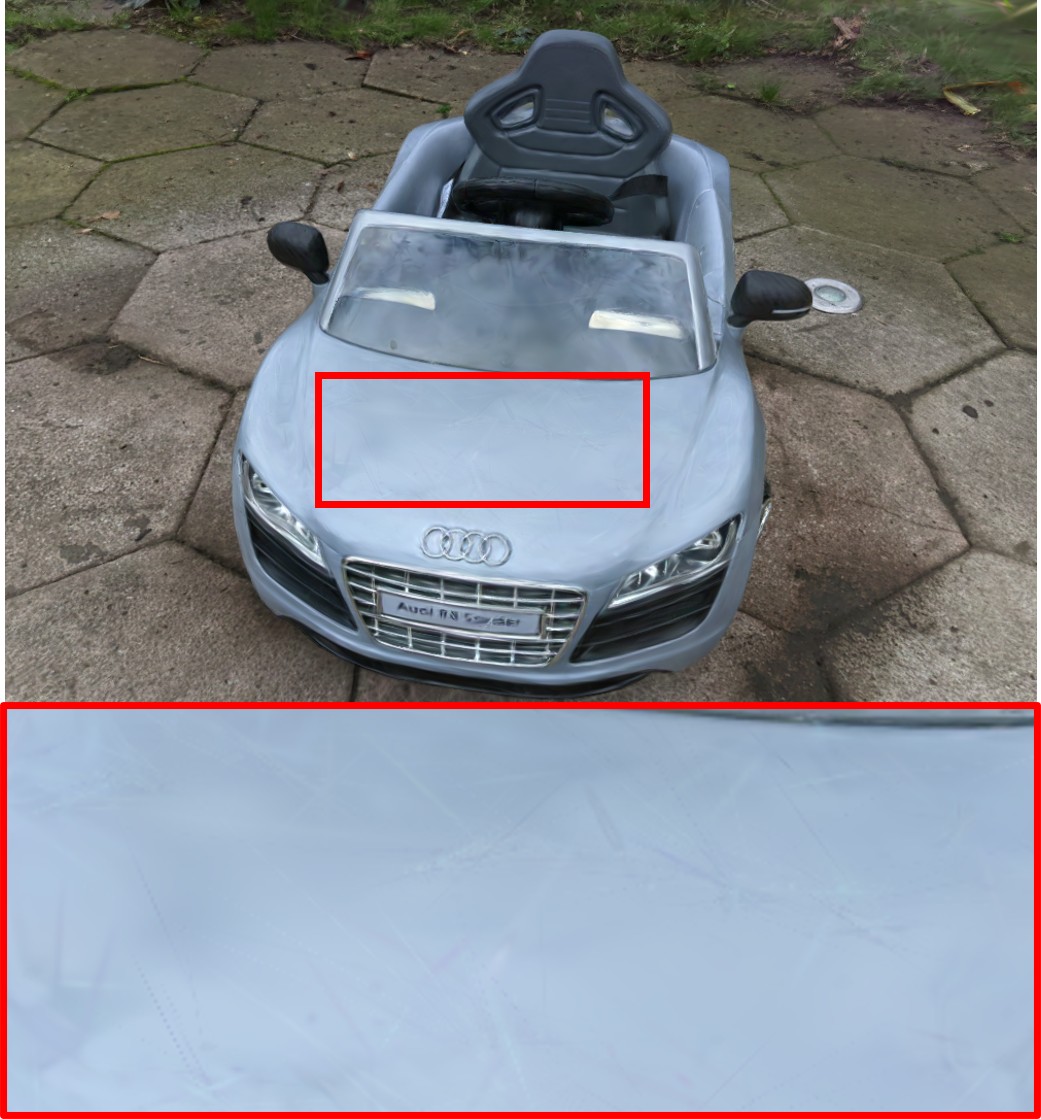} &
    \includegraphics[width=\linewidth]{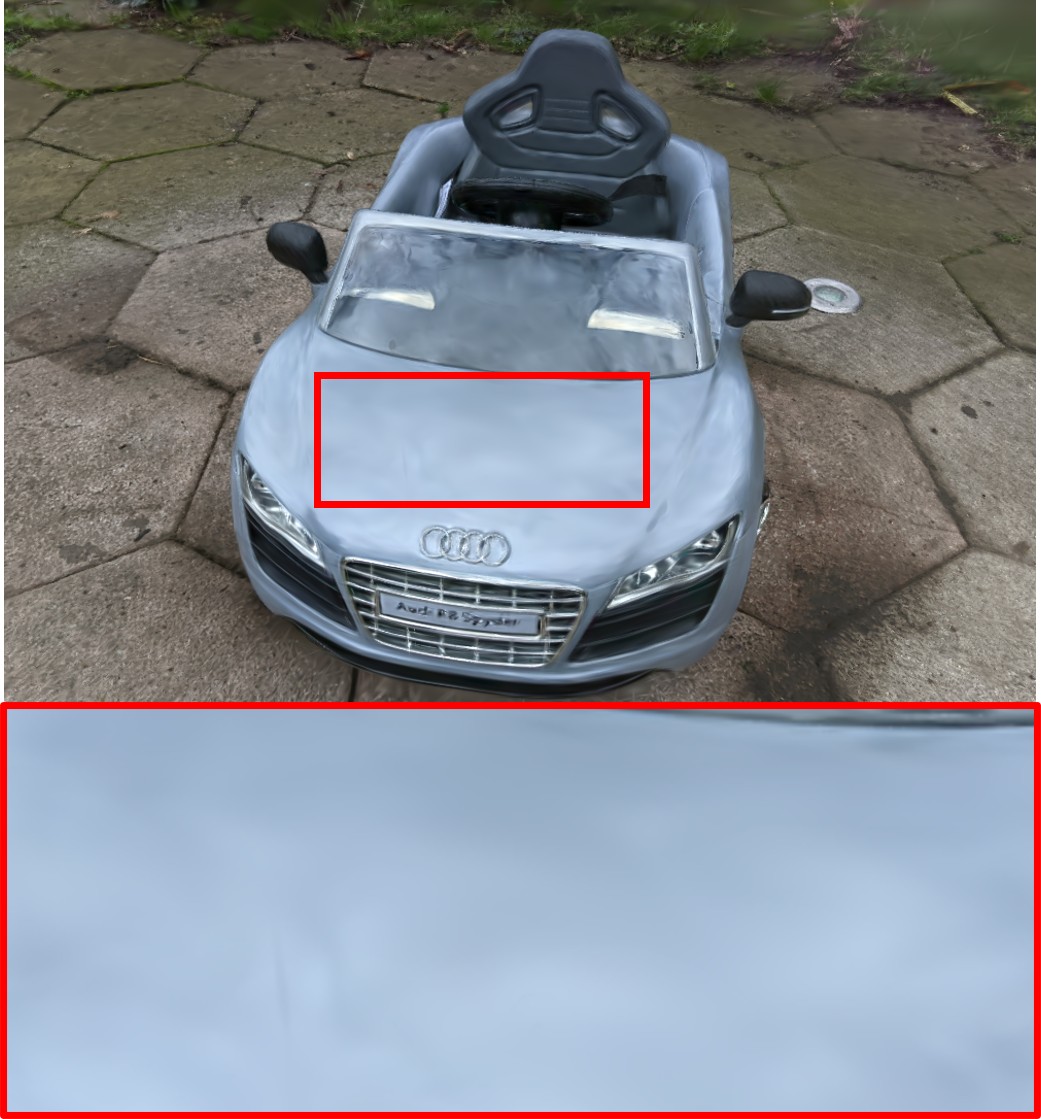} &
    \includegraphics[width=\linewidth]{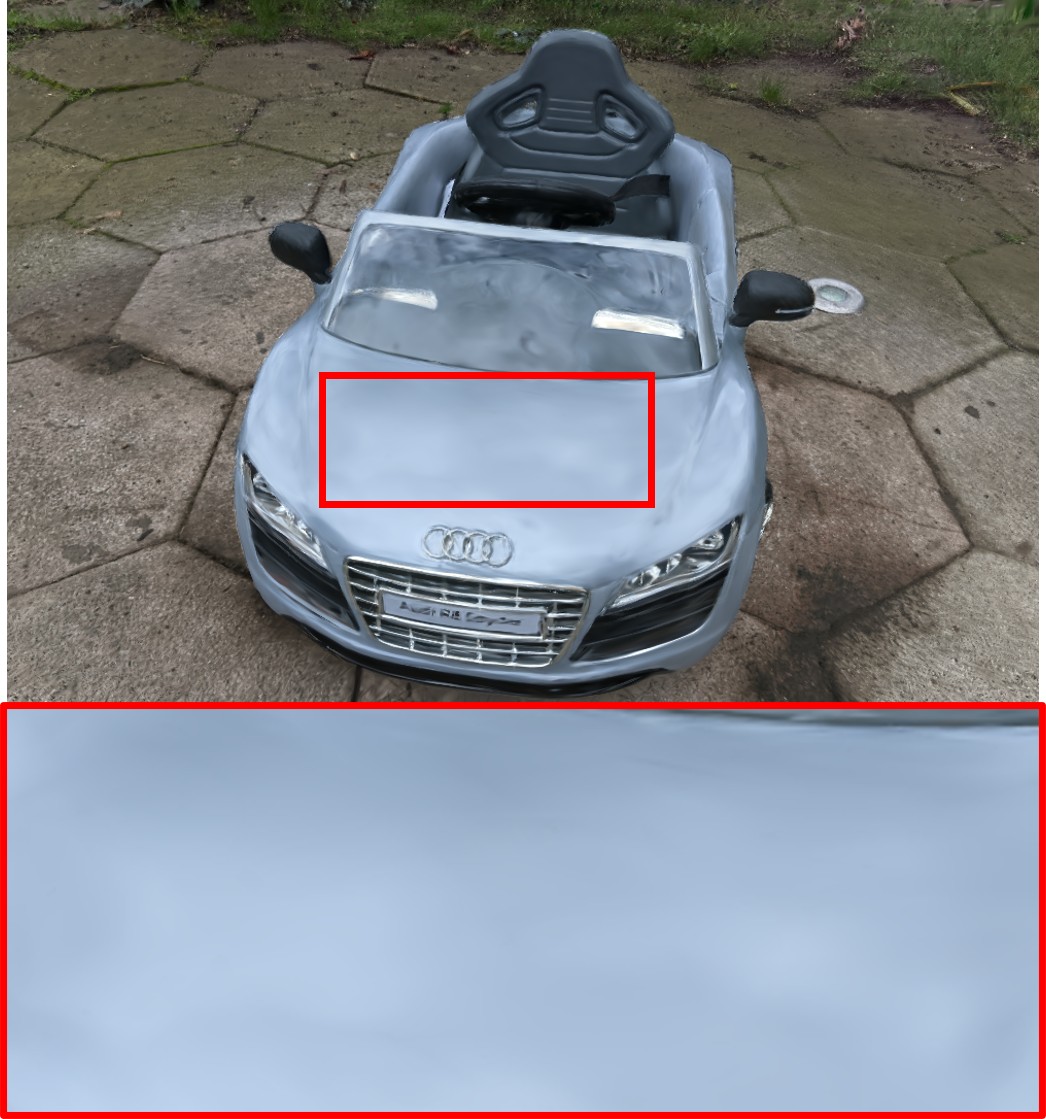} \\[2pt]

    GT & Ours & MaterialRefGS & RGS &  Ref-GS

  \end{tabular}

  \caption{More qualitative comparison of our method with baselines.}

  \label{fig:additional_qual}
\end{figure}

We provide full per-scene photometric results corresponding to the abbreviated \cref{tab:main_compare_avg} in \cref{tab:main_compare_synthetic,tab:main_compare_real}. We also report the Mean Angular Error (MAE) of surface normals in \cref{tab:normal_mae}. Additional qualitative comparisons are shown in \cref{fig:additional_qual}.

\begin{figure*}[t]
  \centering
  \includegraphics[width=0.95\linewidth]{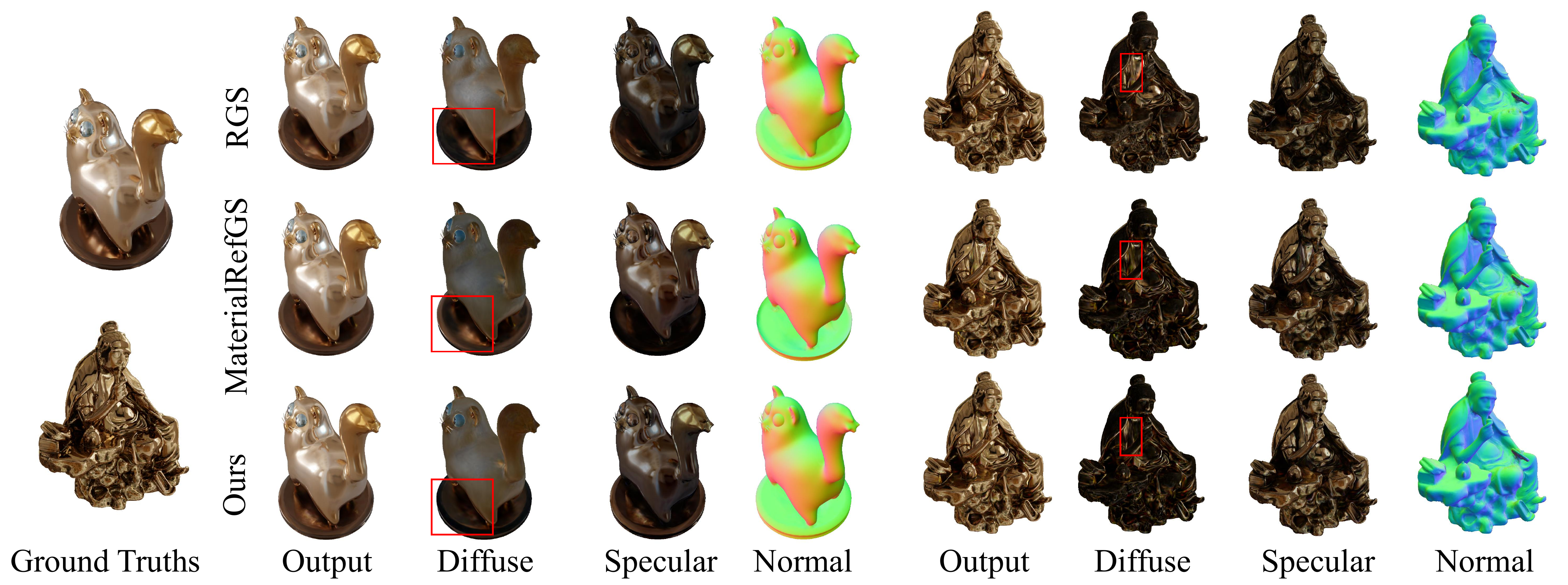}
  \caption{Material visualization. We visualize the rendered output, diffuse color, specular color, and surface normal separately. The red boxes show reflections that are incorrectly baked into the diffuse color instead of the specular color.}

  \label{fig:material_split}
\end{figure*}

\subsection{More analysis}
\begin{figure*}[t]
  \centering
  \includegraphics[width=0.8\linewidth]{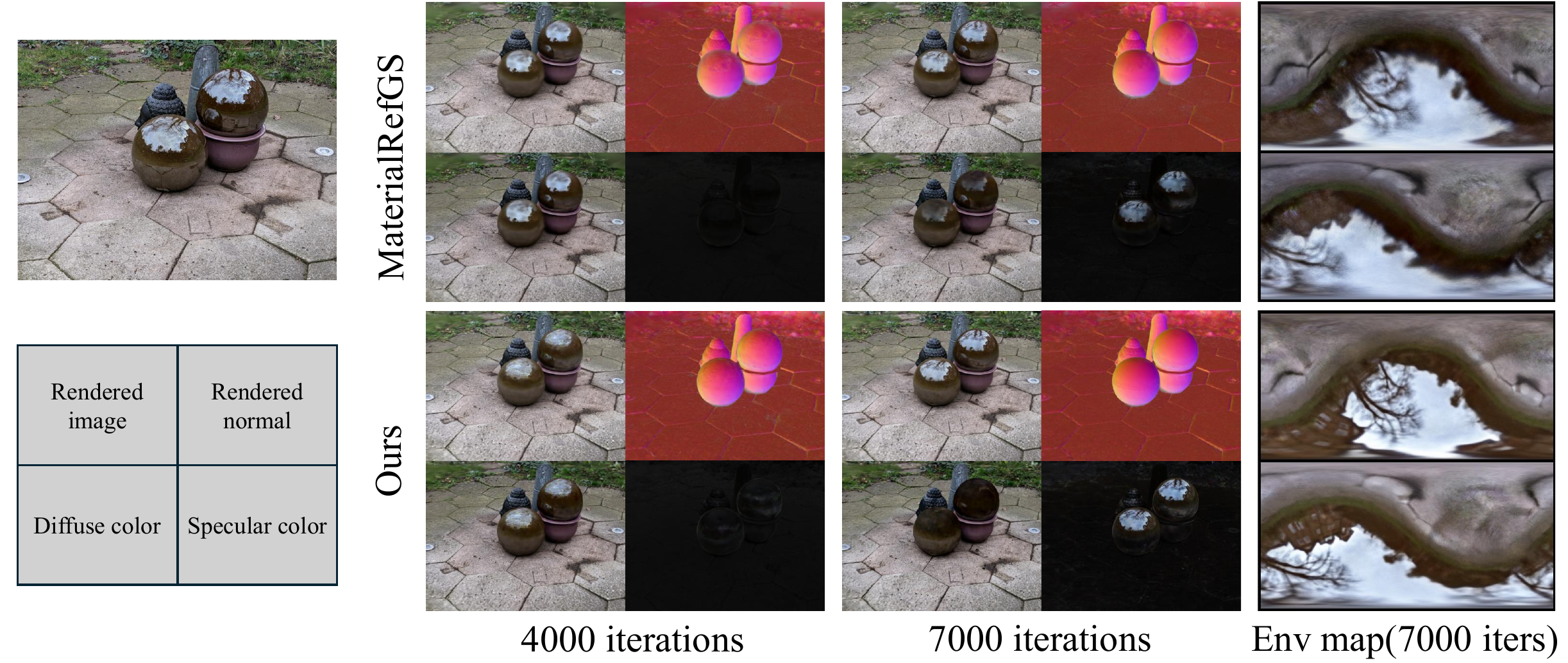}
  \caption{Convergence comparison during training. At both $4{,}000$ and $7{,}000$ iterations, our method produces fewer reflections baked into the diffuse color, and a more faithful environment map than MaterialRefGS, despite using the same normal supervision.}
  \label{fig:vis_convergence}
\end{figure*}

To illustrate why the clay loss benefits geometry learning, we visualize the diffuse color, specular color, and rendered normals at intermediate training iterations in \cref{fig:vis_convergence}. Both methods share the same normal supervision. However, the clay pass bypasses specular evaluation and thus strongly regularizes geometry, preventing Gaussians from distorting their normals and opacities to entangle view-dependent reflections into the diffuse color. Consequently, the environment map also converges faster with early clay supervision. This rapid geometric stabilization improves geometry metrics (\cref{tab:exp_geometry_grouped} and \cref{tab:normal_mae}) and allows the BRDF branch to focus on material refinement, yielding better qualitative results.
In \cref{fig:material_split}, we visualize material decomposition. RGS and MaterialRefGS bake reflections into the diffuse color (red boxes), indicating poor disentanglement between geometry and view-dependent appearance. Our method instead produces clean, view-independent diffuse colors and sharper normal maps, confirming that clay-guided supervision improves both geometry and material decomposition.

\begin{table*}[h]
\caption{Reconstruction on Glossy Synthetic with different diffuse guidance.}
\label{tab:s2dnet}
\centering \scriptsize
\begin{tabular}{ccccc}
\toprule[1pt]
 & CD $\downarrow$ & PSNR $\uparrow$ & SSIM $\uparrow$ & LPIPS $\downarrow$ \\
\midrule
Baseline (MaterialRefGS) & 0.0067 & 30.60 & 0.960 & 0.048 \\
Ours w/ S2Dnet diffuse images & 0.0229 & 28.62 & 0.943 & 0.065 \\
Ours w/ our clay images & \textbf{0.0054} & \textbf{30.68} & \textbf{0.961} & \textbf{0.047} \\
\bottomrule[1pt]
\end{tabular}
\end{table*}
\begin{table*}[t]
\centering
\caption{Ablation on selective gradient routing on the Shiny Blender dataset. We jointly ablate (i) which Gaussian parameters are routed (position, opacity, rotation), (ii) whether a weak residual RGB gradient is applied to the routed parameters, and (iii) the lock duration. Our default configuration is highlighted.}
\scriptsize
\setlength{\tabcolsep}{4pt}
\renewcommand{\arraystretch}{1.1}
\begin{tabular}{cccc c cc}
\toprule[1pt]
\multicolumn{4}{c}{Routed parameters} & \multirow{2}{*}{Lock duration} & \multirow{2}{*}{MAE $\downarrow$} & \multirow{2}{*}{PSNR $\uparrow$} \\
\cmidrule(lr){1-4}
Position & Opacity & Rotation & Weak RGB & & & \\
\midrule
-          & -          & -          & -          & 10K  & 1.32 & 35.01 \\
\checkmark & -          & -          & -          & 10K  & 1.29 & 35.04 \\
\checkmark & \checkmark & -          & -          & 10K  & 1.30 & 35.05 \\
\checkmark & \checkmark & \checkmark & -          & 10K  & 1.31 & 34.94 \\
\checkmark & \checkmark & -          & \checkmark & 5K   & 1.33 & 34.89 \\
\checkmark & \checkmark & -          & \checkmark & 10K  & 1.29 & 35.06 \\
\rowcolor{gray!15}
\checkmark & \checkmark & -          & \checkmark & 20K  & \textbf{1.27} & \textbf{35.12} \\
\checkmark & \checkmark & -          & \checkmark & 30K  & 1.33 & 33.77 \\
\bottomrule[1pt]
\end{tabular}
\label{tab:ablation_routing}
\end{table*}

Here we provide details on the selective gradient routing ablation summarized in \cref{tab:ablation_routing}. All ablations are conducted on the Shiny Blender dataset varying only the routing configuration. As shown in the table, routing both position and opacity gives the best PSNR, and the best overall configuration adds a weak residual RGB gradient with a $20$K-iteration lock period (PSNR $35.12$, MAE $1.27$). Locking for $30$K iterations causes a sharp drop (PSNR $33.77$). When the routing period is too long, the BRDF branch is left with too few iterations to refine the surface geometry using photometric supervision from the RGB images.

\paragraph{View ordering within chunks.}
To separate the effect of view diversity from that of the specific ordering, we additionally evaluate a stratified chunking strategy that draws the views of each chunk from strata over the ordered camera trajectory, guaranteeing angular diversity within every chunk. Evaluated with the same model trained under random ordering, it performs on par with random ordering, and both clearly outperform sorted ordering. This indicates that the key factor behind the benefit of random ordering is view diversity within each chunk rather than the ordering itself. We keep random ordering as the default since exposure to diverse view orders during training additionally acts as data augmentation, while stratified chunking remains a supported inference-time option.

\paragraph{Comparison with specular-to-diffuse translation.}
\label{appx:s2dnet}
We empirically compare with S2Dnet~\cite{wu2018specular} by running its
released pretrained model on Glossy Synthetic. Its diffuse conversion leaves
reflective regions largely intact in most scenes, presumably because its
training data lacks cases where the environment is mirrored on the object
surface. Training our framework with these images degrades reconstruction
below the baseline (\cref{tab:s2dnet}): inaccurate diffuse guidance is worse
than none, underscoring that faithful translation on challenging reflective
data, enabled by the video prior, is central to our approach.

\section{Broader impacts and limitations}

\subsection{Broader social impacts}
\label{appx:social_impacts}
Our work focuses on 3D reconstruction of reflective objects, which can be used for various applications such as digital content creation, AR/VR, and robotic perception. While 3D reconstruction in general could be misused for negative purposes such as military applications, as foundational research, we do not foresee direct negative societal impacts from our specific contribution.

\subsection{Limitations}
\label{sec:limitation}
The quality of geometry learning is bounded by the accuracy of the image-to-clay translation model: incorrect translations can propagate into geometry optimization, particularly for materials or lighting conditions underrepresented in the training data. Training the translator on a broader range of materials and lighting conditions would likely further improve our method. In addition, the clay translation step introduces extra pre-processing cost, although it is negligible compared to the optimization phase. Translations of non-salient objects or multiple objects may fail because the translation model was trained on paired images of single foreground objects and their clay counterparts. Finally,
while the method remains effective with moderately sparse inputs (8 views),
extremely sparse captures (4 or 2 views) break the structure preservation of
the translation.


\end{document}